%% file: main.tex
\documentclass[10pt,twocolumn,letterpaper]{article}

\usepackage{cvpr}              

\input{preamble}

\definecolor{cvprblue}{rgb}{0.21,0.49,0.74}
\usepackage[pagebackref,breaklinks,colorlinks,allcolors=cvprblue]{hyperref}

\def\paperID{} 
\def\confName{CVPR}
\def\confYear{2026}

\title{Stochastic Ray Tracing for the Reconstruction of 3D Gaussian Splatting}

\author{
    Peiyu Xu$^1$ \quad
    Xin Sun$^2$ \quad
    Krishna Mullia$^{2,3}$ \quad
    Raymond Fei$^2$ \quad
    Iliyan Georgiev$^2$ \quad
    Shuang Zhao$^1$ \\[.5em]
    $^1$University of Illinois Urbana-Champaign \qquad $^2$Adobe Research \qquad
    $^3$Canva Research
}

\begin{document}

\vspace{-3em}
\twocolumn[{%
		\renewcommand\twocolumn[1][]{#1}%
		\maketitle
		\input{sec/teaser}
		\vspace{3em}
	}]

\input{sec/abstract}
\input{sec/intro}
\input{sec/related}
\input{sec/preliminaries}
\input{sec/ours}
\input{sec/results}
\input{sec/conculsion}
\clearpage
\input{sec/result_figures}

\clearpage

\appendix
\input{sec/X_suppl}

{
	\small
	\bibliographystyle{ieeenat_fullname}
	\bibliography{main}
}
\end{document}

%% file: preamble.tex
\usepackage{overpic}
\usepackage{microtype}
\usepackage{graphicx} 
\usepackage{tikz}
\usepackage[T1]{fontenc}
\usepackage{amsmath}
\usepackage{pifont}
\makeatletter
\let\oldtheequation\theequation
\renewcommand\tagform@[1]{\maketag@@@{\ignorespaces#1\unskip\@@italiccorr}}
\renewcommand\theequation{(\oldtheequation)}
\makeatother

\usepackage[linesnumbered,ruled,vlined]{algorithm2e} 
\SetAlFnt{\small}
\SetAlCapFnt{\small}
\SetAlCapNameFnt{\small}
\SetAlCapHSkip{0pt}
\let\oldnl\nl
\newcommand{\nonl}{\renewcommand{\nl}{\let\nl\oldnl}}
\definecolor{cmtClr}{HTML}{028A0F}

\usepackage{booktabs}
\usepackage{multirow}
\usepackage{threeparttable}
\usepackage{caption}
\usepackage{siunitx}

\usepackage{nicefrac}
\usepackage{comment}

\usepackage[outline]{contour}
\contourlength{0.075em}

\newlength{\resLen}

\renewcommand{\paragraph}[1]{\vspace{.5em}\noindent\textbf{#1.}}

\newcommand{\sph}{\mathbb{S}^2}
\newcommand{\ex}{\mathbb{E}}
\newcommand{\pr}{\mathbb{P}}
\newcommand{\D}{\mathrm{d}}
\newcommand{\pluseq}{\mathrel{+}=}

\newcommand{\bmu}{\boldsymbol{\mu}}
\newcommand{\bSigma}{\boldsymbol{\Sigma}}
\newcommand{\Cest}{\langle C \rangle}
\newcommand{\loss}{\mathcal{L}}
\newcommand{\img}{\boldsymbol{I}}

\newcommand{\Mf}{M_\mathrm{f}}
\newcommand{\Mb}{M_\mathrm{b}}
\newcommand{\CestTot}{\Cest_\mathrm{tot}}

\newcommand{\brdf}{f_\mathrm{r}}
\newcommand{\vis}{T}
\newcommand{\Li}{L_\mathrm{in}}
\newcommand{\LiDir}{L_\mathrm{e}}

\newcommand{\bc}{\boldsymbol{c}}
\newcommand{\bx}{\boldsymbol{x}}
\newcommand{\bz}{\boldsymbol{z}}
\newcommand{\balpha}{\boldsymbol{\alpha}}
\newcommand{\bom}{\boldsymbol{\omega}}
\newcommand{\bomi}{\bom_\mathrm{in}}
\newcommand{\bomo}{\bom_\mathrm{out}}
\newcommand{\gShadow}{g^\mathrm{shadow}}
\newcommand{\alphaShadow}{\alpha^\mathrm{shadow}}

\newcommand{\dCdbc}{\partial_{\bc} C}
\newcommand{\dCdbcEst}{\langle \dCdbc \rangle}
\newcommand{\dCdbalpha}{\partial_{\balpha} C}
\newcommand{\dCdbalphaEst}{\langle \dCdbalpha \rangle}

\newcommand{\imgwithtext}[3]{%
\begin{tikzpicture}
    \node[inner sep=0pt] (img) {\includegraphics[width=#1]{#3}};
    \node[anchor=south west, font=\scriptsize\bfseries, text=white] at (img.south west) {#2};
\end{tikzpicture}%
}


%% file: sec/teaser.tex
\addtolength{\tabcolsep}{-4.5pt}
\centering
\small
\setlength{\resLen}{.245\textwidth}
\begin{tabular}{cccc}
	\begin{overpic}[width=\resLen]{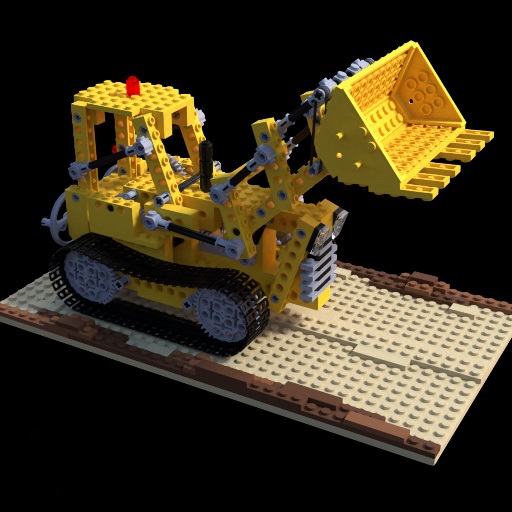}
		\put(2, 92){\color{white} \contour{black}{(a) \textbf{Standard}}}
		\put(0, 0){\includegraphics[trim={10 30 10 0},clip,width=.31\resLen]{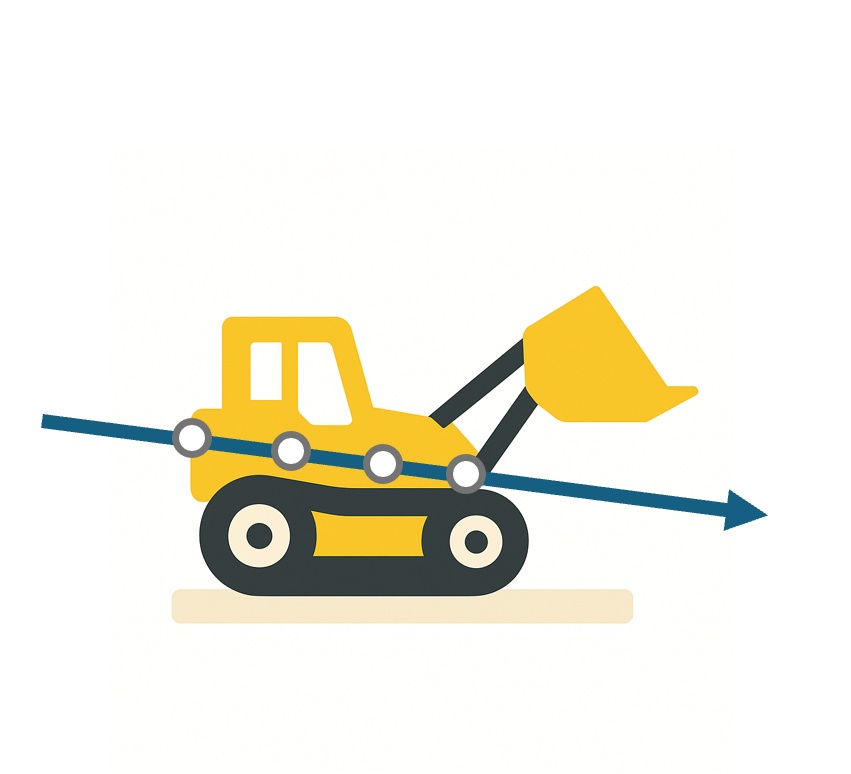}}
	\end{overpic}
	 &
	\begin{overpic}[width=\resLen]{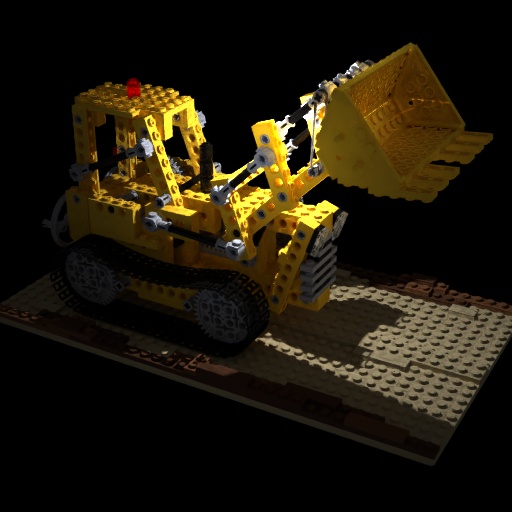}
		\put(2, 92){\color{white} \contour{black}{(b) \textbf{Relightable} (local point)}}
		\put(0, 0){\includegraphics[trim={10 30 10 0},clip,width=.31\resLen]{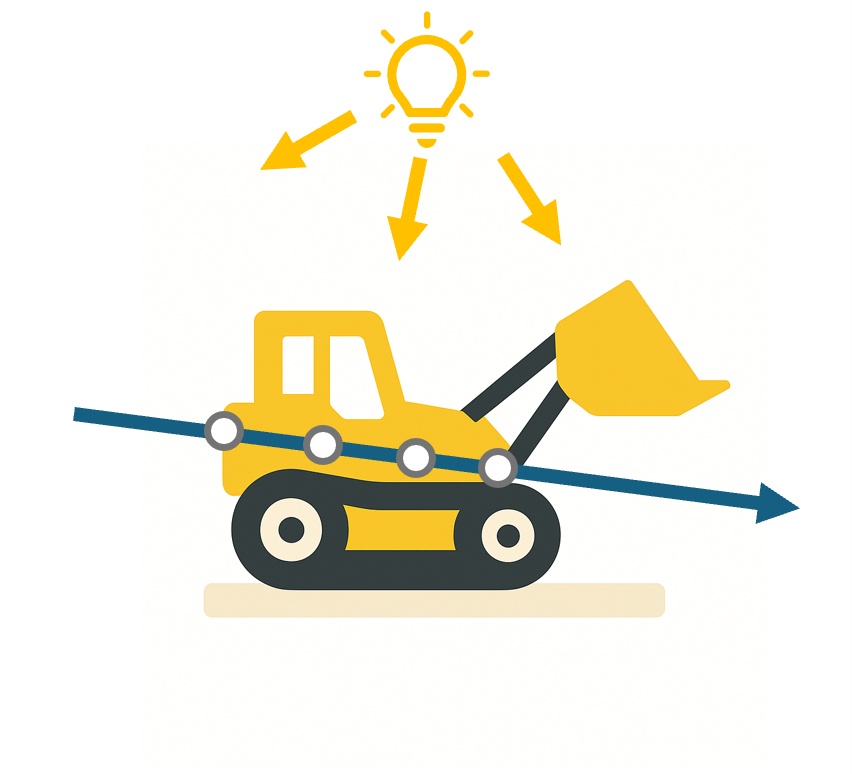}}
	\end{overpic}
	 &
	\begin{overpic}[width=\resLen]{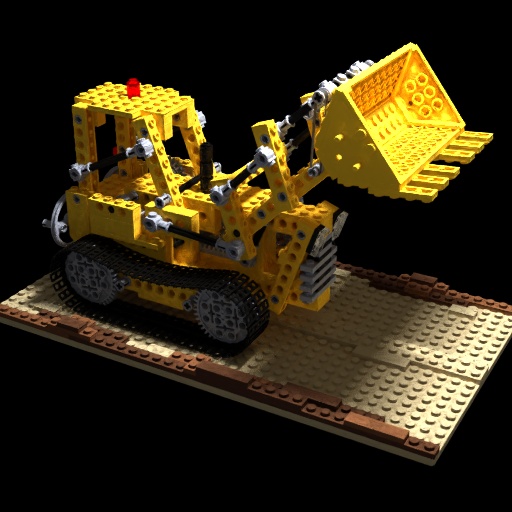}
		\put(2, 92){\color{white} \contour{black}{(c) \textbf{Relightable} (area)}}
		\put(0, 0){\includegraphics[trim={10 30 10 0},clip,width=.31\resLen]{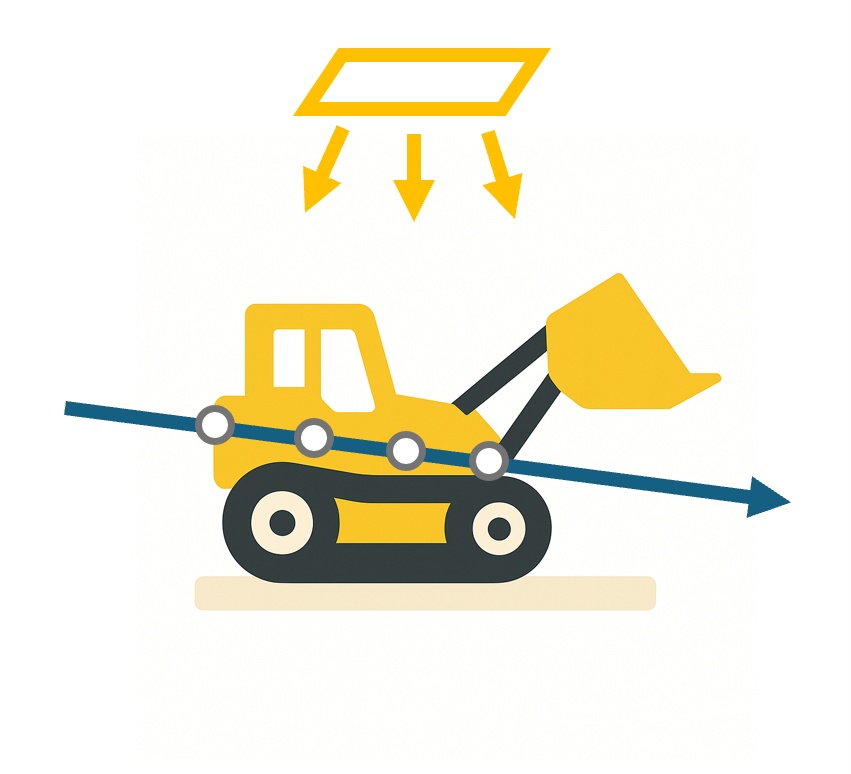}}
	\end{overpic}
	 &
	\begin{overpic}[width=\resLen]{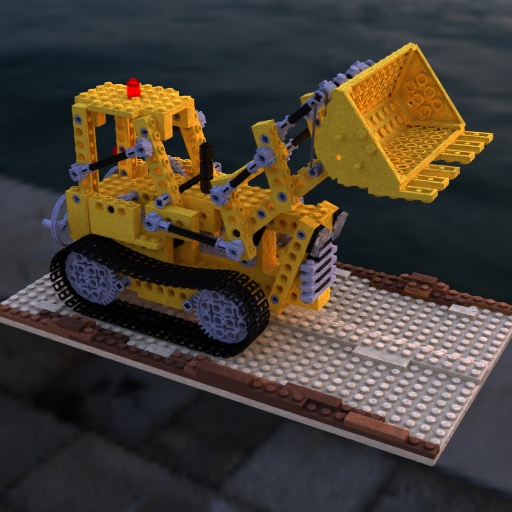}
		\put(2, 92){\color{white} \contour{black}{(d) \textbf{Relightable} (IBL)}}
		\put(0, 0){\includegraphics[trim={10 30 10 0},clip,width=.31\resLen]{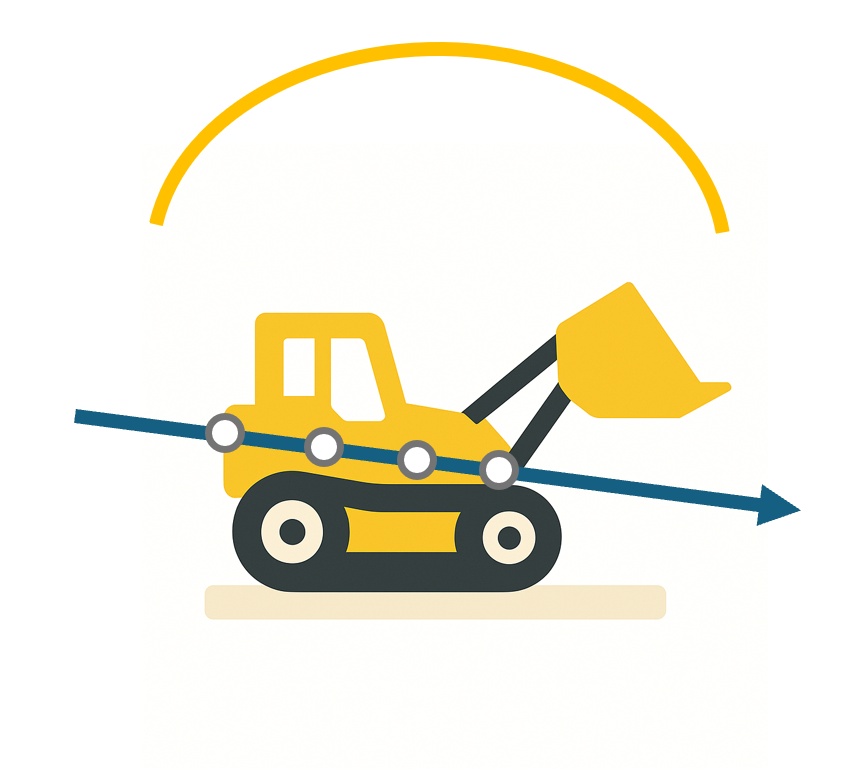}}
	\end{overpic}
\end{tabular}
\captionof{figure}{\label{fig:teaser}
	We introduce a differentiable stochastic formulation for ray-traced 3DGS, enabling efficient reconstruction and rendering of both \emph{standard} and \emph{relightable} 3DGS scenes.
	In this figure, we show re-renderings of our reconstructed standard 3DGS model (a) as well as relightable ones under local point light (b), area light (c), and image-based environmental illumination (d).
}

%% file: sec/abstract.tex
\begin{abstract}
	Ray-tracing-based 3D Gaussian splatting (3DGS) methods overcome the limitations of rasterization---rigid pinhole camera assumptions, inaccurate shadows, and lack of native reflection or refraction---but remain slower due to the cost of sorting all intersecting Gaussians along every ray.
	Moreover, existing ray-tracing methods still rely on rasterization-style approximations such as shadow mapping for relightable scenes, undermining the generality that ray tracing promises.

	We present a differentiable, sorting-free stochastic formulation for ray-traced 3DGS---the first framework that uses stochastic ray tracing to both reconstruct and render standard and relightable 3DGS scenes.
	At its core is an unbiased Monte Carlo estimator for pixel-color gradients that evaluates only a small sampled subset of Gaussians per ray, bypassing the need for sorting.
	For standard 3DGS, our method matches the reconstruction quality and speed of rasterization-based 3DGS while substantially outperforming sorting-based ray tracing.
	For relightable 3DGS, the same stochastic estimator drives per-Gaussian shading with fully ray-traced shadow rays, delivering notably higher reconstruction fidelity than prior work.
\end{abstract}

%% file: sec/intro.tex
\section{Introduction}
\label{sec:intro}
3D Gaussian Splatting (3DGS)~\cite{kerbl20233d} represents scenes as collections of translucent 3D Gaussians, enabling real-time rendering and rapid reconstruction.
Its combination of speed and expressiveness has made 3DGS a compelling choice for applications demanding both fidelity and interactivity---autonomous vehicles, immersive reality, and digital twins.

Most 3DGS methods rely on \emph{rasterization}: individual Gaussians are projected (``splatted'') onto the screen and alpha-blended to form the final image.
Hardware-accelerated splatting makes this fast, but rasterization brings inherent limitations---inaccurate shadows, lacking native support for reflection or refraction, and rigid pinhole camera assumptions.

Several ray-tracing-based 3DGS methods~\cite{loccoz20243dgrt, RaySplats, mai2024everexactvolumetricellipsoid} overcome these limitations by tracing camera rays through the scene and intersecting them with the Gaussians directly.
This unlocks shadows, reflection/refraction, and nonlinear camera models (e.g., fisheye lenses) without the workarounds that rasterization demands.

However, ray-traced 3DGS comes with its own costs.
Sorting all intersecting Gaussians along every camera ray is expensive, leaving these methods slower than their rasterization-based counterparts.
Moreover, when handling relightable scenes, existing ray-tracing methods still fall back on rasterization-style techniques---shadow mapping~\cite{williams1978casting}, deferred shading~\cite{deering1988triangle}---to approximate shadows and reflections, undermining the very generality that ray tracing promises.

\citet{Sun2025} recently showed that stochastic ray tracing can render 3DGS scenes efficiently without sorting.
Their algorithm, unfortunately, is not differentiable and therefore cannot be used for scene \emph{reconstruction}; nor does it address \emph{relightable} representations.

We present a \textbf{differentiable stochastic formulation} for ray-traced 3DGS---the first framework that uses stochastic ray tracing to both \emph{reconstruct} and \emph{render} standard and relightable 3DGS scenes.
At its core is an unbiased Monte Carlo estimator for pixel-color gradients that bypasses the need for sorting Gaussians and evaluates only a small sampled subset per ray.
This is particularly advantageous for relightable settings, where per-Gaussian shading (e.g., tracing shadow rays) makes exhaustive evaluation prohibitively expensive.
Because our formulation is inherently ray-based, it extends naturally to shadow rays and environmental illumination, yielding physically accurate shadows and reflections without specialized rasterization passes.

Concretely, our contributions are:
\begin{enumerate}
	\item An unbiased, sorting-free stochastic algorithm for estimating pixel-color gradients with respect to all Gaussian parameters, enabling efficient differentiable ray tracing of 3DGS scenes (\autoref{ssec:grad_est}).
	\item An extension of this formulation to relightable 3DGS, where the same stochastic estimator drives shadow-ray evaluation and environment-map integration, replacing the approximate shadow-mapping pipelines used by prior work (\autoref{ssec:extensions}).
\end{enumerate}

Across standard and relightable benchmarks (\autoref{sec:results}), our method matches the speed of rasterization-based 3DGS while substantially outperforming sorting-based ray tracing.
For relightable scenes, it delivers notably higher reconstruction fidelity, driven by per-Gaussian shading with accurate, ray-traced shadows and reflections.

%% file: sec/related.tex
\section{Related Work}

\paragraph{3DGS Foundations}
3D Gaussian Splatting (3DGS)~\cite{kerbl20233d} represents a scene as a set of anisotropic 3D Gaussians optimized through hardware-accelerated rasterization with differentiable compositing. This pipeline has become a standard for high-quality novel-view synthesis, and subsequent works extend it to eliminate visual artifacts~\cite{radl2024stopthepop, kv2025stochasticsplats} or support non-pinhole camera models~\cite{wu20253dgut}.

\paragraph{Ray Tracing for 3DGS}
While rasterization is efficient, its reliance on tile-based splatting and per-pixel depth sorting limits both correctness and flexibility. 3D Gaussian Ray Tracing (3DGRT)~\cite{loccoz20243dgrt} overcomes these constraints by incorporating hardware-accelerated ray tracing, enabling per-ray Gaussian sorting, arbitrary camera models, and seamless integration with mesh-based path tracing. Other works follow this direction, extending to non-Gaussian distributions~\cite{dontsplatyourgaussian} or proposing alternative implementations for improved efficiency or robustness~\cite{mai2024everexactvolumetricellipsoid, RaySplats}.

\paragraph{Efficient 3DGS Reconstruction}
A large body of work seeks to accelerate 3DGS reconstruction through model compression or more effective primitive management. Compression-oriented approaches~\cite{Fang2024MiniSplattingRS, HansonSpeedy, chen2025hac100xcompression3d} reduce the number of Gaussians via pruning, quantization, or structure-aware simplification. SpeedySplats~\cite{HansonSpeedy} further tightens Gaussian--tile localization to lower the per-tile primitive count. Complementary efforts improve densification and pruning heuristics for faster convergence without sacrificing quality~\cite{3DGSMCMC, Taming3DGS}. These techniques are largely orthogonal to ours: our method replaces the backward differentiation module and remains compatible with most existing acceleration strategies.

\paragraph{Stochastic Gaussian Splatting}
A growing thread of work replaces deterministic alpha blending with stochastic transparency~\cite{Enderton2010} to bypass the cost of sorting. 3DGRT introduced a biased variant that randomly retains a subset of intersected Gaussians with probability proportional to their blending weights. \citet{Sun2025} developed a principled unbiased algorithm and demonstrated its effectiveness for \emph{rendering} 3DGS assets, but did not provide a differentiable formulation suitable for \emph{reconstruction}. StochasticSplats~\cite{kv2025stochasticsplats} derived a differentiable stochastic blending algorithm to address popping artifacts; however, as we show in \autoref{sec:method} and the appendix, their gradient estimator suffers from high variance due to near-singular opacity terms, making it unsuitable for end-to-end 3DGS reconstruction.

\paragraph{Gaussian-Splatting-Based Relighting}
Several works extend 3DGS to support relighting by replacing view-dependent spherical harmonics with per-Gaussian material attributes~\cite{R3DG2023, liang2023gs, SVG-IR, gu2024IRGS}, handling shadows through baked visibility or learned approximations. More recent methods~\cite{bi2024rgs, RNG} model cast shadows via light-space splatting, achieving stronger performance on relighting benchmarks through more accurate appearance models and explicit light visibility estimation.

Our stochastic ray-tracing formulation offers a fundamentally different approach: the same per-ray visibility estimator used for primary rays extends directly to shadow rays and environmental illumination without specialized splatting passes or shadow-prediction networks, yielding physically accurate light transport within a unified framework.

%% file: sec/preliminaries.tex
\section{Preliminaries}
\label{sec:prelim}

Introduced by \citet{kerbl20233d}, 3DGS represents a scene using a collection of 3D Gaussian primitives $g_i$ with \emph{mean}~$\bmu_i$, \emph{covariance}~$\bSigma_i$, \emph{density}~$\sigma_i$, and \emph{color}~$c_i$.

Consider a camera ray that intersects $n$ Gaussians $g_1, g_2, \ldots, g_n$ in \emph{arbitrary (unsorted) order}.
The pixel color $C$ is obtained by alpha blending:
\begin{equation}
	\label{Eq:alpha_blend}
	C = \sum_{i=1}^{n} c_i \,\alpha_i \prod_{j \prec i} (1 - \alpha_j),
\end{equation}
where $j \prec i$ denotes Gaussians $g_j$ \emph{in front of} (i.e., with smaller depth than) $g_i$, and $\alpha_i$ is the \emph{opacity} of $g_i$:
\begin{equation}
	\label{Eq:particle_opacity}
	\alpha_i := \sigma_i \exp\left(-(\bx_i - \bmu_i)^{\top} \bSigma_i \,(\bx_i - \bmu_i) \right),
\end{equation}
with $\bx_i$ being the maximum-response point of $g_i$ along the camera ray~\cite{loccoz20243dgrt}.
Directly evaluating \autoref{Eq:alpha_blend}---\emph{deterministic blending}---requires sorting all $n$ Gaussians by depth, which remains expensive despite various acceleration efforts~\cite{HansonSpeedy, Fang2024MiniSplattingRS, chen2025hac100xcompression3d}, especially when $n$ is large.

\paragraph{Stochastic blending}
Following prior work~\cite{Enderton2010, Sun2025}, the pixel color $C$ can instead be estimated stochastically.
Drawing an index $I$ with probability mass $p_I$ and defining the random variable
\begin{equation}
	\label{Eq:forward_mc_0}
	\Cest = \frac{1}{p_I} \,c_I \,\alpha_I \prod_{j \prec I}(1 - \alpha_j),
\end{equation}
it is straightforward to verify that $\ex[\Cest] = C$, making $\Cest$ an \emph{unbiased Monte Carlo estimator}%
\footnote{Throughout this paper, we use $\langle h \rangle$ to denote a Monte Carlo estimator of $h$.}
of $C$.
Choosing the probability mass
\begin{equation}
	\label{Eq:mc_pdf}
	p_I = \alpha_I \prod_{j \prec I}(1 - \alpha_j)
\end{equation}
cancels most terms in \autoref{Eq:forward_mc_0}, yielding the simple estimator
\begin{equation}
	\label{Eq:forward_mc}
	\Cest = c_I.
\end{equation}

\citet{Sun2025} realized this idea with \autoref{alg:stoc_trans}.
This method examines Gaussians along the camera ray in arbitrary order, maintaining a running selection: for each Gaussian~$g_i$, it computes the opacity~$\alpha_i$ (\autoref{Eq:particle_opacity}) and depth~$z_i$, then replaces the current selection with $g_i$ (i.e., $I \gets i$) with probability~$\alpha_i$, provided $g_i$ is closer than the current choice (i.e., $z_i < z$; see \autoref{line:update_selection}).
After all Gaussians have been visited, the color~$c$ of the selected Gaussian~$g_I$ is computed (\autoref{line:compute_color}).
This process is repeated $\Mf$ times to reduce variance.

Compared with deterministic blending, \autoref{alg:stoc_trans} offers two key advantages: it requires \emph{no sorting}, since Gaussians can be visited in arbitrary order; and it evaluates the color of only \emph{one} selected Gaussian per sample---yielding significant savings when per-Gaussian shading is expensive, as we demonstrate in \autoref{sec:method}.

\begin{algorithm}[t]
	\caption{\label{alg:stoc_trans}
		Monte Carlo estimate of pixel color~$C$
	}
	\SetCommentSty{mycmtfn}
	\SetKwComment{tccinline}{// }{}
	%
	$\CestTot \gets 0$\;
	\For{$m = 1$ to $\Mf$}{
	\tcc{Draw index $I$}
	$I \gets 0$; $z \gets \infty$\;
	\ForEach{Gaussian $g_i$ along the camera ray}{
	Compute its opacity $\alpha_i$ and depth $z_i$\;
	Draw $\xi$ uniformly from $[0, 1)$\;
	\If{$\xi < \alpha_i$ and $z_i < z$}{ \label{line:alpha}
		$I \gets i$; $z \gets z_i$\; \label{line:update_selection}
	}
	}
	\tcc{Obtain Gaussian color}
	\If{$I > 0$}{
		Compute the color $c$ for the selected Gaussian $g_I$\; \label{line:compute_color}
	}
	\Else{
		$c \gets 0$\;
	}
	$\CestTot \pluseq c$ \tccinline*{\autoref{Eq:forward_mc}}
	}
	\Return $\nicefrac{\CestTot}{\Mf}$\;
\end{algorithm}

%% file: sec/ours.tex
\section{Stochastic Reconstruction of 3DGS Scenes}
\label{sec:method}
Reconstructing 3DGS scenes requires differentiating pixel colors $C$ defined in \autoref{Eq:alpha_blend} with respect to the Gaussian parameters.
Since the color derivatives $\nicefrac{\partial C}{\partial c_i}$ and opacity derivatives $\nicefrac{\partial C}{\partial \alpha_i}$ together determine all remaining parameter gradients (of $\bmu_i$, $\bSigma_i$, and $\sigma_i$) via the chain rule through \autoref{Eq:particle_opacity}, estimating these two quantities is the core computational task.

Although stochastic blending (\autoref{alg:stoc_trans}) enables efficient \emph{forward rendering} of 3DGS scenes, it alone is insufficient for \emph{reconstruction}.
The difficulty is that the opacity $\alpha_i$ governs the stochastic branching decision (\autoref{line:alpha}): na\"ive automatic differentiation (AD) treats this branch as fixed and therefore does not correctly differentiate through it, producing incorrect estimates of $\nicefrac{\partial C}{\partial \alpha_i}$.
While more sophisticated compiler techniques exist for differentiating through such discontinuities~\cite{Bangaru:2021}, their use within general-purpose GPU computation frameworks (e.g., CUDA or OptiX) remains limited in practice.

To address this challenge, we introduce a simple and efficient Monte Carlo procedure that produces \emph{unbiased} estimates of opacity gradients (\autoref{ssec:grad_est}).
We then extend this formulation with a technique for rendering and reconstructing \emph{relightable} 3DGS scenes (\autoref{ssec:extensions}).

\subsection{Stochastic Gradient Estimation}
\label{ssec:grad_est}
Let $\bc := (c_1, c_2, \ldots, c_n)$ and $\balpha := (\alpha_1, \alpha_2, \ldots, \alpha_n)$ denote the colors and opacities of all $n$ Gaussians along a camera ray.
Our goal is to estimate the gradient vectors $\dCdbc := (\nicefrac{\partial C}{\partial c_1}, \ldots, \nicefrac{\partial C}{\partial c_n})$ and $\dCdbalpha := (\nicefrac{\partial C}{\partial\alpha_1}, \ldots, \nicefrac{\partial C}{\partial\alpha_n})$.
Differentiating \autoref{Eq:alpha_blend} yields their components:
\begin{align}
	\frac{\partial C}{\partial c_i}
	 & = \alpha_i \prod_{j \prec i} (1 - \alpha_j),
	\\
	\frac{\partial C}{\partial\alpha_i}
	 & = \left[ \prod_{j \prec i}(1 - \alpha_j) \right] \!\!\left( c_i - \sum_{k \succ i} c_k \alpha_k \prod_{i \prec t \prec k}(1 - \alpha_t) \right),
\end{align}
where $k \succ i$ indicates Gaussians $g_k$ \emph{behind} (i.e., with greater depth than) $g_i$; and $i \prec t \prec k$ denotes Gaussians $g_t$ \emph{between} $g_i$ and $g_k$.

We now introduce Monte Carlo estimators $\dCdbcEst$ and $\dCdbalphaEst$ for the gradient vectors $\dCdbc$ and $\dCdbalpha$.
The key idea is to reuse the same stochastic sampling from the forward pass (\autoref{alg:stoc_trans}): we draw an index $I \in \{1, 2, \ldots, n\}$ with the probability mass $p_I$ from \autoref{Eq:mc_pdf}, which exactly cancels the product $\alpha_i \prod_{j \prec i}(1 - \alpha_j)$ appearing in both gradient expressions above.
We then set the $I$-th components $\dCdbcEst_I$ and $\dCdbalphaEst_I$ to
\begin{align}
	\label{Eq:ours_dCdbc}
	\dCdbcEst_I     & = 1,
	\\
	\label{Eq:ours_dCdbalpha_0}
	\dCdbalphaEst_I & = \frac{1}{\alpha_I} \left(c_I - \sum_{k \succ I} c_k \alpha_k \prod_{I \prec t \prec k}(1 - \alpha_t) \right),
\end{align}
while leaving all the other components at zero.

Intuitively, \autoref{Eq:ours_dCdbalpha_0} measures how much the color $c_I$ of the selected Gaussian differs from the alpha-blended color of all Gaussians behind it---capturing the effect of ``removing'' $g_I$ from the blend.
However, a brute-force evaluation of this sum would require the Gaussians~$g_k$ behind the sampled $g_I$ to be sorted.
To avoid this, we apply Monte Carlo a second time, drawing an index $K \succ I$ with probability mass
\begin{equation}
	p_{K|I} = \alpha_K \prod_{I \prec t \prec K}(1 - \alpha_t).
\end{equation}
Note that $p_{K|I}$ has the same form as \autoref{Eq:mc_pdf} but restricted to Gaussians behind $g_I$, so it can be sampled with the same stochastic procedure.
This reduces \autoref{Eq:ours_dCdbalpha_0} to a simple color difference:
\begin{equation}
	\label{Eq:ours_dCdbalpha}
	\dCdbalphaEst_I = \frac{1}{\alpha_I} (c_I - c_K).
\end{equation}
We prove the unbiasedness of our Monte Carlo estimators in the appendix.

\begin{figure}[t]
	\centering
	\includegraphics[width=.8\linewidth]{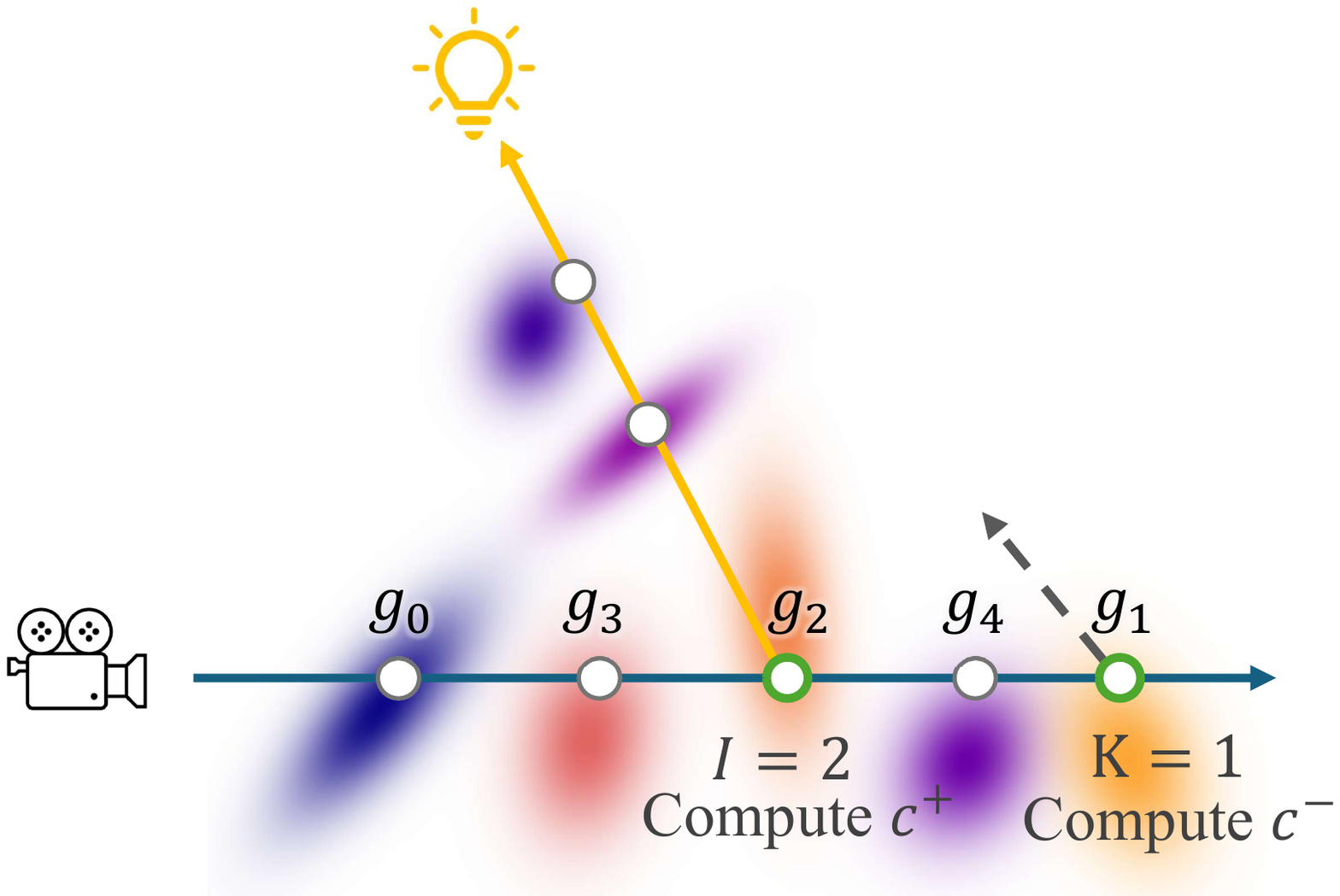}
	\vspace{-1.5em}
	\caption{\label{fig:backward_illust}
		Our stochastic gradient estimation (\autoref{alg:ours}) works by drawing a Gaussian $g_I$ along each camera ray (illustrated in dark blue) followed by another $g_K$ behind it.
		Then, the colors $c^+$ and $c^-$ of these Gaussians are computed for evaluating \autoref{Eq:ours_dCdbalpha}.
		For standard 3DGS, these colors can be obtained by evaluating the associated spherical harmonics (SH).
		For relightable 3DGS, on the contrary, we compute $c^+$ and $c^-$ using Monte Carlo by tracing additional shadow rays (shown in yellow) toward a light source.
		\vspace{-1.0em}
	}
\end{figure}

\begin{algorithm}[t]
	\caption{\label{alg:ours}
		Our Monte Carlo estimates of pixel color gradients $\dCdbc$ and $\dCdbalpha$.
	}
	\SetCommentSty{mycmtfn}
	\SetKwComment{tccinline}{// }{}
	$\dCdbcEst \gets \boldsymbol{0}$; $\dCdbalphaEst \gets \boldsymbol{0}$\;
	\For{$m = 1$ to $\Mb$}{
	\tcc{Draw index $I$}
	$I \gets 0$; $z \gets \infty$; $\alpha \gets 1$\; \label{line:draw_I_begin}
	\ForEach{Gaussian $g_i$ along the camera ray}{
	Obtain its opacity $\alpha_i$ and depth $z_i$\;
	Draw $\xi$ uniformly from $[0, 1)$\;
	\If{$\xi < \alpha_i$ and $z_i < z$}{
		$I \gets i$; $z \gets z_i$; $\alpha \gets \alpha_i$\; \label{line:draw_I_end}
	}
	}
	\If{$I > 0$}{
	\tcc{Draw index $K$}
	$K \gets 0$; $z \gets \infty$\; \label{line:draw_K_begin}
	\ForEach{Gaussian $g_k$ along the camera ray}{
	Obtain its opacity $\alpha_k$ and depth $z_k$\;
	Draw $\xi$ uniformly from $[0, 1)$\;
	\If{$z_k > z_I$ and $\xi < \alpha_k$ and $z_k < z$}{
		$K \gets k$; $z \gets z_k$\; \label{line:draw_K_end}
	}
	}
	\tcc{Obtain Gaussian colors}
	Compute the color $c^+$ of the Gaussian $g_I$\; \label{line:comp_clr_begin}
	\If{$K > 0$}{
		Compute the color $c^-$ of the Gaussian $g_K$\; \label{line:comp_clr_end}
	}
	\Else
	{
		$c^- \gets 0$\;
	}
	\tcc{Update the gradient estimates}
	$\dCdbcEst_I \pluseq 1$ \tccinline*{\autoref{Eq:ours_dCdbc}}
	$\dCdbalphaEst_I \pluseq \nicefrac{(c^+ - c^-)}{\alpha}$ \tccinline*{\autoref{Eq:ours_dCdbalpha}}
	}
	}
	\Return $\nicefrac{\dCdbcEst}{\Mb}$, $\nicefrac{\dCdbalphaEst}{\Mb}$\;
\end{algorithm}

\paragraph{Monte Carlo algorithm}
\autoref{alg:ours} summarizes the complete multi-sample procedure for estimating both $\dCdbc$ and $\dCdbalpha$.
Each of the $\Mb$ rounds draws a Gaussian~$g_I$ (\autoref{line:draw_I_begin}--\autoref{line:draw_I_end}) and a second Gaussian $g_K$ behind it (\autoref{line:draw_K_begin}--\autoref{line:draw_K_end}), as illustrated in \autoref{fig:backward_illust}.
The colors $c^+, c^-$ of the two selected Gaussians are then computed (\autoref{line:comp_clr_begin}--\autoref{line:comp_clr_end}), and the $I$-th components of the gradient estimates are updated using \autoref{Eq:ours_dCdbc} and \autoref{Eq:ours_dCdbalpha}.

Much like its forward-rendering counterpart (\autoref{alg:stoc_trans}), \autoref{alg:ours} requires no sorting and maintains only a minimal per-ray state---the selected indices $I$, $K$ and their depths---as the ray traverses the scene in the GPU ray-tracing pipeline.

	Furthermore, when we use the sorting-based forward-rendering algorithm, the sample indices $I$ and $K$ required for the backward pass can be collected at almost negligible cost in memory and time. In practice, we find that for simple reconstruction tasks, this approach further boosts the performance of our algorithm.

\paragraph{Reconstruction pipeline}
	Being able to stochastically compute gradients of pixel colors (\autoref{alg:ours}), we reconstruct 3DGS scenes (using multi-view input images) with a standard two-pass process as follows.

	In the \emph{forward pass}, we render images $\img$ (of a mini-batch) using the sorting-based algorithm introduced by~\citet{loccoz20243dgrt}. Meanwhile, we run our stochastic sampling algorithm (\autoref{alg:ours}) without the color computation to sample the indices $I$ and $K$.

In the \emph{backward pass}, we first compute the rendering loss $\loss$ by comparing rendered images $\img$ against the input views, along with the pixel-wise gradient $\nicefrac{\partial\loss}{\partial\img}$.
	We then apply \autoref{alg:ours} using the pre-computed indices to further backpropagate the gradients onto the color $c_i$ and opacity $\alpha_i$ of each Gaussian $g_i$ as well as its mean~$\bmu_i$, covariance~$\bSigma_i$, and density $\sigma_i$ via \autoref{Eq:particle_opacity}.

Given the resulting gradients with respect to all Gaussian parameters, we update them following the same optimization scheme as prior work \cite{kerbl20233d, loccoz20243dgrt}.

\subsection{Handling Relightable 3DGS Scenes}
\label{ssec:extensions}
%


%
The stochastic computation of pixel color gradients (\autoref{alg:ours}) can also significantly benefit the rendering and reconstruction of \emph{relightable} 3DGS scenes.
In the following, we first present a physically inspired relightable 3DGS formulation, and then discuss the rendering and reconstruction of 3D scenes under this formulation.


\paragraph{Relightable 3DGS formulation}
In standard 3DGS, each Gaussian $g_i$ carries a fixed color~$c_i$---it essentially \emph{emits} light.
In a relightable formulation, $c_i$ must instead depend on the incident illumination, since Gaussians effectively \emph{reflect} light.
Motivated by the rendering equation~\cite{Kajiya:RE}, we express the color~$c_i$ as a spherical integral over incoming directions:
\begin{equation}
	\label{Eq:RE}
	c_i(\bomo) = \int_{\sph} \brdf(\bz_i, \bomi, \bomo) \,\Li(\bomi) \,\D\bomi,
\end{equation}
where $\bomo$ and $\bomi$ are the viewing and incident lighting directions, respectively.
Here, $\brdf$ denotes the (cosine-weighted) bidirectional reflectance distribution function (BRDF), parameterized by a per-Gaussian feature $\bz_i$ that can encode either physical quantities (e.g., surface albedo, roughness) or learned latent vectors for neural BRDFs.

\input{sec/tab_benchmarks}

To instantiate \autoref{Eq:RE} in practice, we adopt a lightweight neural decoder $\Theta$, shared across all Gaussians, following RNG~\cite{RNG}:
\begin{equation}
	\label{Eq:neural_appearance}
	\Theta(\bz_i, \bomi, \bomo, \LiDir, \vis) \approx \brdf(\bz_i, \bomi, \bomo) \,\Li(\bomi),
\end{equation}
The decoder takes as input the incident and outgoing directions $\bomi$ and $\bomo$, together with the per-Gaussian latent feature~$\bz_i$.
To help the network learn baked-in global illumination effects such as interreflections, $\Theta$ receives two additional inputs: the direct emission $\LiDir$ from a light source in direction~$\bomi$, and the transmittance~$\vis$ between that light and the Gaussian $g_i$, defined as
\begin{equation}
	\label{Eq:T}
	\vis = 1 - \sum_{i' = 1}^{n'} \,\alphaShadow_{i'} \prod_{j' \prec i'} (1 - \alphaShadow_{j'}),
\end{equation}
where $\alphaShadow_1, \ldots, \alphaShadow_{n'}$ are the opacities of the unsorted Gaussians $\gShadow_1, \ldots, \gShadow_{n'}$ along the shadow ray, and $j' \prec i'$ denotes indices of Gaussians between $\gShadow_{i'}$ and $g_i$.
Intuitively, the product $\LiDir \,\vis$ represents the direct illumination arriving at the Gaussian $g_i$ from direction $\bomi$, attenuated by any occluding Gaussians along the way.

\paragraph{Rendering and training}
Due to the increased integrand dimensionality, the sorting-based forward rendering algorithm becomes unaffordable for relightable Gaussians. Therefore, we switch to stochastic computation for pixel colors using \autoref{alg:stoc_trans}.
In practice, for point or directional light sources, \autoref{Eq:RE} reduces to a summation over directions toward individual lights (see \autoref{fig:backward_illust}).
For environmental illumination, on the other hand, we estimate the integral in \autoref{Eq:RE} using Monte Carlo integration.

To evaluate $\vis$, we estimate the transmittance along the shadow ray using a modified version of \autoref{alg:stoc_trans}: after visiting all Gaussians $\gShadow_1, \ldots, \gShadow_{n'}$ along the shadow ray, we set the estimated transmittance to one if no Gaussian was selected (i.e., the sampled depth $z = \infty$, meaning the ray is unoccluded) and zero otherwise.
A detailed description of this procedure is provided in the appendix.

Training follows the same two-pass pipeline described in \autoref{ssec:grad_est}, with one modification: the per-Gaussian color is now produced by the neural decoder $\Theta$ rather than being a stored attribute.
In the backward pass, the color gradient $\dCdbcEst$ is backpropagated through $\Theta$ to update both the network weights and the per-Gaussian latent features~$\bz_i$, while the opacity gradient $\dCdbalphaEst$ is backpropagated to update the shape parameters of each Gaussian as before.

\paragraph{Discussion}
Existing relightable 3DGS methods approximate transmittance using shadow-mapping passes or costly shadow-prediction networks~\cite{RNG, bi2024rgs}.
In contrast, our stochastic ray tracing computes transmittance both efficiently and accurately, enabling faster and more faithful reconstructions---as demonstrated in \autoref{sec:results}.
Moreover, because our formulation is inherently ray-traced, it naturally supports complex environmental illumination, whereas shadow-mapping---based techniques require substantial---and often imperfect---extensions to handle such scenarios.

Additionally, our rendering algorithm is agnostic to the choice of reflectance model $\brdf$, and can therefore be combined with other per-Gaussian reflectance representations (e.g., GS$^3$~\cite{bi2024rgs}, Relightable-3DGS~\cite{R3DG2023}, GS-IR~\cite{liang2023gs}).

\paragraph{Relation to StochasticSplats}
	StochasticSplats~\cite{kv2025stochasticsplats} also estimates gradients of the alpha-blended pixel colors $C$ stochastically but operates within a rasterization framework rather than ray tracing.
	Their gradient estimator could serve as an alternative to ours; however, as we show through visual comparisons in \autoref{sec:results} and a detailed analysis in the appendix, our estimator yields superior reconstruction quality.

%% file: sec/tab_benchmarks.tex
\begin{table*}[t]
    \caption{\label{tab:benchmarks}
        \textbf{Novel view synthesis}: Results of our approach and baselines on standard novel view synthesis benchmarks. As a \textbf{ray tracing} algorithm, our method runs at a similar speed to the rasterization-based baseline and achieves comparable quality on all benchmarks.
    }
    \centering
    \small
    \begin{threeparttable}
        \footnotesize 
        \setlength{\tabcolsep}{4.2pt} 
        \begin{tabular}{
            l
            S[table-format=2.2] S[table-format=1.3] S[table-format=1.3] l
            S[table-format=2.2] S[table-format=1.3] S[table-format=1.3] l
            S[table-format=2.2] S[table-format=1.3] S[table-format=1.3] l
        }
        \toprule
        \multirow{2}{*}{Method\textbackslash Metric}
         & \multicolumn{4}{c}{\textbf{MipNeRF360}} 
         & \multicolumn{4}{c}{\textbf{Tanks \& Temples}}
         & \multicolumn{4}{c}{\textbf{Deep Blending}} \\
        \cmidrule(lr){2-5} \cmidrule(lr){6-9} \cmidrule(lr){10-13}
         & {PSNR$\uparrow$} & {SSIM$\uparrow$} & {LPIPS$\downarrow$} & {Time$\downarrow$}
         & {PSNR$\uparrow$} & {SSIM$\uparrow$} & {LPIPS$\downarrow$} & {Time$\downarrow$}
         & {PSNR$\uparrow$} & {SSIM$\uparrow$} & {LPIPS$\downarrow$} & {Time$\downarrow$} \\
        \midrule
        3DGS           & 28.69 & 0.867 & 0.224 & 24m & 23.14 & 0.853 & \text{-} & 14m & 29.41 & 0.903 & \text{-} & 20m \\
        3DGRT          & 28.40 & 0.862 & 0.233 & 69m & 22.95 & 0.838 & 0.221 & 41m & 29.69 & 0.904 & 0.318 & 54m \\
        \textbf{Ours}     & 28.31 & 0.857 & 0.254 & 33m & 22.57 & 0.830 & 0.221 & 20m & 29.87 & 0.906 & 0.324 & 25m \\
        \bottomrule
        \end{tabular}
    \end{threeparttable}
\end{table*}

%% file: sec/results.tex
\begin{table}[t]
	\caption{\label{tab:timings}
		Per-iteration timing breakdown on \textsl{MipNeRF360} \cite{barron2022mip}.
		Our method significantly accelerates the ray tracing baseline.
	}
	\centering
	\small
	\begin{tabular}{c|c c}
		\toprule
		\textbf{Time} (ms) & \textbf{Total} & \textbf{Backward} \\
		\midrule
		3DGS               & \textbf{31.4}  & 20.6              \\
		3DGRT              & 87.5           & 50.5              \\
		\textbf{Ours}      & 39.8           & \textbf{17.9}     \\
		\bottomrule
	\end{tabular}
\end{table}

\section{Results}
\label{sec:results}
We evaluate our method against state-of-the-art baselines on standard benchmarks for both novel view synthesis and relightable reconstruction.

Our implementation builds on the 3DGRT codebase~\cite{loccoz20243dgrt}, using OptiX~\cite{optix} for hardware-accelerated ray tracing.
Sampling is performed in an OptiX \emph{any-hit} program, and the forward and backward passes run inside \emph{ray-gen} programs. Neural network evaluation and backpropagation use OptiX's Cooperative Vector interface~\cite{optix_coopvec}. All experiments are conducted on an Nvidia RTX 5880 Ada Generation GPU.

We set $\Mb=8$ backward samples for all experiments.
Since our stochastic algorithm tends to produce more Gaussians, we increase the densification interval to 400 in the novel view synthesis experiments to ensure a fair comparison.
Multiple forward samples are drawn via independent trials within a single BVH traversal for maximum efficiency; for relighting, we use $\Mf=15$.

By replacing sorted alpha blending with our stochastic estimator, we achieve substantially faster optimization with comparable visual quality on novel view synthesis benchmarks. Our method also naturally extends to relightable Gaussians, where accurate light transport estimation yields clear improvements over baselines.

\subsection{Novel View Synthesis}
\paragraph{Baselines}
Many recent works accelerate 3DGS optimization while preserving quality. Since our method only replaces the backward differentiation module, these techniques are largely orthogonal; we therefore compare against vanilla 3DGS and 3DGRT baselines.

For fair evaluation, all reconstructed scenes are rendered with 3DGRT's deterministic alpha blending.

\paragraph{Datasets}
We evaluate on four standard datasets: \textsl{MipNeRF-360} \cite{barron2022mip}, \textsl{Tanks \& Temples} \cite{Knapitsch2017}, \textsl{Deep Blending} \cite{DeepBlending2018}, and \textsl{NeRF Synthetic} \cite{mildenhall2020nerf}.

From \textsl{MipNeRF-360}, we select four indoor scenes (\textsl{room, counter, kitchen, bonsai}) and three outdoor scenes (\textsl{bicycle, garden, stump}).
From \textsl{Tanks \& Temples} we use \textsl{train} and \textsl{truck}; from \textsl{Deep Blending}, \textsl{playroom} and \textsl{drjohnson}.

\paragraph{Results}
Quantitative results are shown in \autoref{tab:benchmarks}. Our method delivers a substantial speed-up over 3DGRT while matching the speed of rasterization-based 3DGS, and achieves comparable reconstruction quality across all benchmarks. We test several forward sample counts in \autoref{tab:benchmarks} and adopt 30\,spp for the remaining experiments as the best speed--quality trade-off.

\begin{figure}[t]
	\centering
	\includegraphics[width=0.8\linewidth]{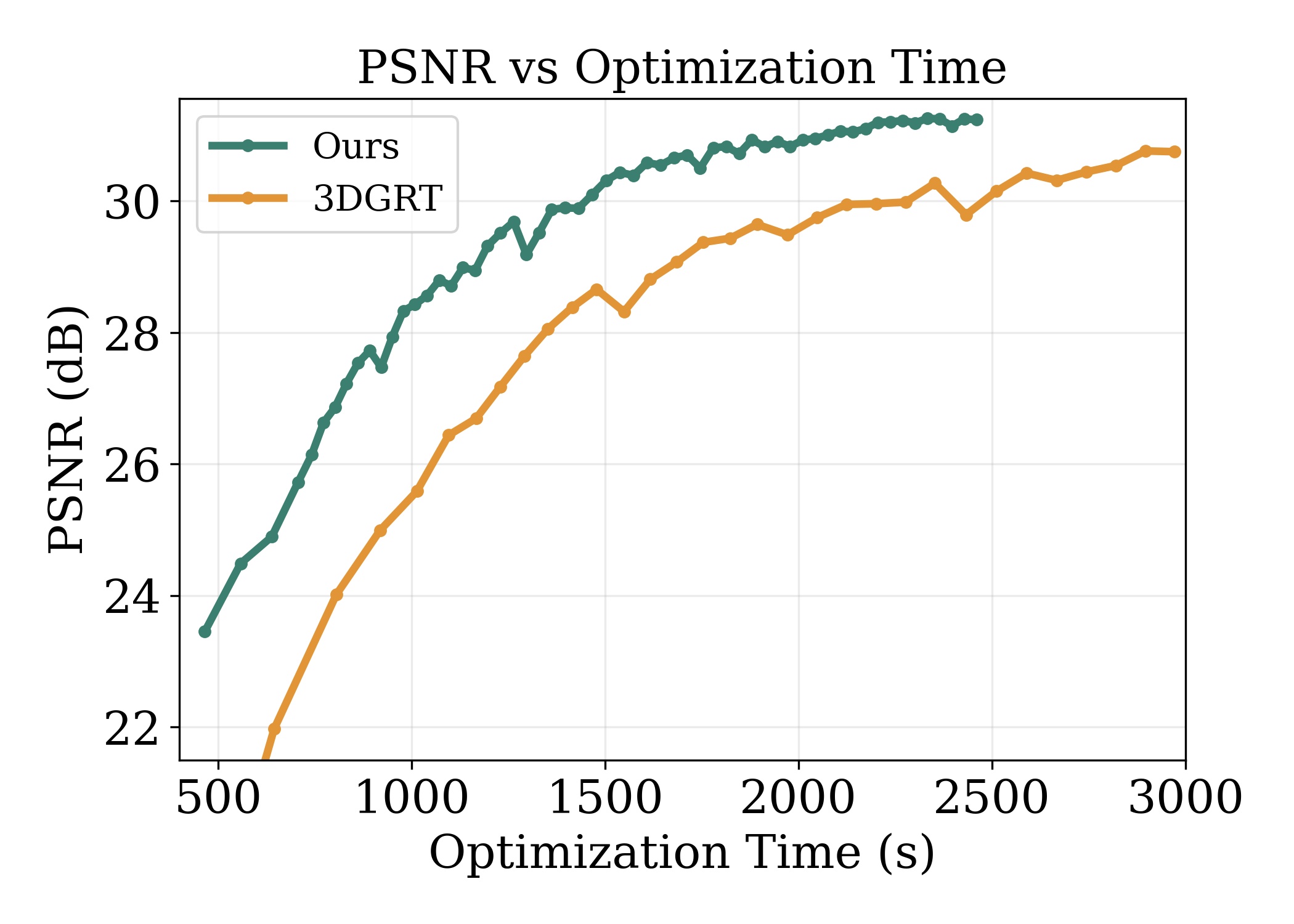}
	\caption{\label{fig:psnr_vs_time}
		PSNR vs.\ optimization time for sorted alpha blending (3DGRT) and our method.
	}
\end{figure}

\autoref{fig:nvs_figures} presents an equal-time visual comparison: all methods run for the same wall-clock budget. \autoref{fig:psnr_vs_time} plots PSNR against optimization time, confirming that our method consistently converges faster than the sorted ray tracing baseline.

A per-stage timing breakdown appears in \autoref{tab:timings}.
Relative to 3DGS, our main overhead is BVH construction; relative to 3DGRT, our stochastic backward pass cuts iteration time by more than half.


\input{sec/ssplats_figure}
	\paragraph{StochasticSplats comparison}
	We re-implement StochasticSplats in our ray tracing backend, denoting this variant \textsl{SS-Tracing}.
	\autoref{fig:benchmark_ssplats} compares novel-view synthesis results on the same datasets; quantitative metrics are reported in the appendix.

\subsection{Relightable Gaussian Splatting}

\begin{table}[t]
	\caption{\label{tab:relightable_benchmarks}
		Results of our method and baselines on the \textsl{NRHints} dataset \cite{zeng2023nrhints}.
	}
	\centering
	\small
	\begin{tabular}{c|cccc}
		\toprule
		\footnotesize{PSNR$\uparrow$\;$|$\;SSIM$\uparrow$} & \textbf{Ours}                   & RNG                     & GS$^3$         \\
		\midrule
		\textsl{Lego}                                      & \textbf{30.40}$|$\textbf{0.949} & 26.72$|$0.924           & 26.62$|$0.923  \\
		\textsl{Basket}                                    & \textbf{28.02}$|$\textbf{0.956} & 19.97$|$0.853           & 23.22$|$0.936  \\
		\textsl{Pixiu}                                     & \textbf{30.89}$|$0.936          & 30.35$|$\textbf{0.941}  & 30.38$|$0.937  \\
		\textsl{Hotdog}                                    & \textbf{31.88}$|$0.955          & 30.38$|$\textbf{0.960}  & 25.40$|$ 0.949 \\
		\textsl{FurBall}                                   & \textbf{33.69}$|$\textbf{0.949} & 27.82$|$0.926           & 26.36$|$0.931  \\
		\textsl{Cat}                                       & \textbf{28.42}$|$0.870          & 28.39 $|$\textbf{0.888} & 26.09 $|$0.882 \\
		\bottomrule
	\end{tabular}
\end{table}

We compare against state-of-the-art relightable baselines RNG and GS$^3$.

Following RNG, we adopt a two-stage pipeline. Stage~1 optimizes the Gaussians as standard 3DGS with view-dependent spherical harmonics for 15,000 iterations to initialize geometry. Stage~2 switches to the neural appearance model of \autoref{Eq:neural_appearance} and fine-tunes for 85,000 iterations. The network has 4 hidden layers of dimension 64, and each Gaussian stores a 16-dimensional latent feature. Stage~1 hyperparameters match 3DGRT's novel view synthesis configuration; in Stage~2, densification, pruning, and density reset intervals are set to 3,000, 1,000, and 12,000 respectively, with all other settings unchanged.

We evaluate on the NRHints dataset~\cite{zeng2023nrhints}, consistent with the baselines. \autoref{tab:relightable_benchmarks} reports quantitative metrics; \autoref{fig:relightable_figures} visualizes reconstruction quality.

Because our pipeline traces exact shadow rays, it produces high-quality shadows and geometry without dedicated shadow-handling modules. As shown in \autoref{fig:relightable_figures}, this is consistently preferable to neural approximations.

\autoref{fig:environment_map_relighting} shows our reconstructions relit under novel environment maps. Despite training only with point-light illumination, the results are visually plausible---and our full ray-tracing pipeline incurs no additional overhead for environment lighting.

%% file: sec/ssplats_figure.tex
\begin{figure}[]
    \setlength{\resLen}{1.65in}
    \centering
    \small
    \addtolength{\tabcolsep}{-5.5pt}
    \begin{tabular}{cc}
        (a) \textbf{Ours} & (b) \textbf{SS-Tracing}
        \\ \vspace{-0.1em}
        \includegraphics[width=\resLen]{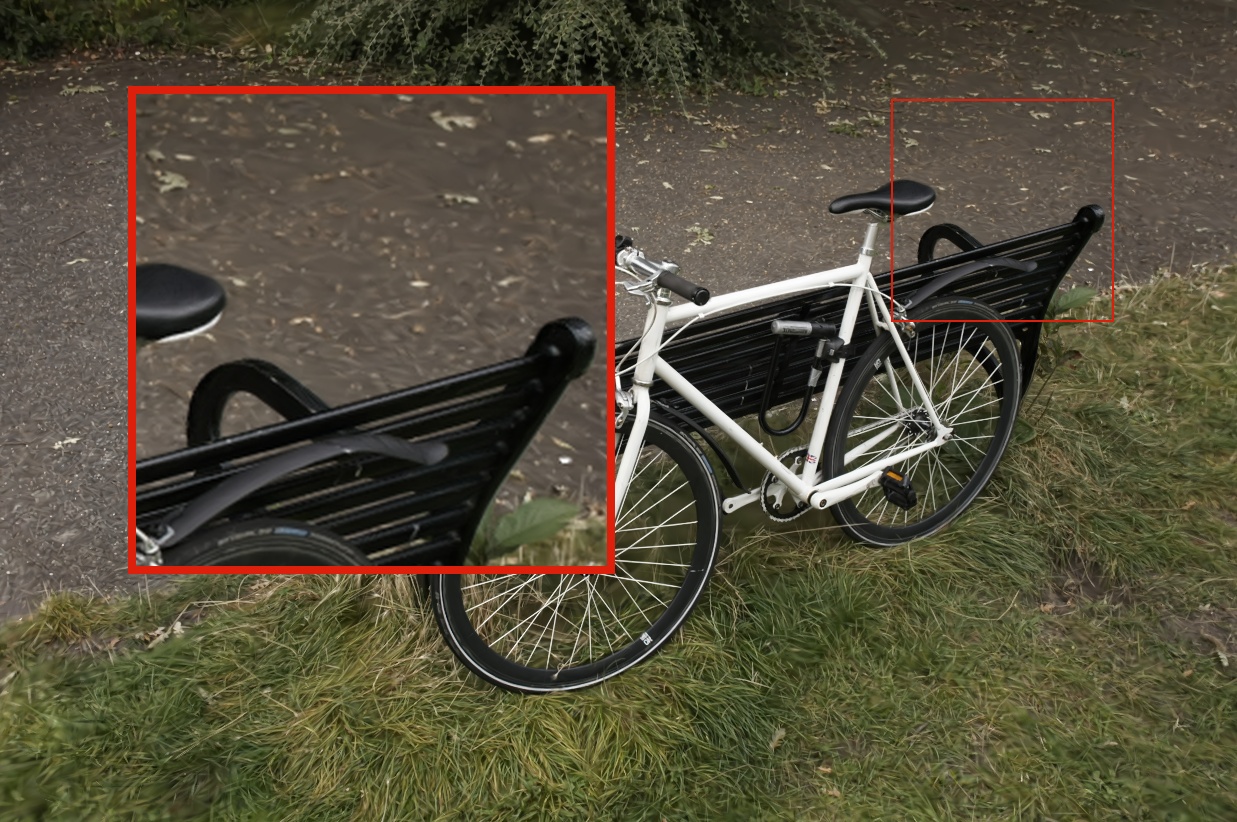} &
        \includegraphics[width=\resLen]{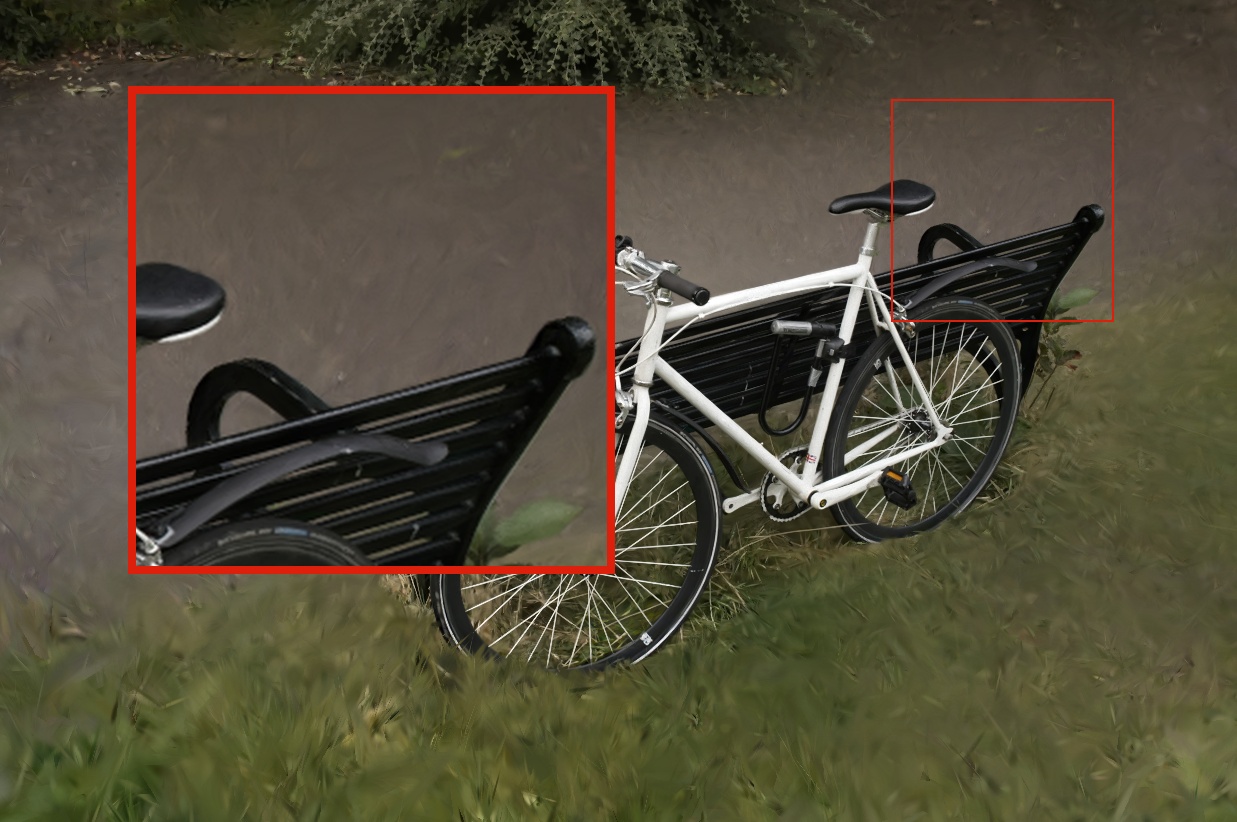} \\
        \includegraphics[width=\resLen]{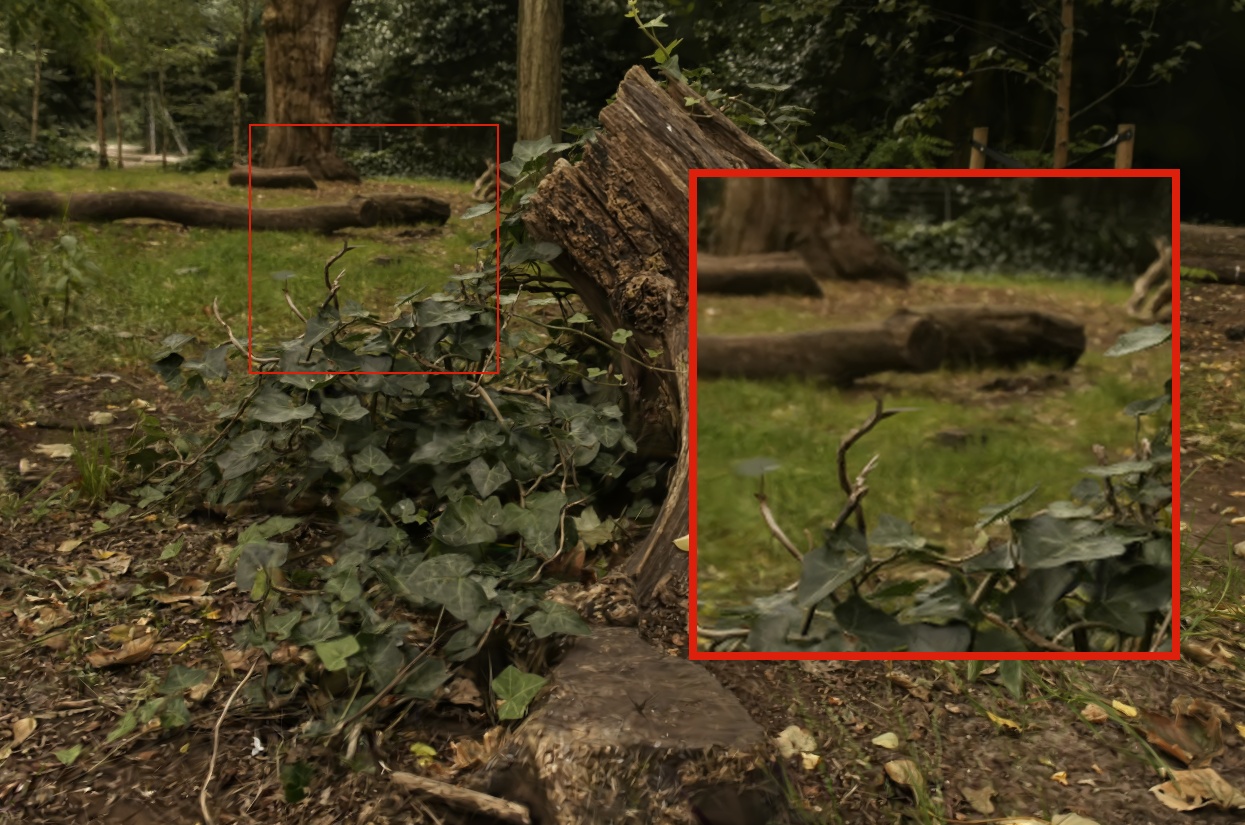} &
        \includegraphics[width=\resLen]{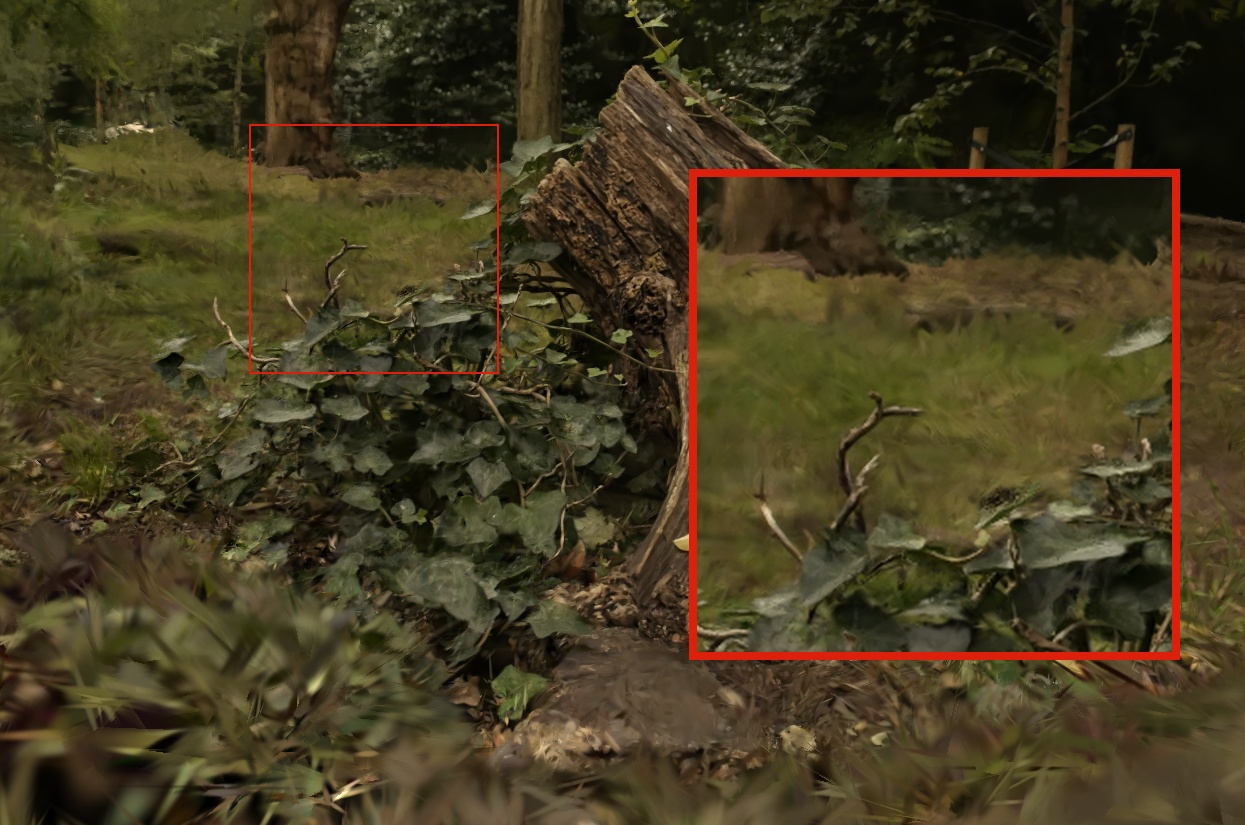}
    \end{tabular}
    \caption{
        Reconstruction quality comparison between our method and StochasticSplats~\cite{kv2025stochasticsplats}.
    }
    \label{fig:benchmark_ssplats}
\end{figure}

%% file: sec/conculsion.tex
\section{Discussion and Conclusion}
\label{sec:conclusion}
\paragraph{Limitations and future work}
Our stochastic formulation introduces variance into the gradient estimates, which we observed affects the behavior of the splitting and pruning heuristics used for Gaussian densification. A more detailed investigation of how the densification scheme interacts with this variance is left to future work.

\paragraph{Conclusion}
We presented a differentiable, sorting-free stochastic ray tracing framework for 3D Gaussian Splatting---the first to use stochastic ray tracing for both reconstruction and rendering of standard and relightable scenes.
By replacing exhaustive sorted evaluation with an unbiased Monte Carlo estimator that samples only a small subset of Gaussians per ray, our method matches the speed and quality of rasterization-based 3DGS while substantially outperforming sorting-based ray tracing.
For relightable scenes, the same estimator enables per-Gaussian shading with fully ray-traced shadow rays, delivering notably higher reconstruction fidelity than prior approaches that rely on rasterization-style approximations.

\section*{Acknowledgments}
We thank the anonymous reviewers for their feedback and suggestions.
This work was partially supported by NSF grant 2553564. This work started when Peiyu Xu was an intern at Adobe Research.

%% file: sec/result_figures.tex
\begin{figure*}[t]
    \setlength{\resLen}{1.75in}
    \centering
    \small
    \setlength{\tabcolsep}{1pt}
    \renewcommand{\arraystretch}{0}
    \begin{tabular}{cccc}
        \textbf{Reference} & (a) \textbf{Ours} & (b) \textbf{3DGS} & (c) \textbf{3DGRT}
        \\[3pt]
        \includegraphics[width=\resLen]{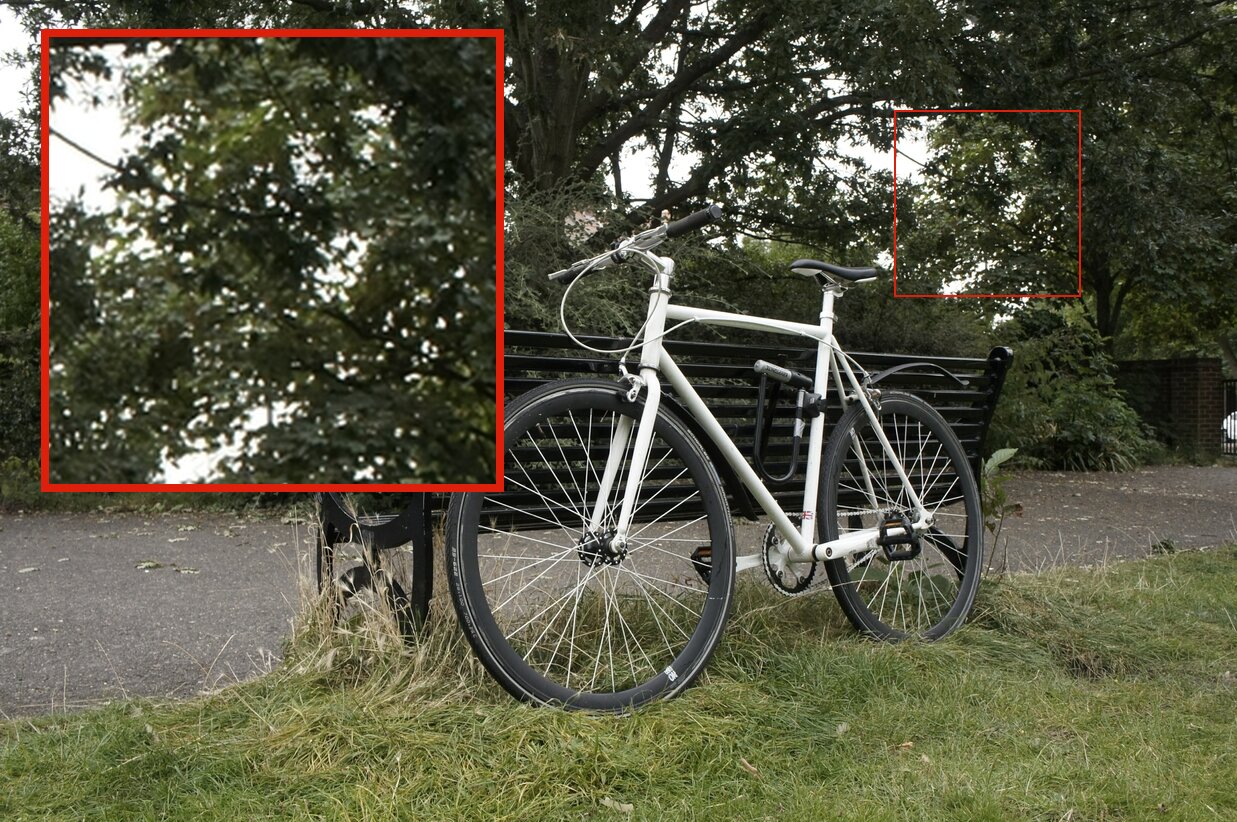} &
        \imgwithtext{\resLen}{\contour{black}{PSNR 19.0 / SSIM 0.541}}{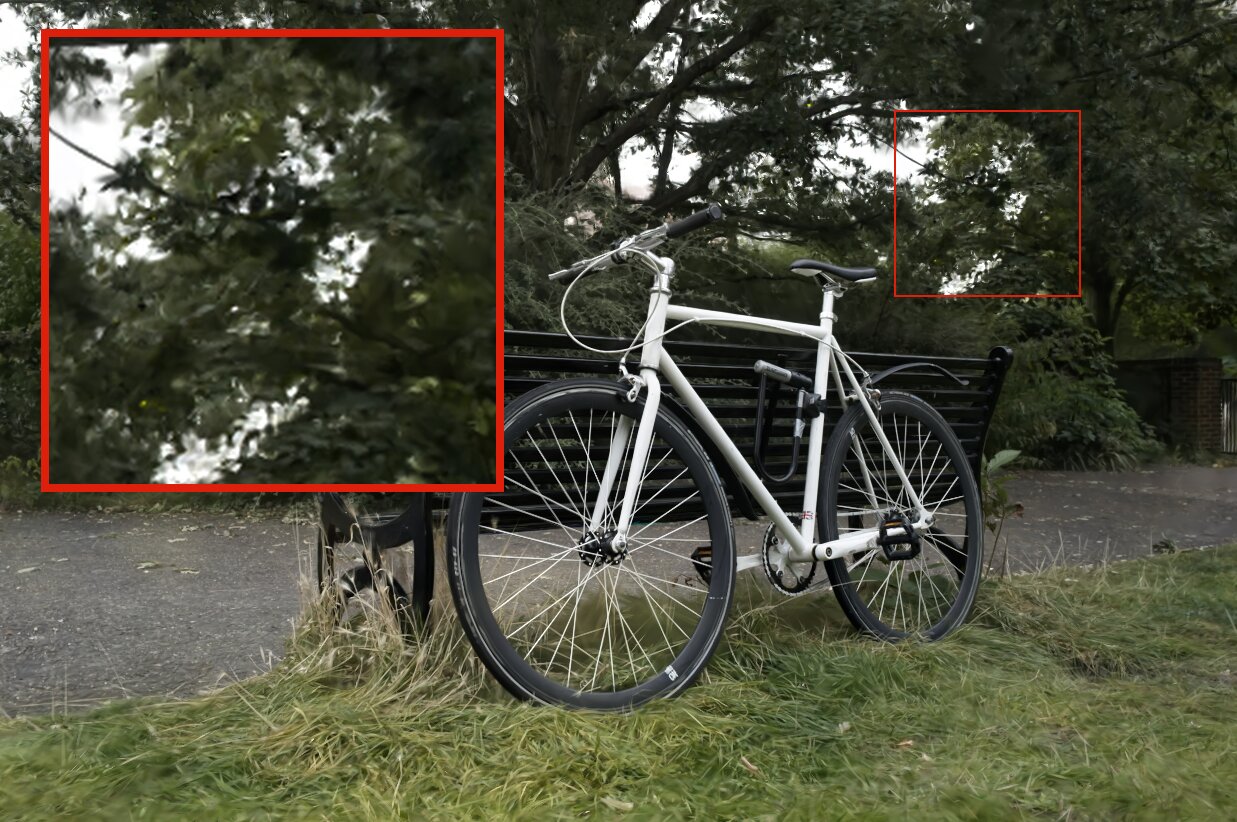} &
        \imgwithtext{\resLen}{\contour{black}{PSNR 19.0 / SSIM 0.544}}{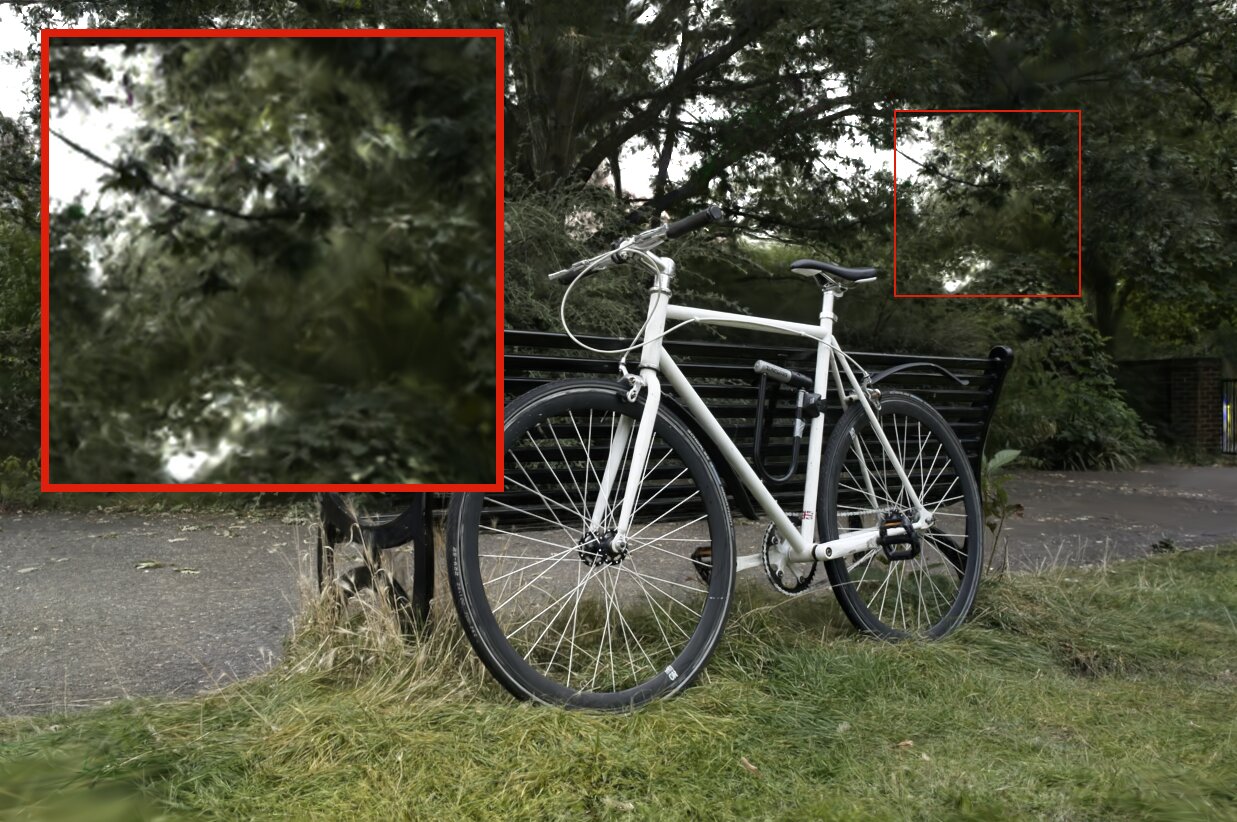} &
        \imgwithtext{\resLen}{\contour{black}{PSNR 18.2 / SSIM 0.529}}{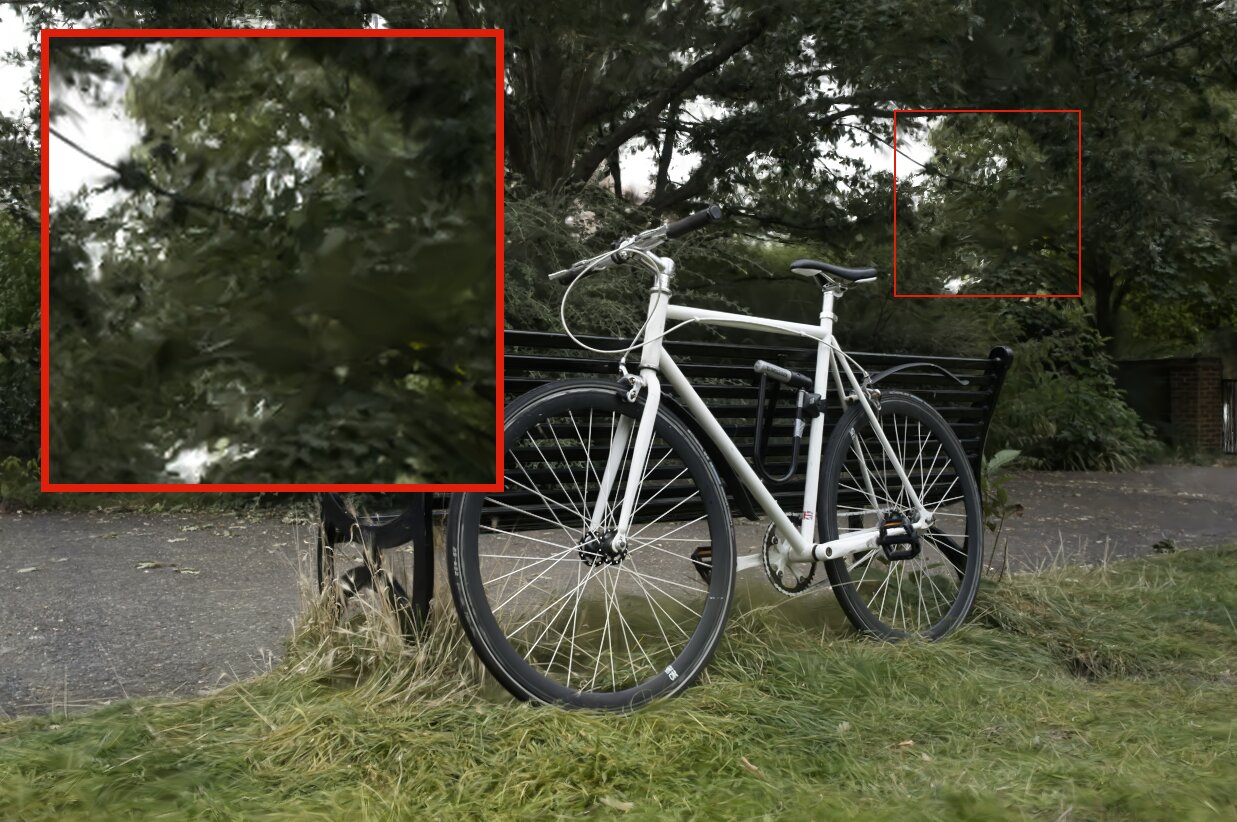} 
        \\[3pt]
        \includegraphics[width=\resLen]{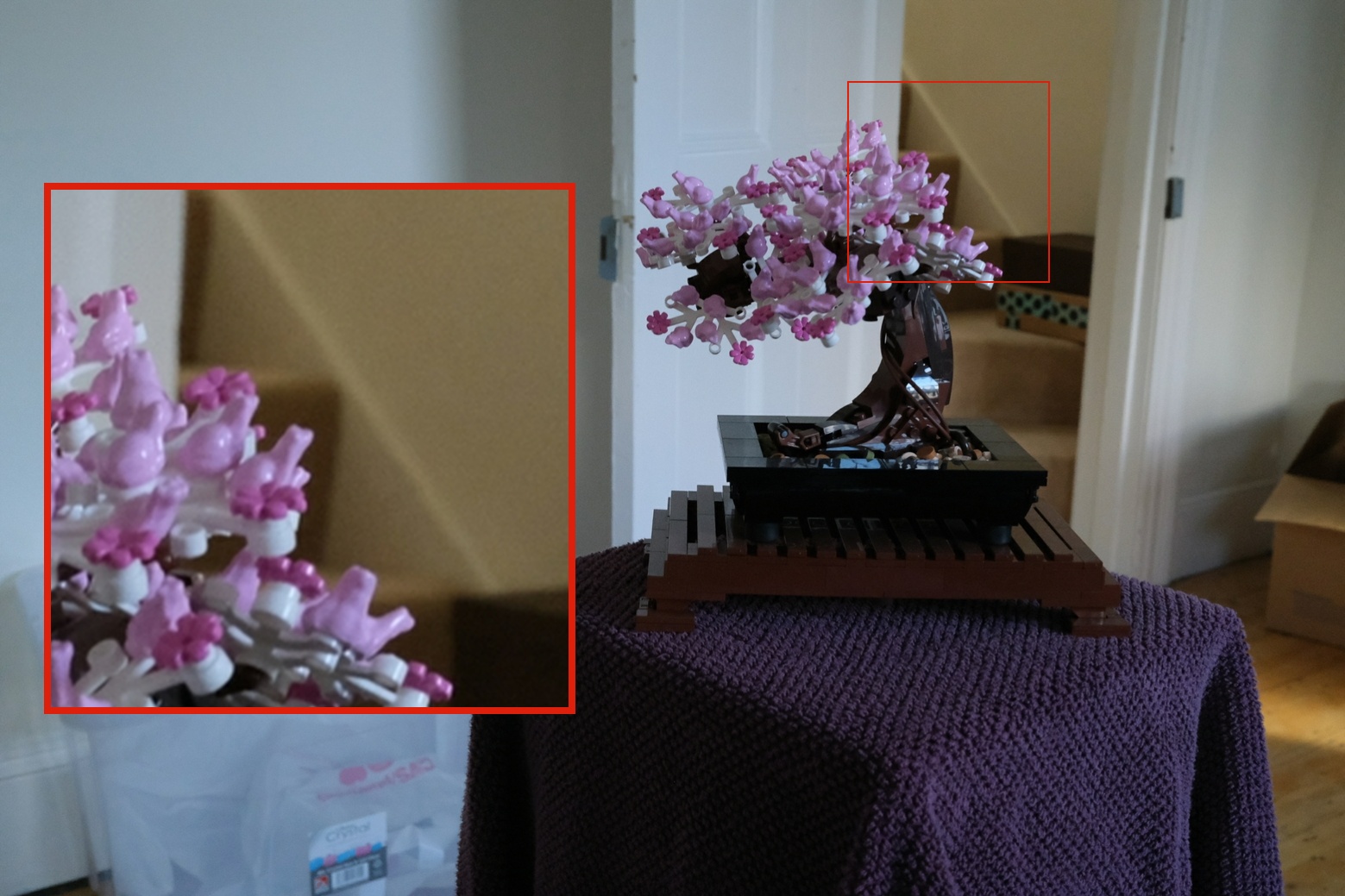} &
        \imgwithtext{\resLen}{\contour{black}{PSNR 30.5 / SSIM 0.931}}{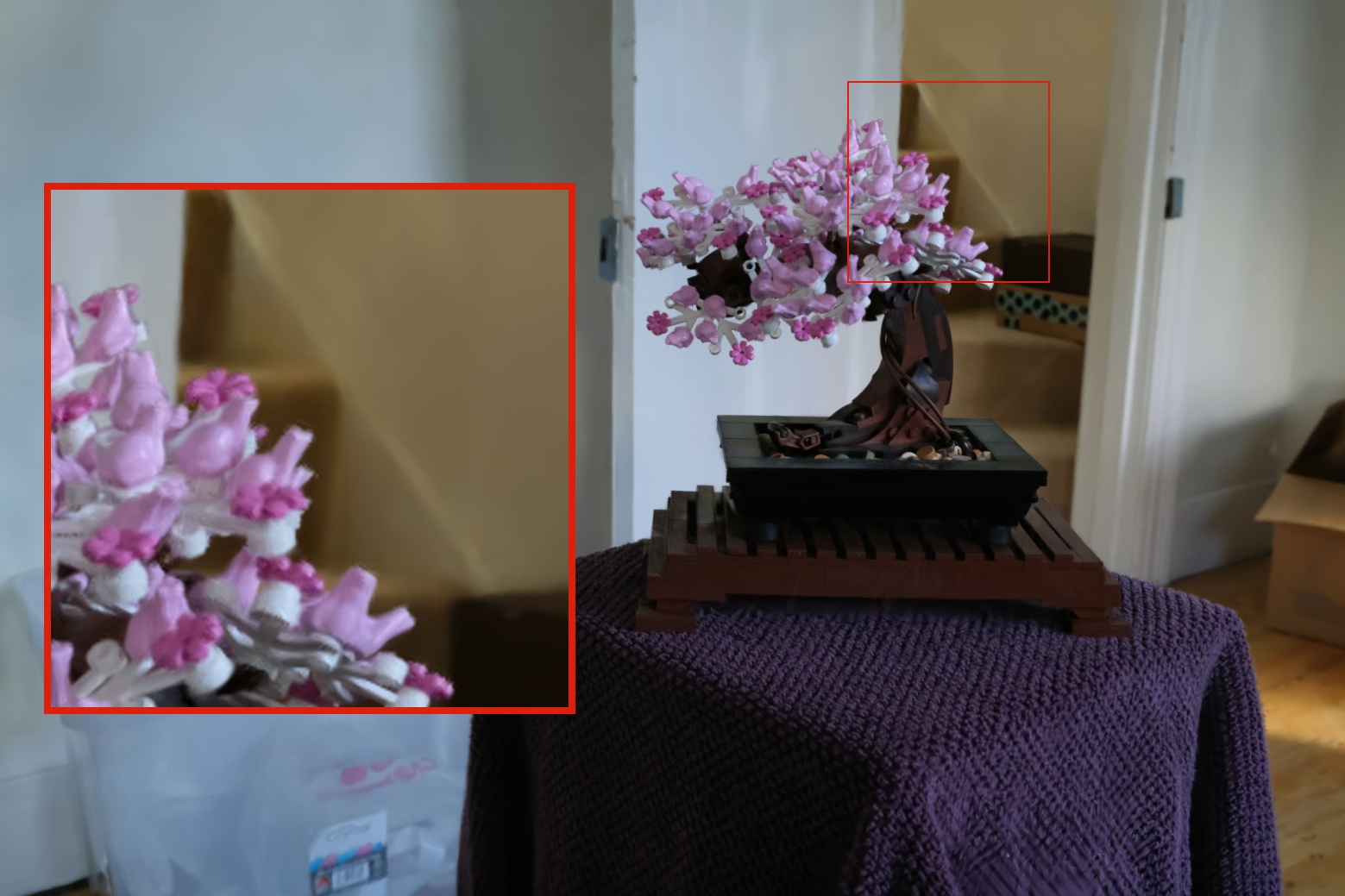} &
        \imgwithtext{\resLen}{\contour{black}{PSNR 31.5 / SSIM 0.942}}{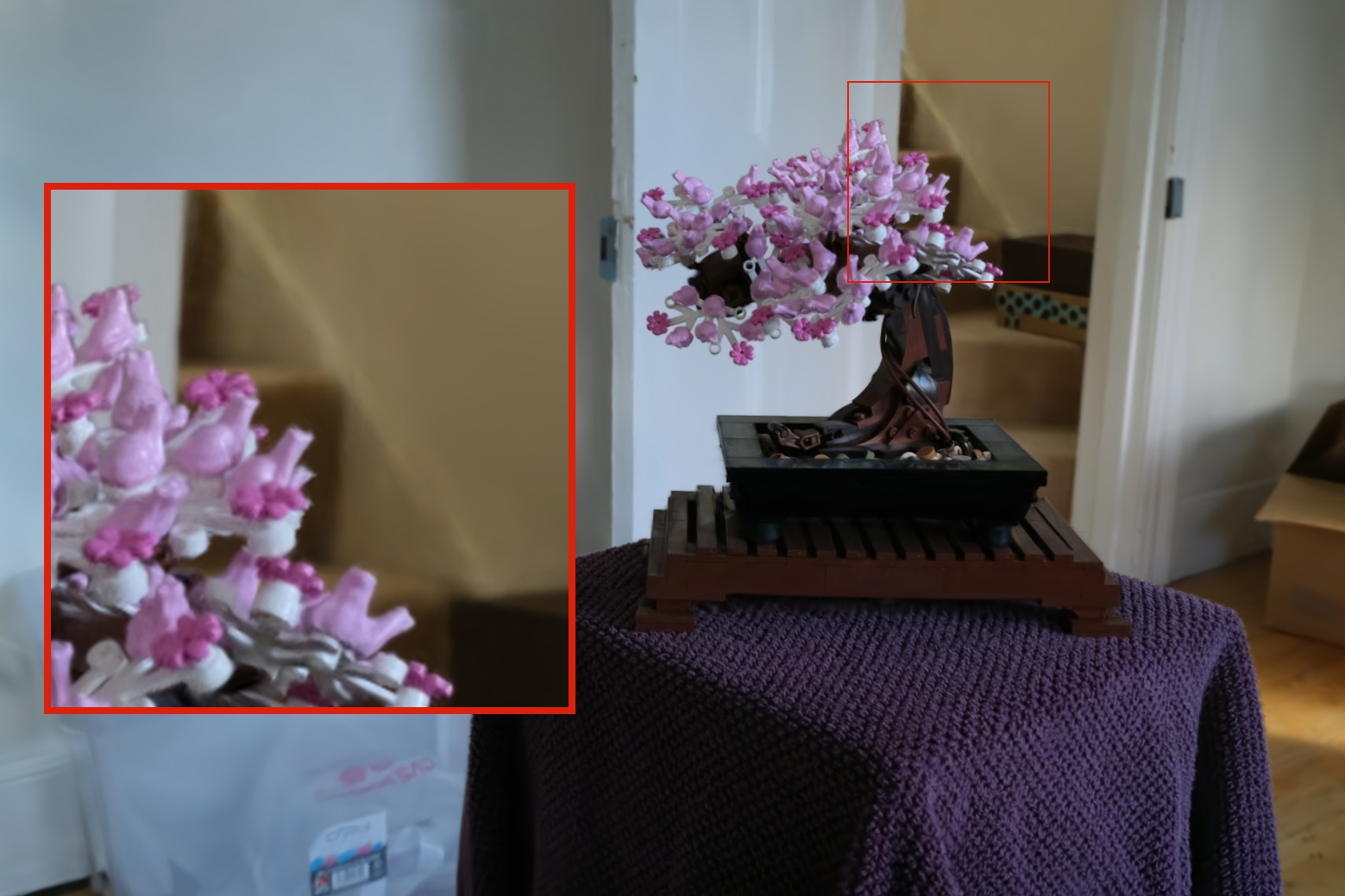} &
        \imgwithtext{\resLen}{\contour{black}{PSNR 29.6 / SSIM 0.925}}{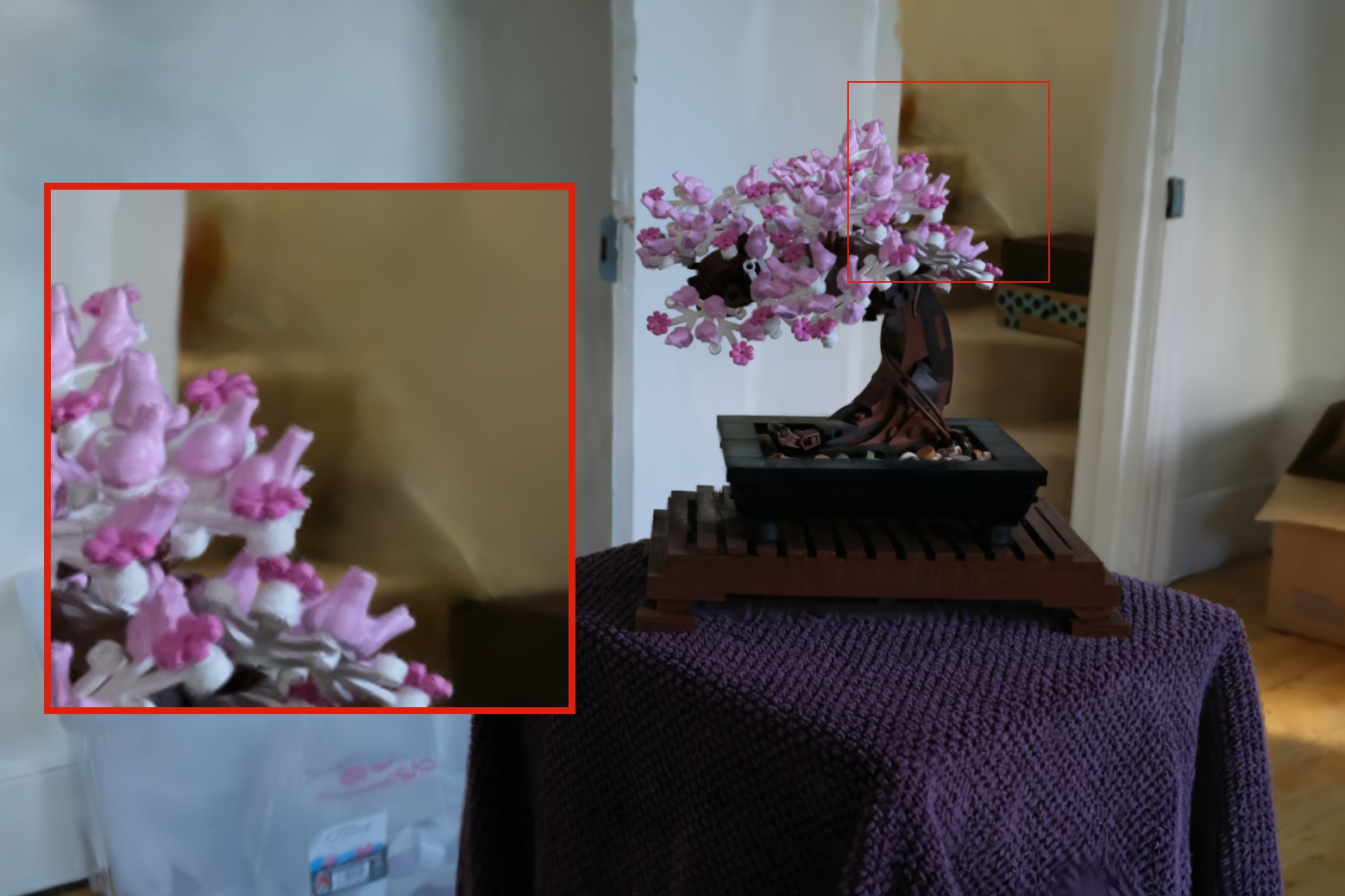} 
        \\[3pt]
        \includegraphics[width=\resLen]{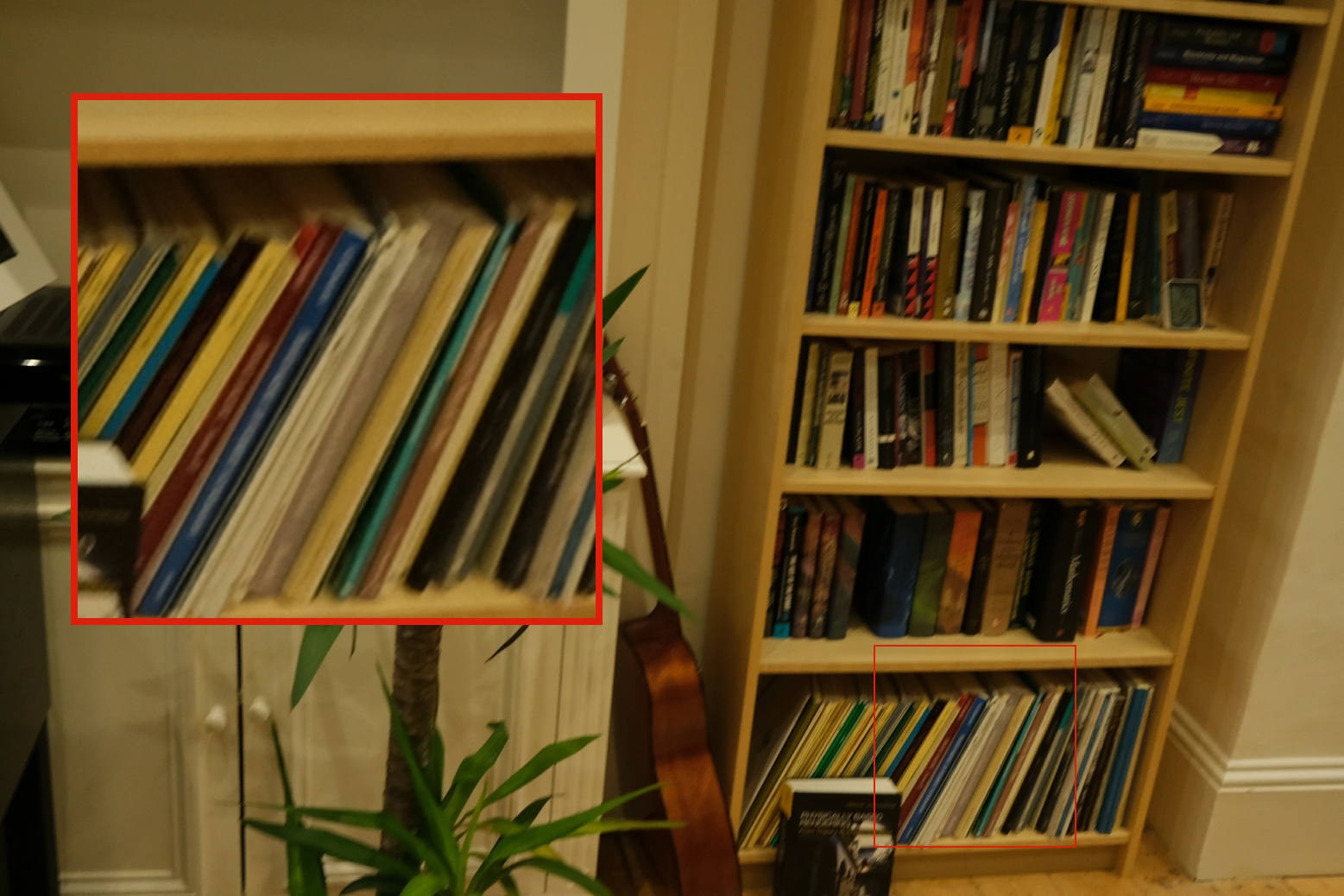} &
        \imgwithtext{\resLen}{\contour{black}{PSNR 29.8 / SSIM 0.918}}{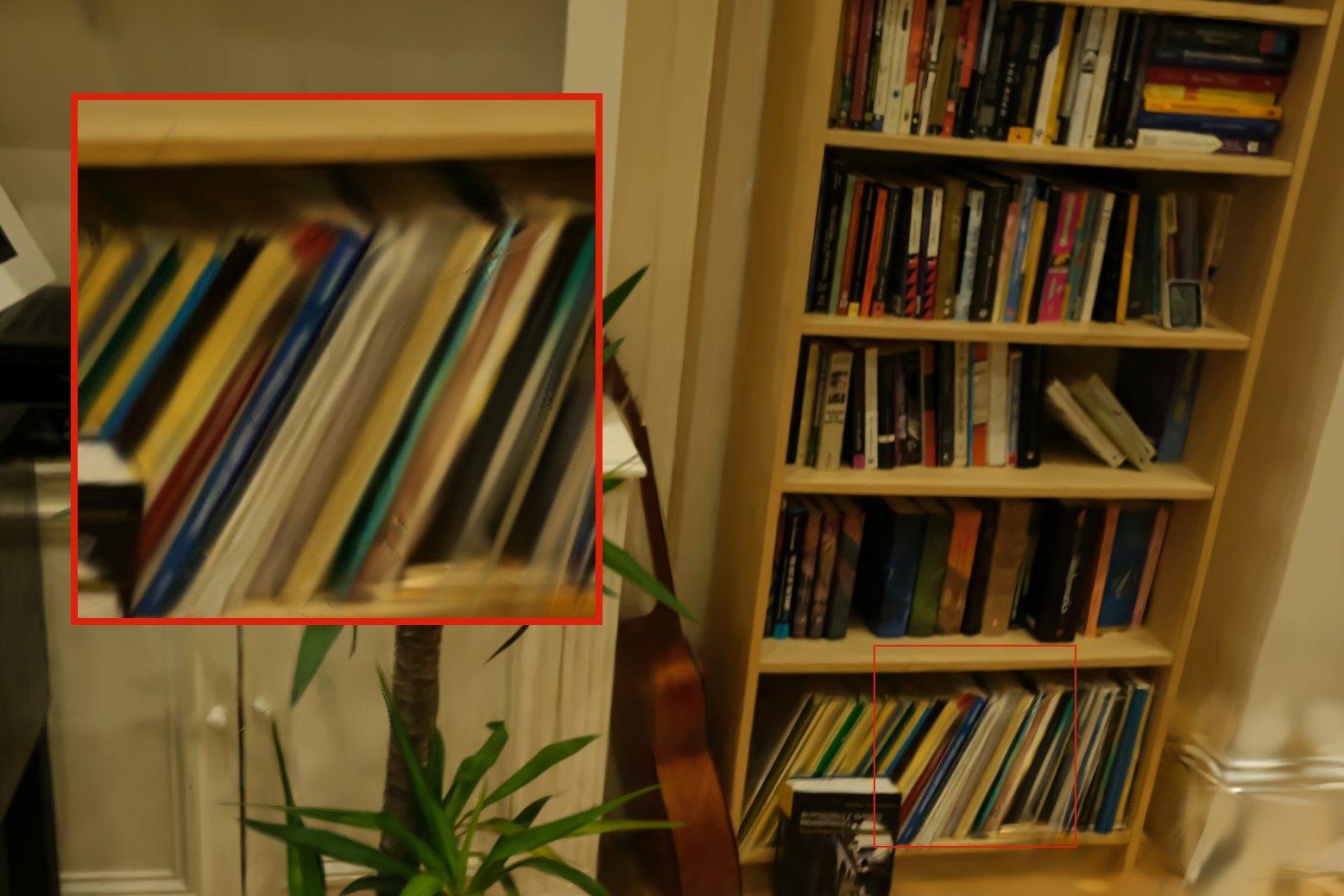} &
        \imgwithtext{\resLen}{\contour{black}{PSNR 28.8 / SSIM 0.908}}{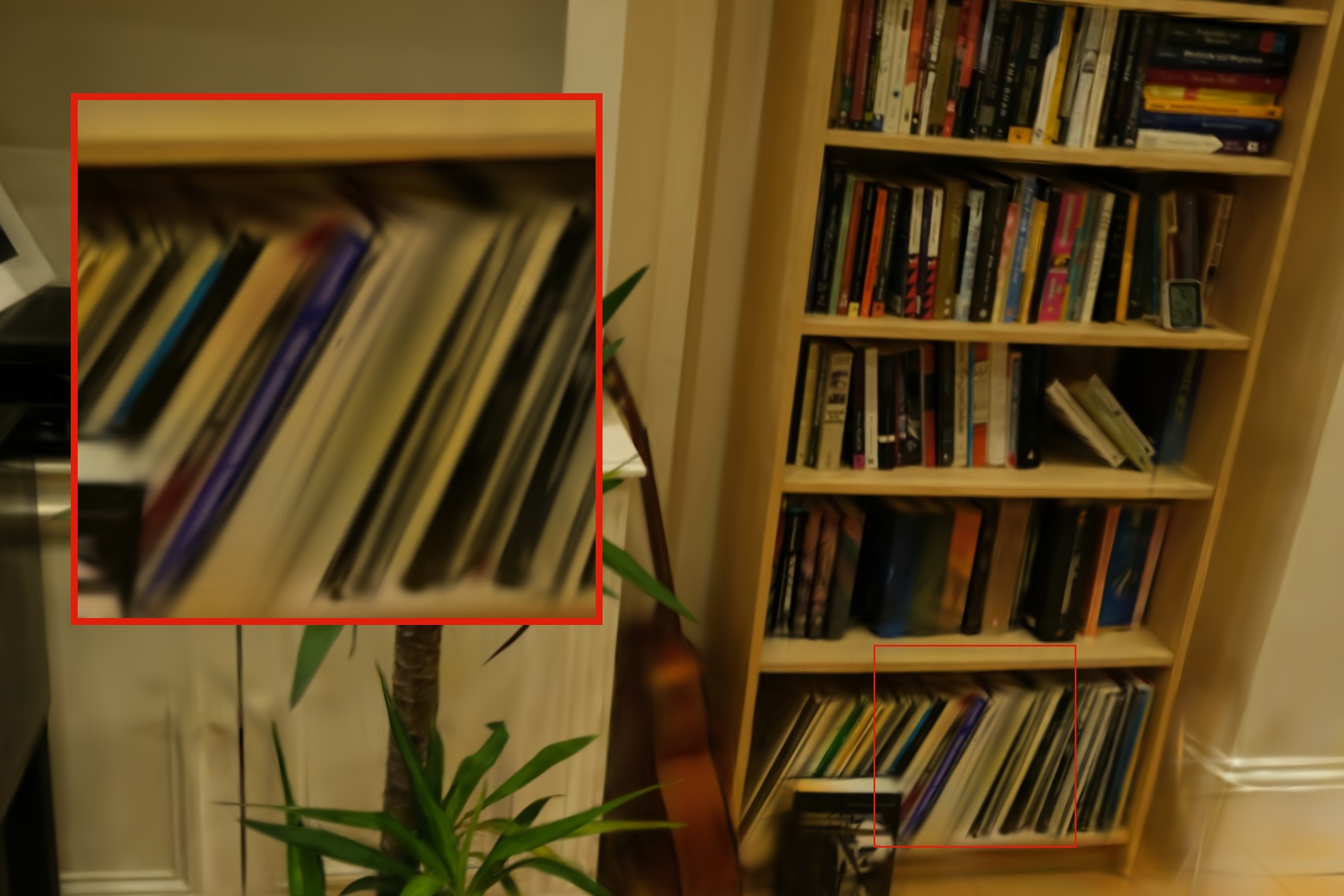} &
        \imgwithtext{\resLen}{\contour{black}{PSNR 27.4 / SSIM 0.880}}{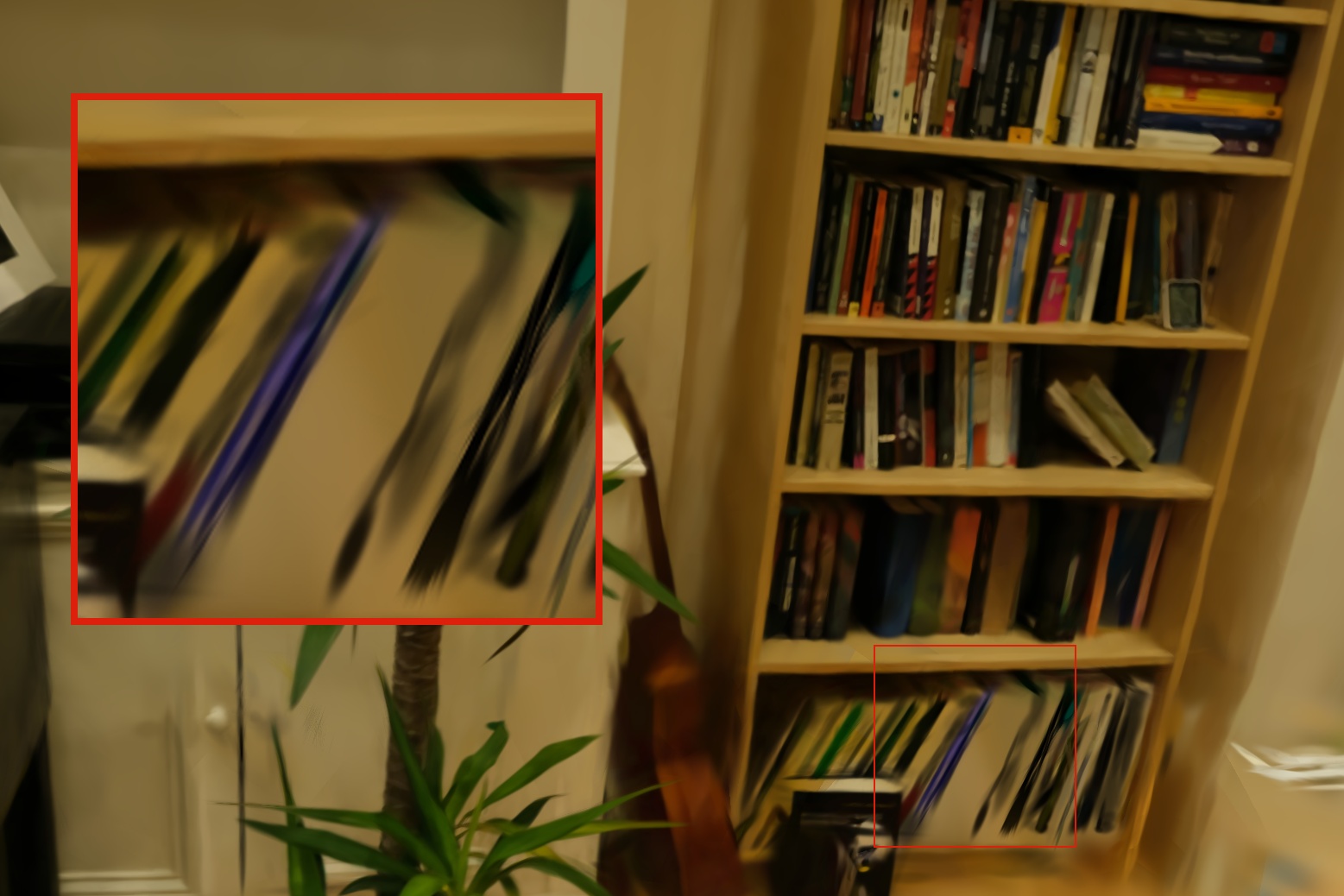} 

    \end{tabular}
    \caption{\textbf{Equal-time comparison} between our method and baselines. All methods run for the same wall-clock time. Our method produces comparable visual quality to 3DGS and outperforms 3DGRT.
    }
    \label{fig:nvs_figures}
\end{figure*}

\begin{figure*}[t]
    \setlength{\resLen}{1.75in}
    \centering
    \small
    \setlength{\tabcolsep}{1pt}
    \renewcommand{\arraystretch}{0}
    \begin{tabular}{cccc}
        \textbf{Reference} & (a) \textbf{Ours} & (b) \textbf{RNG} & (c) \textbf{GS$^3$}
        \\[3pt]
        \includegraphics[width=\resLen]{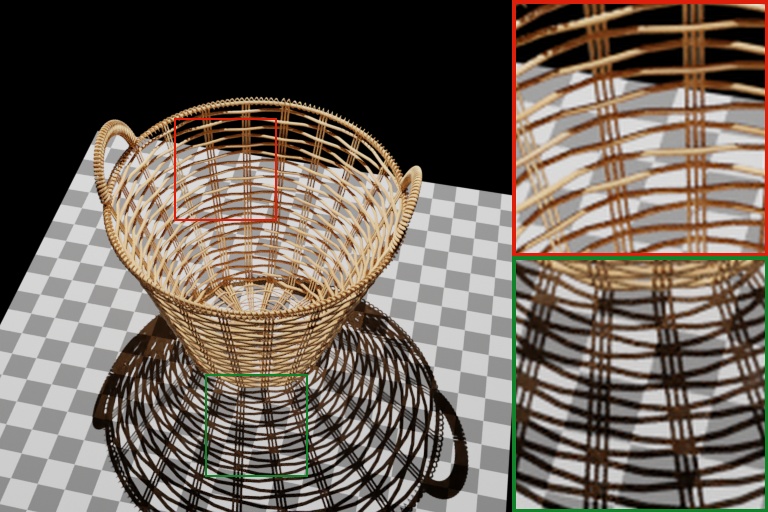} &
        \imgwithtext{\resLen}{\contour{black}{PSNR 28.0 / SSIM 0.956}}{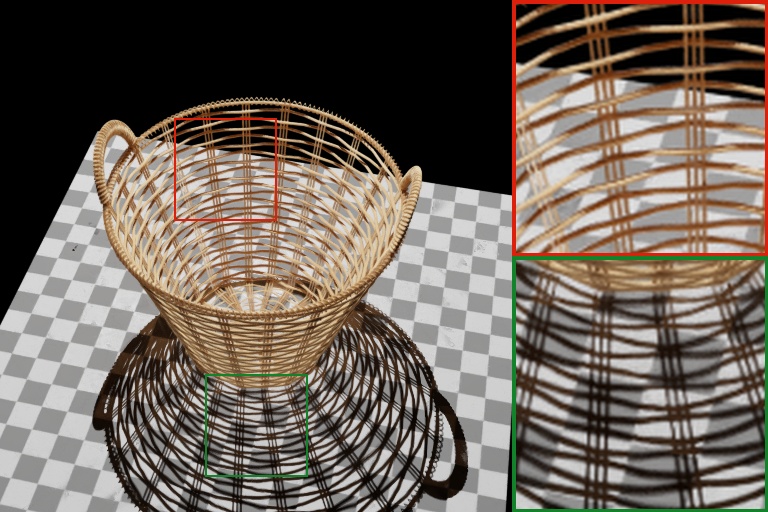} &
        \imgwithtext{\resLen}{\contour{black}{PSNR 20.0 / SSIM 0.853}}{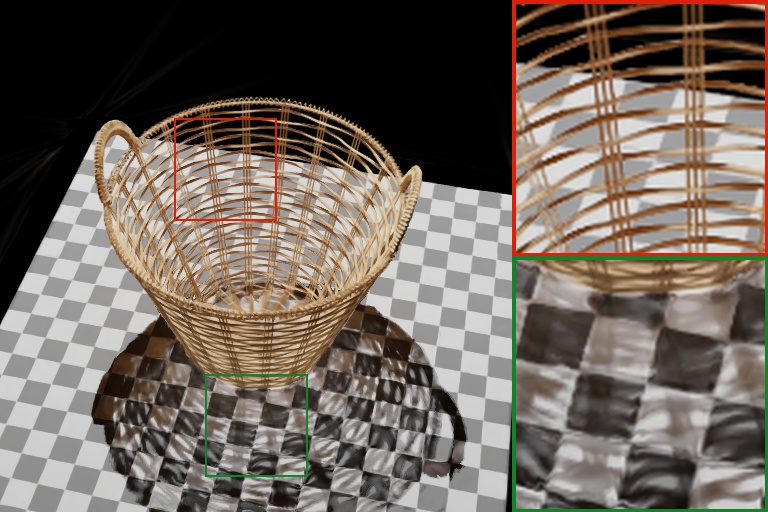} &
        \imgwithtext{\resLen}{\contour{black}{PSNR 23.2 / SSIM 0.936}}{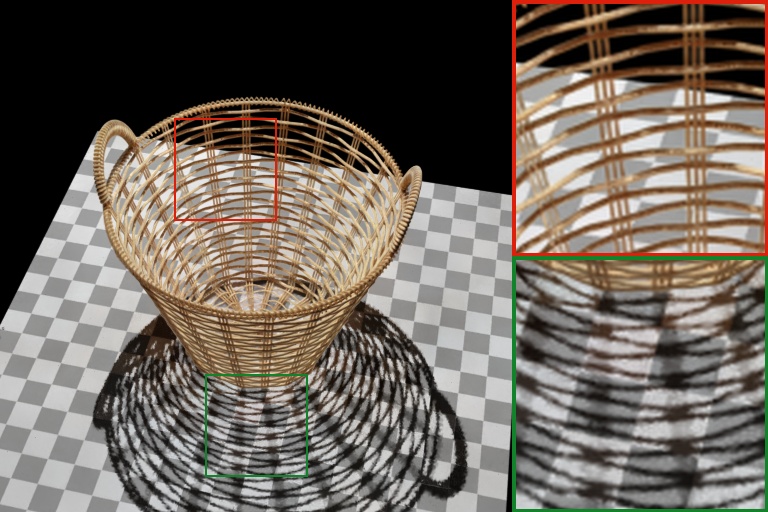}
        \\[2pt]
        \includegraphics[width=\resLen]{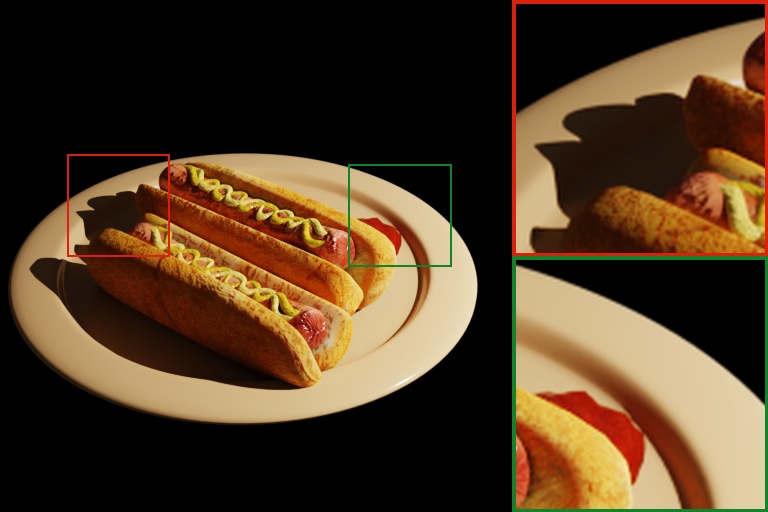} &
        \imgwithtext{\resLen}{\contour{black}{PSNR 31.9 / SSIM 0.955}}{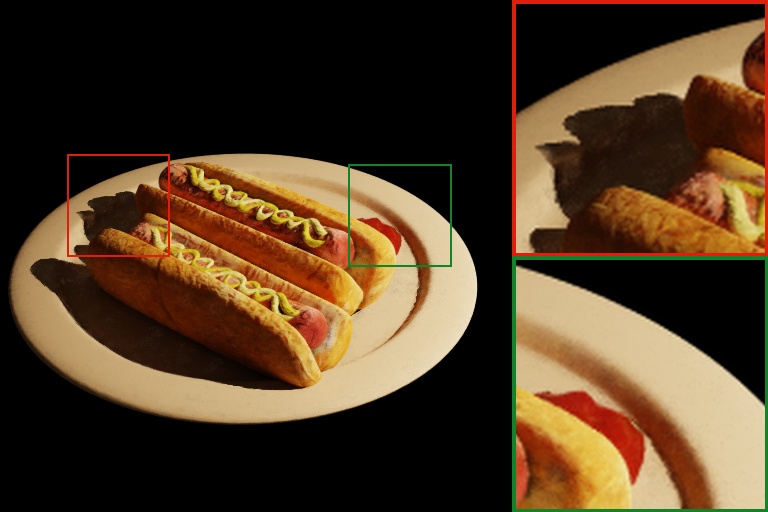} &
        \imgwithtext{\resLen}{\contour{black}{PSNR 30.4 / SSIM 0.960}}{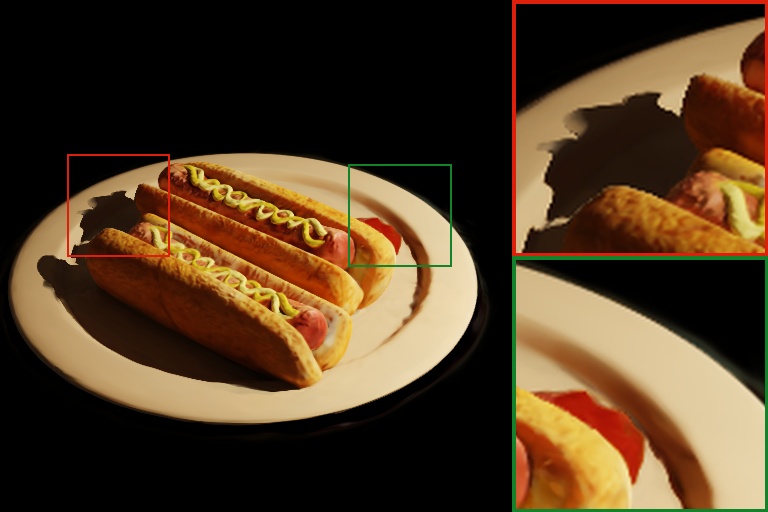} &
        \imgwithtext{\resLen}{\contour{black}{PSNR 25.4 / SSIM 0.949}}{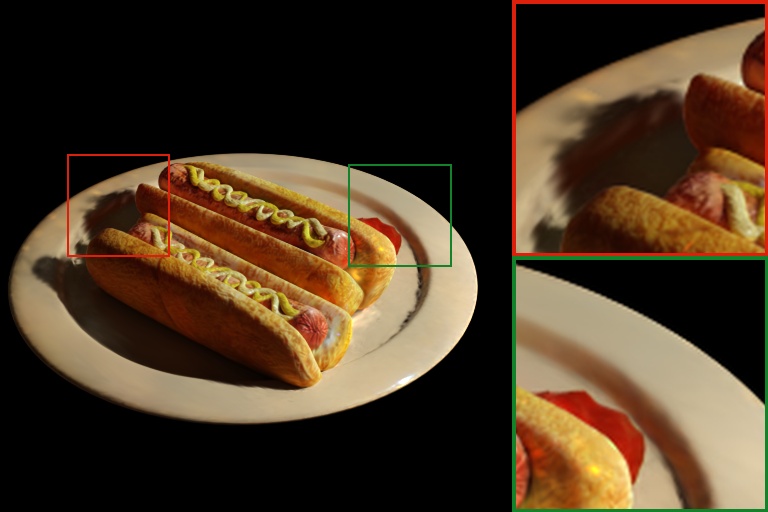}
    \end{tabular}
    \caption{Reconstruction quality of our method and baselines on the relightable benchmark. Our method produces significantly better geometry, especially shadow quality.
    }
    \label{fig:relightable_figures}
\end{figure*}

\begin{figure*}[t]
    \setlength{\resLen}{1.2in}
    \centering
    \small
    \addtolength{\tabcolsep}{-6.5pt}
    \begin{tabular}{cccccc}
        (a1) \textbf{Ours, original} & (b1) \textbf{Reference} & (c1) \textbf{Ours, relit} & (a2) \textbf{Ours, original} & (b2) \textbf{Reference} & (c2) \textbf{Ours, relit}
        \\
        \includegraphics[width=\resLen]{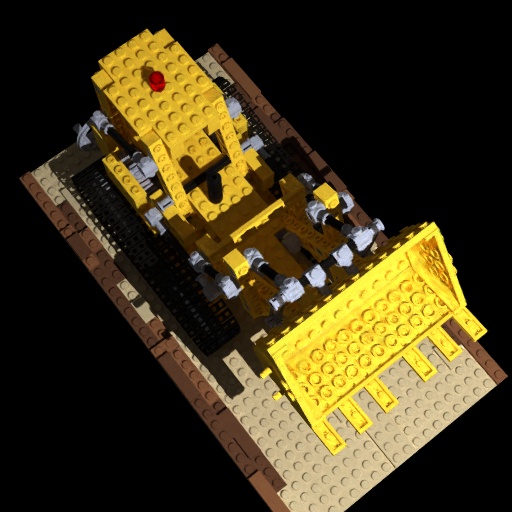} &
        \includegraphics[width=\resLen]{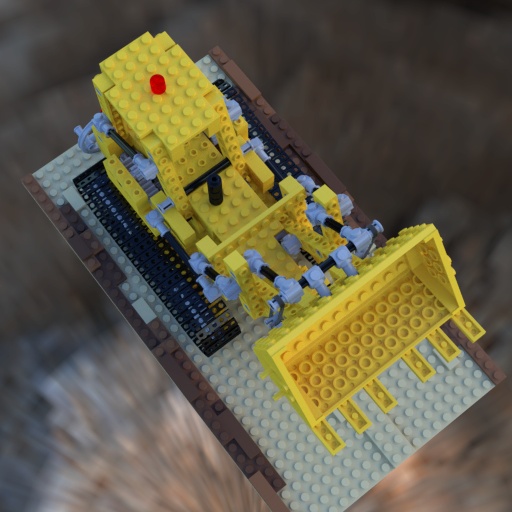} &
        \includegraphics[width=\resLen]{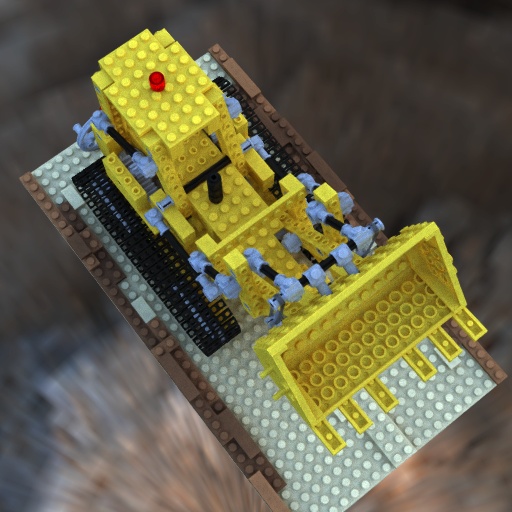} &
        \includegraphics[width=\resLen]{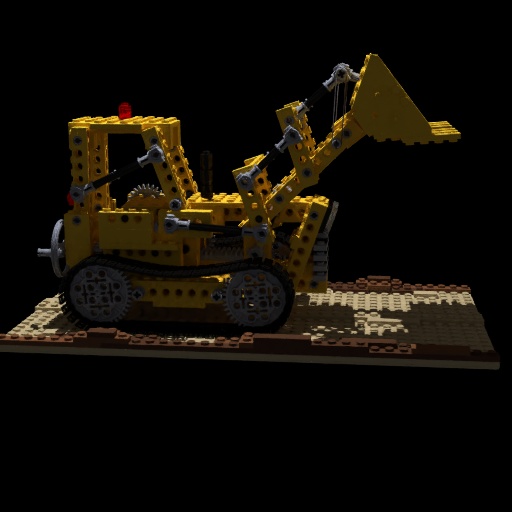} &
        \includegraphics[width=\resLen]{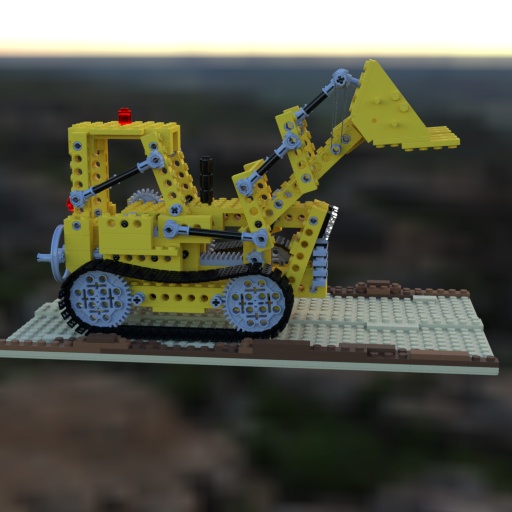} &
        \includegraphics[width=\resLen]{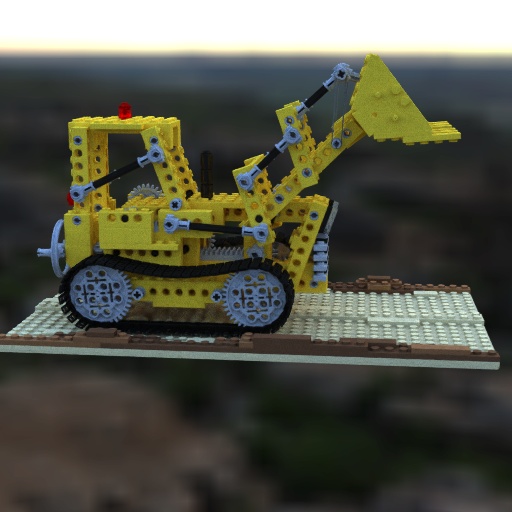}
    \end{tabular}
    \caption{
        Re-renderings of our reconstructed 3DGS objects under novel environmental illumination.
    }
    \label{fig:environment_map_relighting}
\end{figure*}

%% file: sec/X_suppl.tex
\clearpage
\setcounter{page}{1}
\maketitlesupplementary

\input{sec/app_proof}

\input{sec/app_ssplats}

\input{sec/app_results}

%% file: sec/app_proof.tex
\section{Proof of Unbiasedness}
\label{app:proof}
We now provide a proof of the unbiasedness for our Monte Carlo estimators $\dCdbcEst$ and $\dCdbalphaEst$ presented in Section 4.1 of the main paper.
We recall that these estimators work by first drawing an index $I$ and then setting the $I$-th components using Eq. (8) and Eq. (11) of the main paper while leaving all the other components to zero.

Then, it holds that
\begin{equation}
    \label{Eq:ex_dCdbcEst_i}
    \ex[\dCdbcEst_i] = \pr[I = i] \cdot 1,
\end{equation}
and
\begin{equation}
    \label{Eq:ex_dCdbalphaEst_i}
    \begin{aligned}
        \ex[\dCdbalphaEst_i] &= \pr[I = i] \cdot \ex\left[ \left. \frac{c_I - c_K}{\alpha_I} \right| I = i \right]\\
        &= \pr[I = i] \,\frac{c_i - \ex[c_K \,|\, I = i]}{\alpha_i}\\
        &= \frac{\pr[I = i]}{\alpha_i} \left( c_i - \sum_{k \succ i} \pr[K = k \,|\, I = i] \,c_k \right).
    \end{aligned}
\end{equation}
for all $i = 1, 2, \ldots, n$.

We recall that, according to Eq. (4) and Eq. (10) of the main paper, we have
\begin{align}
    \label{Eq:pr_I}
    \pr[I = i] &= \alpha_i \prod_{j \prec i} (1 - \alpha_j),\\
    \label{Eq:pr_K}
    \pr[K = k \,|\, I = i] &= \alpha_k \prod_{i \prec t \prec k} (1 - \alpha_t).
\end{align}
Substituting \autoref{Eq:pr_I} and \autoref{Eq:pr_K} into \autoref{Eq:ex_dCdbcEst_i} and \autoref{Eq:ex_dCdbalphaEst_i} yields
\begin{equation}
    \label{Eq:ex_dCdbcEst_i_1}
    \ex[\dCdbcEst_i] = \alpha_i \prod_{j \prec i} (1 - \alpha_j),
\end{equation}
and
\begin{multline}
    \label{Eq:ex_dCdbalphaEst_i_1}
    \ex[\dCdbalphaEst_i] =\\
    \left( \prod_{j \prec i} (1 - \alpha_j) \right) \left( c_i - \sum_{k \succ i} c_k \alpha_k \prod_{i \prec t \prec k} (1 - \alpha_t) \right).
\end{multline}
\autoref{Eq:ex_dCdbcEst_i_1} and \autoref{Eq:ex_dCdbalphaEst_i_1} match the desired partial derivatives $\nicefrac{\partial C}{\partial c_i}$ and $\nicefrac{\partial C}{\partial\alpha_i}$ given by Eq. (6) and Eq. (7) of the main paper, respectively.

%% file: sec/app_ssplats.tex
\section{Comparison to StochasticSplats}
\label{app:stocSplats}

\begin{table*}[t]
    \caption{\label{tab:ssplats_benchmarks}
        \textbf{Novel view synthesis}: Comparison between our method, our method with stochastic forward pass, and \cite{kv2025stochasticsplats}, in terms of both speed and quality.
    }
    \centering
    \small
    \begin{threeparttable}
        \footnotesize 
        \setlength{\tabcolsep}{4.2pt} 
        \begin{tabular}{
            l
            S[table-format=2.2] S[table-format=1.3] S[table-format=1.3] l
            S[table-format=2.2] S[table-format=1.3] S[table-format=1.3] l
            S[table-format=2.2] S[table-format=1.3] S[table-format=1.3] l
        }
        \toprule
        \multirow{2}{*}{Method\textbackslash Metric}
         & \multicolumn{4}{c}{\textbf{MipNeRF360}} 
         & \multicolumn{4}{c}{\textbf{Tanks \& Temples}}
         & \multicolumn{4}{c}{\textbf{Deep Blending}} \\
        \cmidrule(lr){2-5} \cmidrule(lr){6-9} \cmidrule(lr){10-13}
         & {PSNR$\uparrow$} & {SSIM$\uparrow$} & {LPIPS$\downarrow$} & {Time$\downarrow$}
         & {PSNR$\uparrow$} & {SSIM$\uparrow$} & {LPIPS$\downarrow$} & {Time$\downarrow$}
         & {PSNR$\uparrow$} & {SSIM$\uparrow$} & {LPIPS$\downarrow$} & {Time$\downarrow$} \\
        \midrule
        \textbf{Ours}     & 28.31 & 0.857 & 0.254 & 33m & 22.57 & 0.830 & 0.221 & 20m & 29.87 & 0.906 & 0.324 & 25m \\
        \textbf{Ours-FullStoch} (15spp)     & 27.90 & 0.843 & 0.265 & 37m & 22.37 & 0.821 & 0.233 & 21m & 29.51 & 0.902 & 0.330 & 24m \\
        \textbf{Ours-Fullstoch} (30spp)     & 28.14 & 0.850 & 0.254 & 52m & 22.75 & 0.827 & 0.229 & 25m & 29.75 & 0.904 & 0.327 & 29m \\
        \textbf{SS-Tracing}  & 22.12 & 0.648 & 0.451 & 89m & 14.94 & 0.532 & 0.560 & 54m & 18.08 & 0.726 & 0.578 & 68m \\
        \bottomrule
        \end{tabular}
    \end{threeparttable}
\end{table*}

Proposed by \citet{kv2025stochasticsplats}, StochasticSplats introduces a stochastic method for estimating gradients of the alpha-blended pixel colors $C$ with respect to the Gaussian parameters.
In what follows, we present the pseudo-code for the alternative estimator in \autoref{alg:ssplats}, and compare the efficiency of both estimators via empirical experiments.

\begin{algorithm}[t]
    \caption{\label{alg:ssplats}
        Monte Carlo estimator for pixel color gradient $\dCdbalpha$ introduced by StochasticSplats~\cite{kv2025stochasticsplats}.
    }
	\SetCommentSty{mycmtfn}
	\SetKwComment{tccinline}{// }{}
    %
    $\dCdbalphaEst^\mathrm{ssplats} \gets \boldsymbol{0}$\;
    \For{$m = 1$ to $\Mb$}{
        \tcc{Draw index $I$}
        $I \gets 0$; $z \gets \infty$; $\alpha \gets 1$\;
        \ForEach{Gaussian $g_i$ along the camera ray}{ 
            Obtain its opacity $\alpha_i$ and depth $z_i$\;
            Draw $\xi$ uniformly from $[0, 1)$\;
            \If{$\xi < \alpha_i$ and $z_i < z$}{
                $I \gets i$; $z \gets z_i$; $\alpha \gets \alpha_i$\; 
                
            }
        }
        \If{$I > 0$}{
            \tcc{Obtain Gaussian color}
            Compute the color $c_I$ of the Gaussian $g_I$\; 
            \ForEach{Gaussian $g_k$ along the camera ray}{
                Obtain its opacity $\alpha_k$ and depth $z_k$\;
                \If{$z_k < z_I$}{
                    \tcc{Update the gradient estimates}
                    $\dCdbalphaEst^\mathrm{ssplats}_k \pluseq \nicefrac{-c_I}{(1 - \alpha_k)}$
                        \label{line:div_by_zero}
                        \tccinline*{\autoref{Eq:ssplats_dCdbalpha_1}}
                }
            }            
            \tcc{Update the gradient estimates}
            $\dCdbalphaEst^\mathrm{ssplats}_I \pluseq \nicefrac{c_I}{\alpha}$
                \tccinline*{\autoref{Eq:ssplats_dCdbalpha_0}}
        }
    }
    \Return 
        $\nicefrac{\dCdbalphaEst^\mathrm{ssplats}}{\Mb}$\;
\end{algorithm}

\subsection{Alternative Gradient Estimator}
\label{ssec:ssplats_grad_est}

StochasticSplats~\cite{kv2025stochasticsplats} uses the same procedure as our method to estimate the gradient $\dCdbc$ but a different one for $\dCdbalpha$.
In the following, we focus on the estimation of the latter.

As described in \autoref{alg:ssplats}, to estimate the gradient $\dCdbalpha := (\nicefrac{\partial C}{\partial\alpha_1}, \nicefrac{\partial C}{\partial\alpha_2}, \ldots, \nicefrac{\partial C}{\partial\alpha_n})$ with respect to the Gaussian opacities $\alpha_1, \alpha_2, \ldots, \alpha_n$, StochasticSplats starts with drawing an index $I$ the same way as our method, and setting
\begin{equation}
    \label{Eq:ssplats_dCdbalpha_0}
    \dCdbalphaEst^\mathrm{ssplats}_I = \frac{c_I}{\alpha_I}.
\end{equation}
Then, instead of randomly drawing some $K \succ I$, StochasticSplats lets
\begin{equation}
    \label{Eq:ssplats_dCdbalpha_1}
    \dCdbalphaEst^\mathrm{ssplats}_k = \frac{-c_I}{1 - \alpha_k},
\end{equation}
%
for all Gaussians $g_k$ located in front of the Gaussian $g_I$ (i.e., with $k \prec I$).

Although this approach returns denser gradient estimates by spatting gradients to not only the $I$-th component but also all $k \prec I$, it suffers from two major drawbacks.
First, when executed on the GPU, writing to all $k \prec I$ components in parallel requires significantly more atomic operations, which can lead to reduced computational efficiency.
Second, the division by $(1 - \alpha_k)$ in \autoref{Eq:ssplats_dCdbalpha_1} can produce near-infinite gradients---and therefore high variance---whenever high-opacity Gaussians make $\alpha_k \approx 1$, which occurs frequently.

\input{sec/grad_img}

The key difference between our method and \cite{kv2025stochasticsplats} is the order of derivation. StochasticSplats~\cite{kv2025stochasticsplats} first designs a Monte Carlo estimator for alpha blending, and subsequently takes the derivative of the estimator. Our method, in contrast, first takes the derivative of alpha blending, and designs a Monte Carlo estimator for the derivative. This ordering introduces a term $\nicefrac{1}{1-\alpha_{K\prec I}}$ in the density gradient, where the denominator easily evaluates to a near-$0$ value when an opaque Gaussian exists, causing problematic gradients.

\subsection{Comparisons}
\label{ssec:ssplats_comp}
To compare the efficiency of our gradient estimator (Algorithm~2 of the main paper) and the alternative one proposed by StochasticSplats (\autoref{alg:ssplats}), we implement the latter on our stochastic ray tracing codebase, which we refer to as \textsl{SS-Tracing}.

We give an intuition on the variance of both methods in \autoref{fig:gradient_image} by visualizing the gradient image. The images show the magnitude of screen space gradient with respect to the horizontal motion of the Gaussians, as well as a plot of the opacities of Gaussians along a ray. Our experiment shows that high opacity Gaussians frequently occur, and can easily lead to large variance for \textsl{SS-Tracing}. Empirically, it takes $16\times$ samples for SS-Tracing to reach the same level of variance as our method.


We compare the numerical metrics of our method and our re-implemented \textsl{SS-Tracing} in \autoref{tab:ssplats_benchmarks}.

%% file: sec/grad_img.tex
\begin{figure*}[t]
    \setlength{\resLen}{1.5in}
    \centering
    \small
    \addtolength{\tabcolsep}{-5.5pt}
    \begin{tabular}{cc}
        (a) \textbf{Forward} & (b) \textbf{Opacity Values Along the Ray}
        \\
        \includegraphics[trim={30 0 30 100}, clip, width=\resLen]{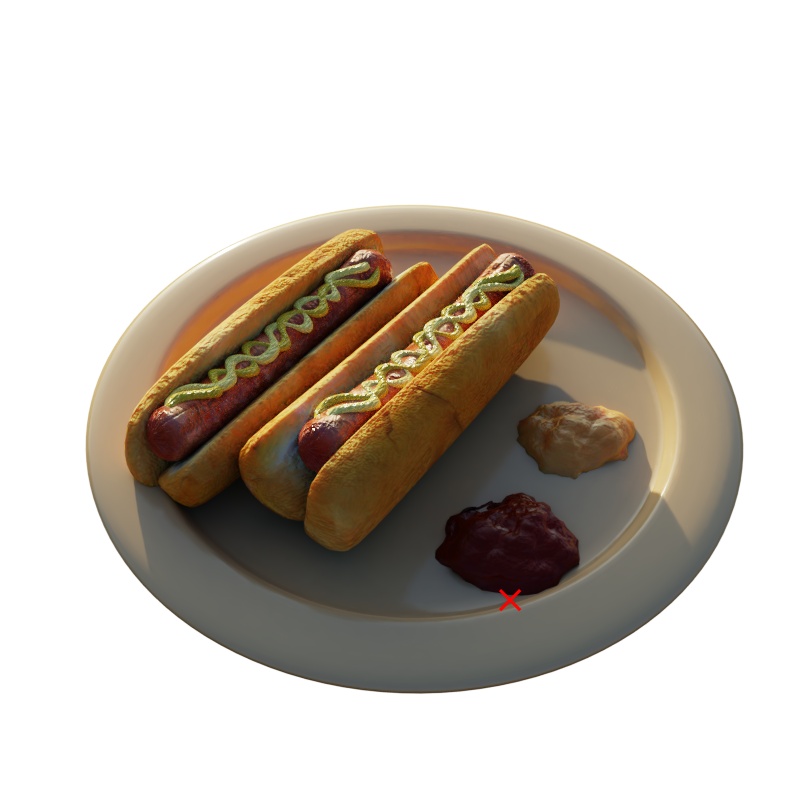} &
        \includegraphics[height=\resLen]{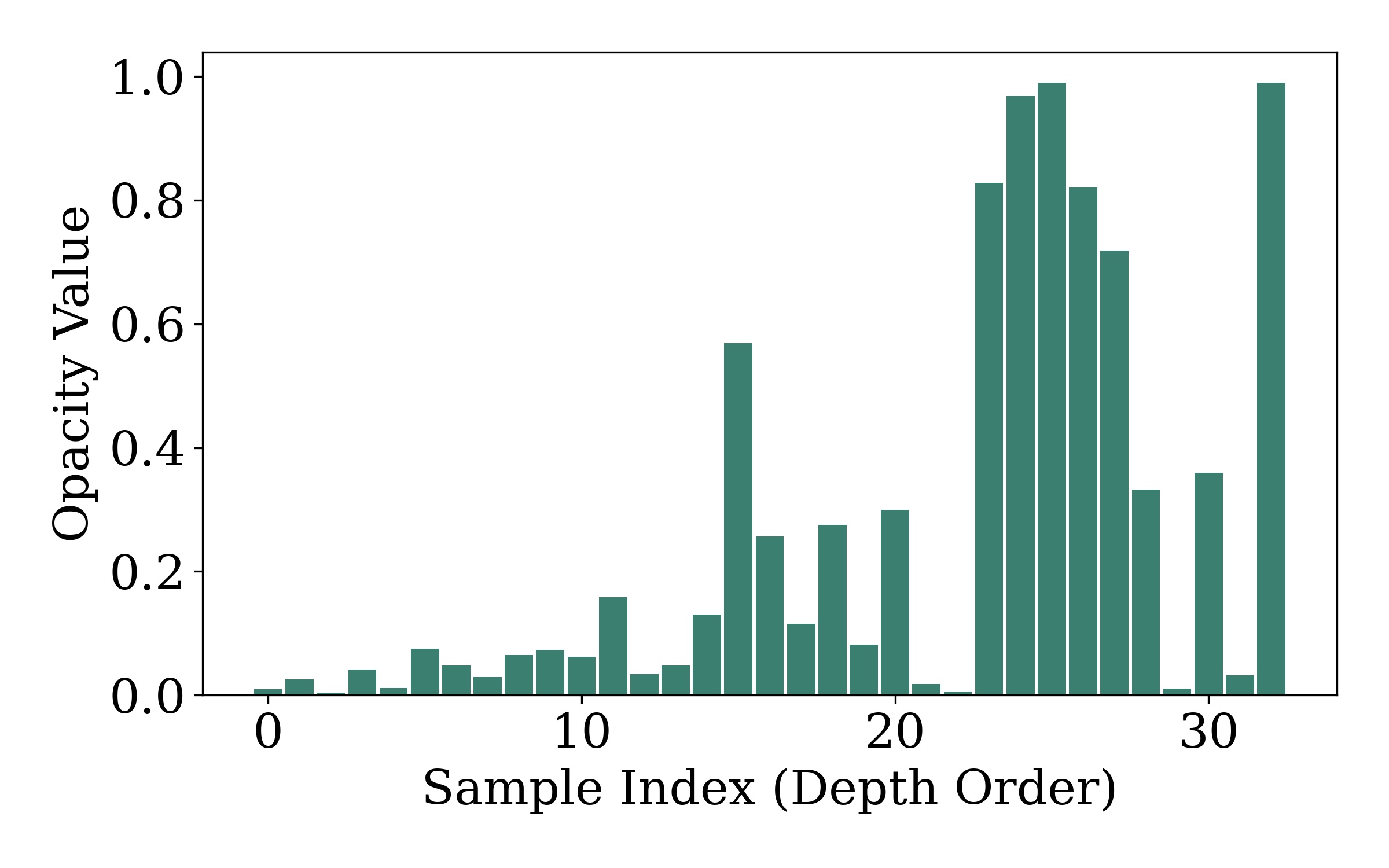}
    \end{tabular}
    \begin{tabular}{cccc}
        (c) \textbf{Sorted} & (d) \textbf{Ours, $M_b=8$}  & (e) \textbf{SS-Tracing, $M_b=8$}  & (f) \textbf{SS-Tracing, $M_b=128$}
        \\
        \begin{overpic}[trim={30 0 30 100}, clip, width=\resLen]{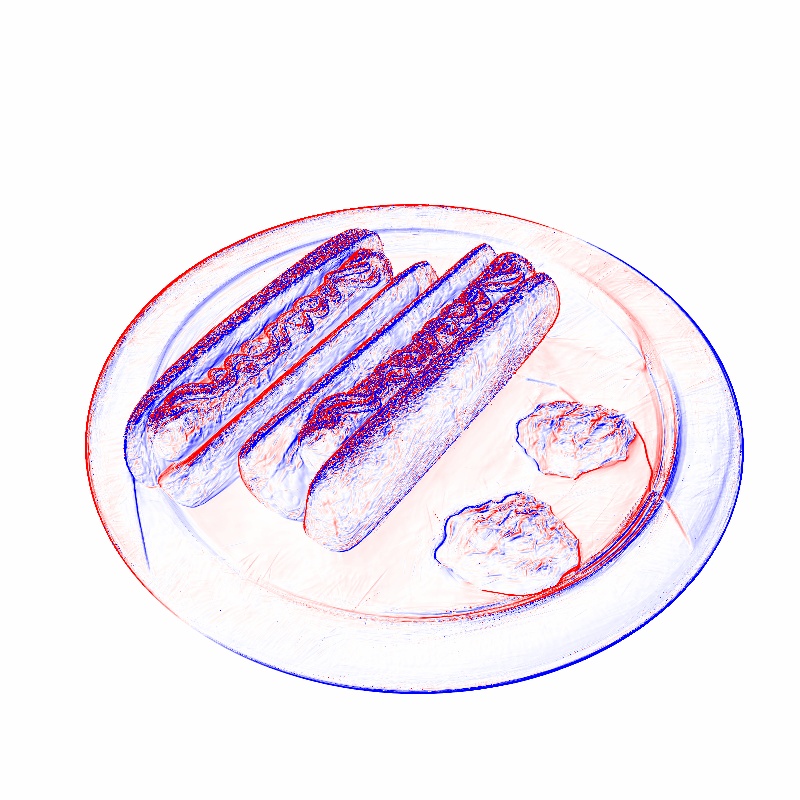}
            \put(50, 0){\makebox(0,0){\color{black}{\scriptsize \textbf{Time: 21.46ms}}}}
        \end{overpic} &
        \begin{overpic}[trim={30 0 30 100}, clip, width=\resLen]{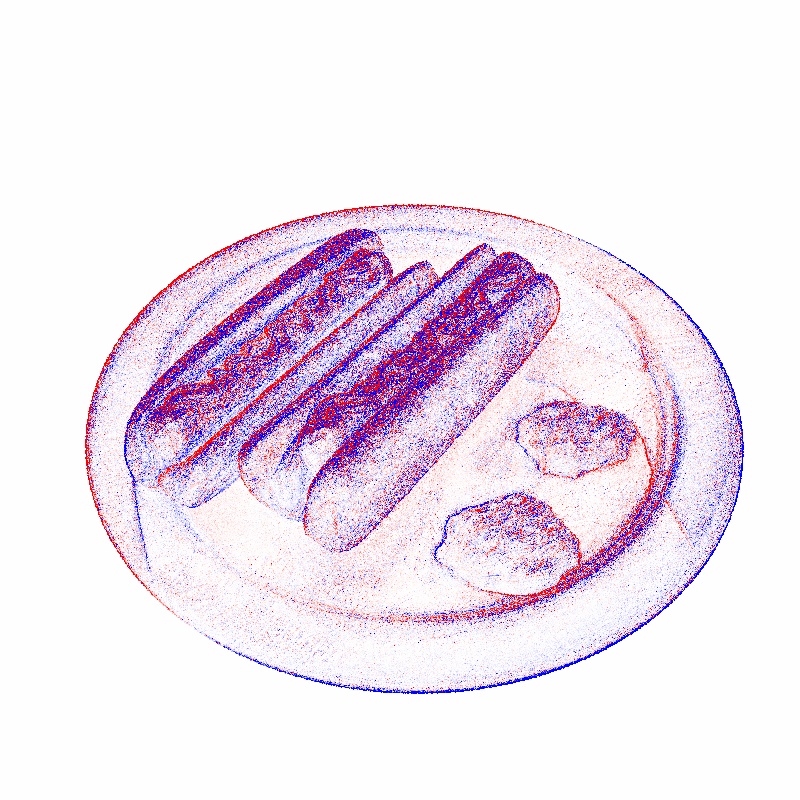}
            \put(50, 0){\makebox(0,0){\color{black}{\scriptsize \textbf{Time: 10.87ms | RMSE: 0.408}}}}
        \end{overpic} &
        \begin{overpic}[trim={30 0 30 100}, clip, width=\resLen]{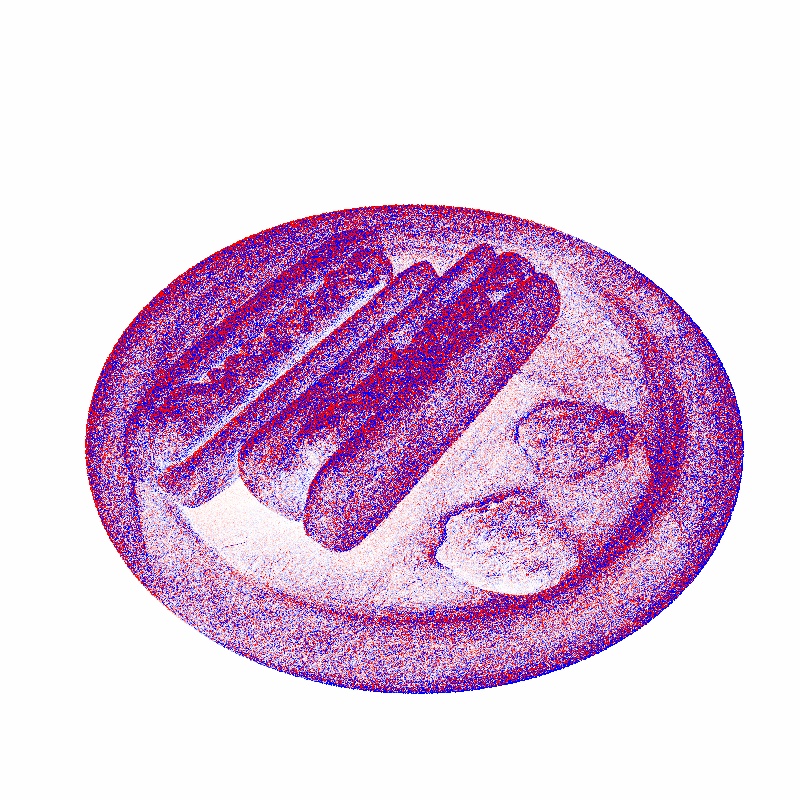}
            \put(50, 0){\makebox(0,0){\color{black}{\scriptsize \textbf{Time: 28.22ms | RMSE: 1.432}}}}
        \end{overpic} &
        \begin{overpic}[trim={30 0 30 100}, clip, width=\resLen]{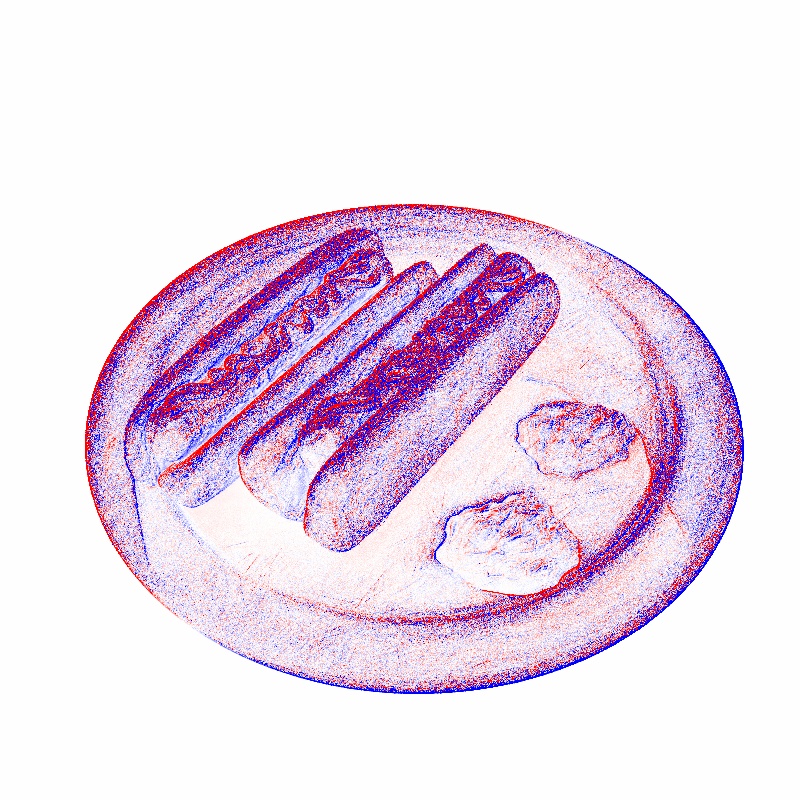}
            \put(50, 0){\makebox(0,0){\color{black}{\scriptsize \textbf{Time: 484.04ms | RMSE: 0.427}}}}
        \end{overpic}
    \end{tabular}
    \caption{
        Visualization of the variance of the gradient obtained by our method and the baseline method. (a) shows the forward rendering result, and (b) visualizes the opacities of each Gaussian along the ray at the red cross in (a). We further visualize the gradient that the Gaussians along each pixel receive when applying: (c) Sorted alpha blending (3DGRT), (d) Ours, with $M_b=8$, (e) Our implementation of~\cite{kv2025stochasticsplats}, with $M_b=8$, and (f) Our implementation of~\cite{kv2025stochasticsplats}, with $M_b=128$. The time for the backward pass and RMSE are provided with each image.
    }
    \label{fig:gradient_image}
\end{figure*}

%% file: sec/app_results.tex
\section{Additional Results}
\label{app:results}
We provide additional evaluation of our method, following Section 5 of the main text.

In \autoref{tab:ssplats_benchmarks}, we provide an ablation study of an important variant of our algorithm. We replace the sorting-based forward pass with the stochastic algorithm in Algorithm 1 of the main paper, proposed by \citet{Sun2025}, vary the number of samples used in the forward pass and evaluate the performance on novel view synthesis tasks. Experiment shows that our hybrid approach consistently outperforms the full stochastic variants in terms of both reconstruction quality and time. 

\begin{table}[t]
    \centering
    \caption{
        NeRF Synthetic Dataset
    }
    \small
    \begin{tabular}{c|c c c c}
        \toprule
        & \textbf{PSNR$\uparrow$} & \textbf{SSIM$\uparrow$} & \textbf{LPIPS$\downarrow$} & \textbf{Time$\downarrow$}\\
        \midrule
        3DGRT & 33.84 & 0.970 & 0.038 & 10m21s \\
        SS-Tracing & 31.67 & 0.958 & 0.050 & 24m31s\\
        \textbf{Ours} & 33.99 & 0.970 & 0.039 & 6m57s\\
        \bottomrule
    \end{tabular}
    \label{tab:benchmark_nerf}
\end{table}

In \autoref{tab:benchmark_nerf}, we show the metrics of our method, 3DGRT and SS-Tracing on the NeRF-Synthetic dataset. Our performance is consistent with the results we show in the main paper.

\begin{figure}[t]
    \setlength{\resLen}{1.5in}
    \centering
    \small
    \addtolength{\tabcolsep}{-5.5pt}
    \begin{tabular}{cc}
        (a) \textbf{Original} & (b) \textbf{Ours}
        \\
        \includegraphics[width=\resLen]{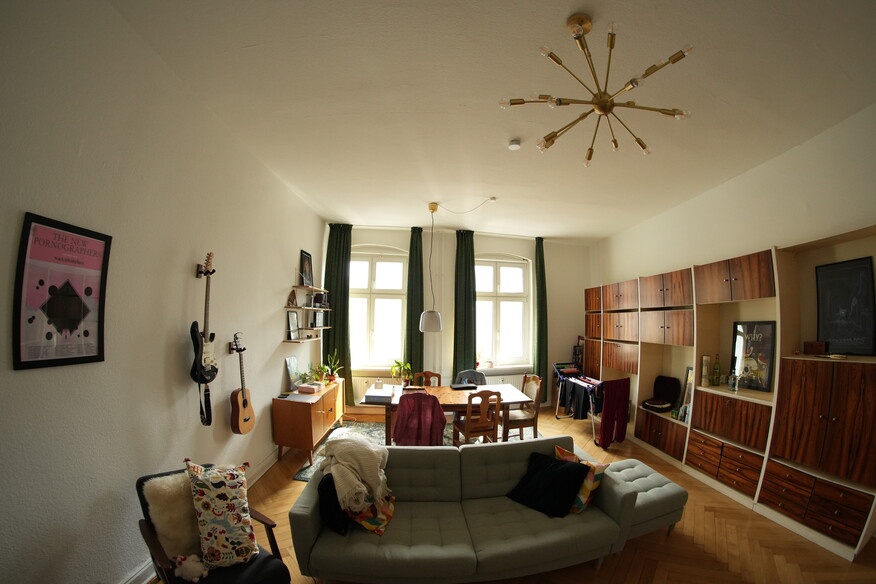} &
        \includegraphics[width=\resLen]{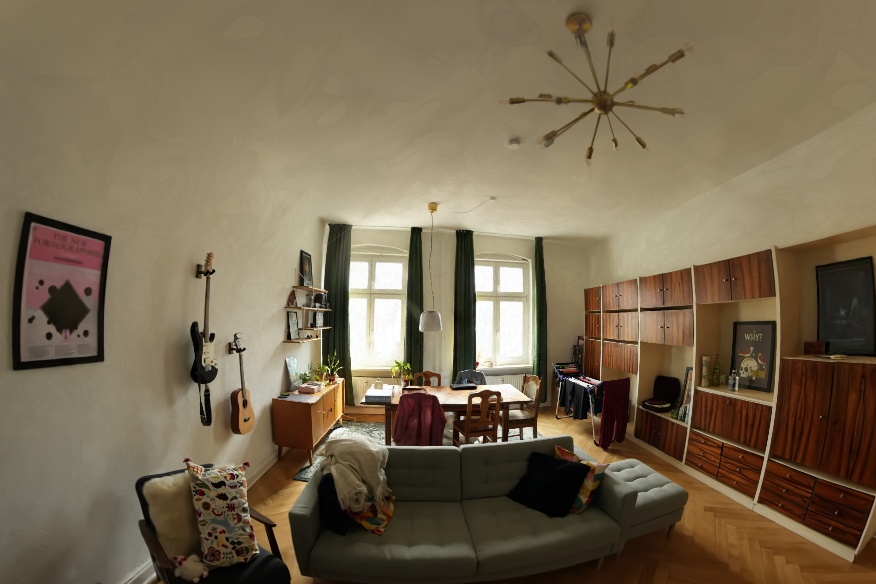} 
        \\
        \includegraphics[width=\resLen]{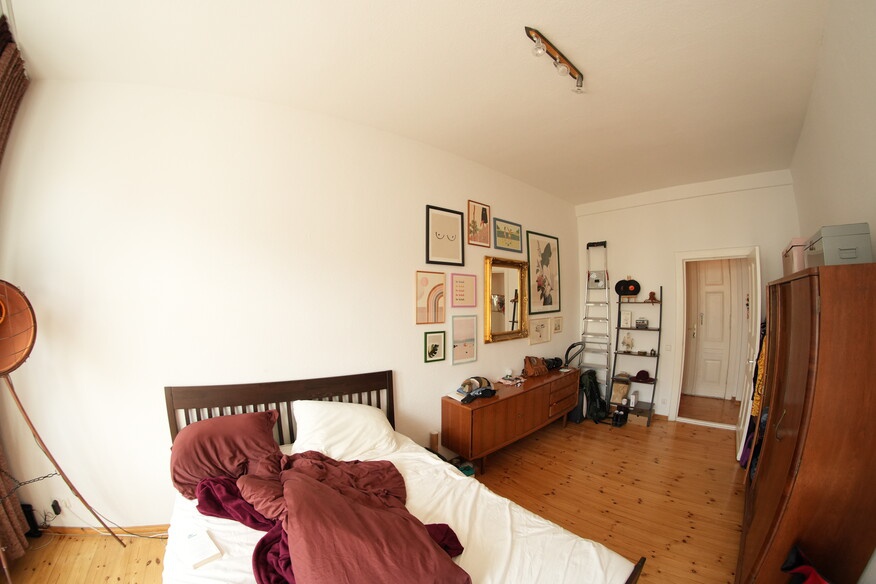} &
        \includegraphics[width=\resLen]{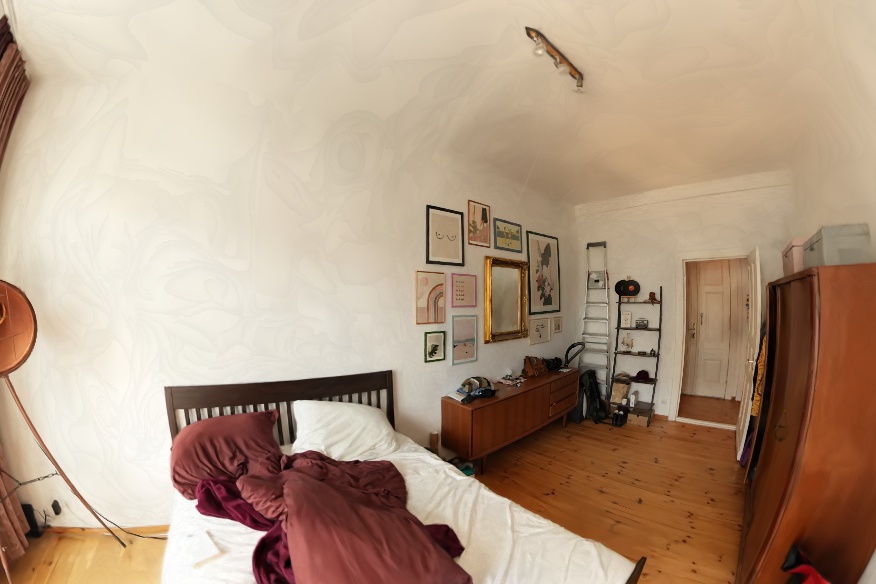}
    \end{tabular}
    \caption{
        Similar to our ray tracing baseline, our method also seamlessly supports distorted cameras, e.g., fisheye.
    }
    \label{fig:fisheye}
\end{figure}

In \autoref{fig:fisheye}, we demonstrate that our method is seamlessly compatible with distorted camera models due to the usage of ray tracing.

\begin{figure*}[t]
    \setlength{\resLen}{1.75in}
    \centering
    \small
    \setlength{\tabcolsep}{1pt}
    \renewcommand{\arraystretch}{0}
    \begin{tabular}{cccc}
        \textbf{Reference} & (a) \textbf{Ours} & (b) \textbf{3DGS} & (c) \textbf{3DGRT}
        \\[3pt]
        \includegraphics[width=\resLen]{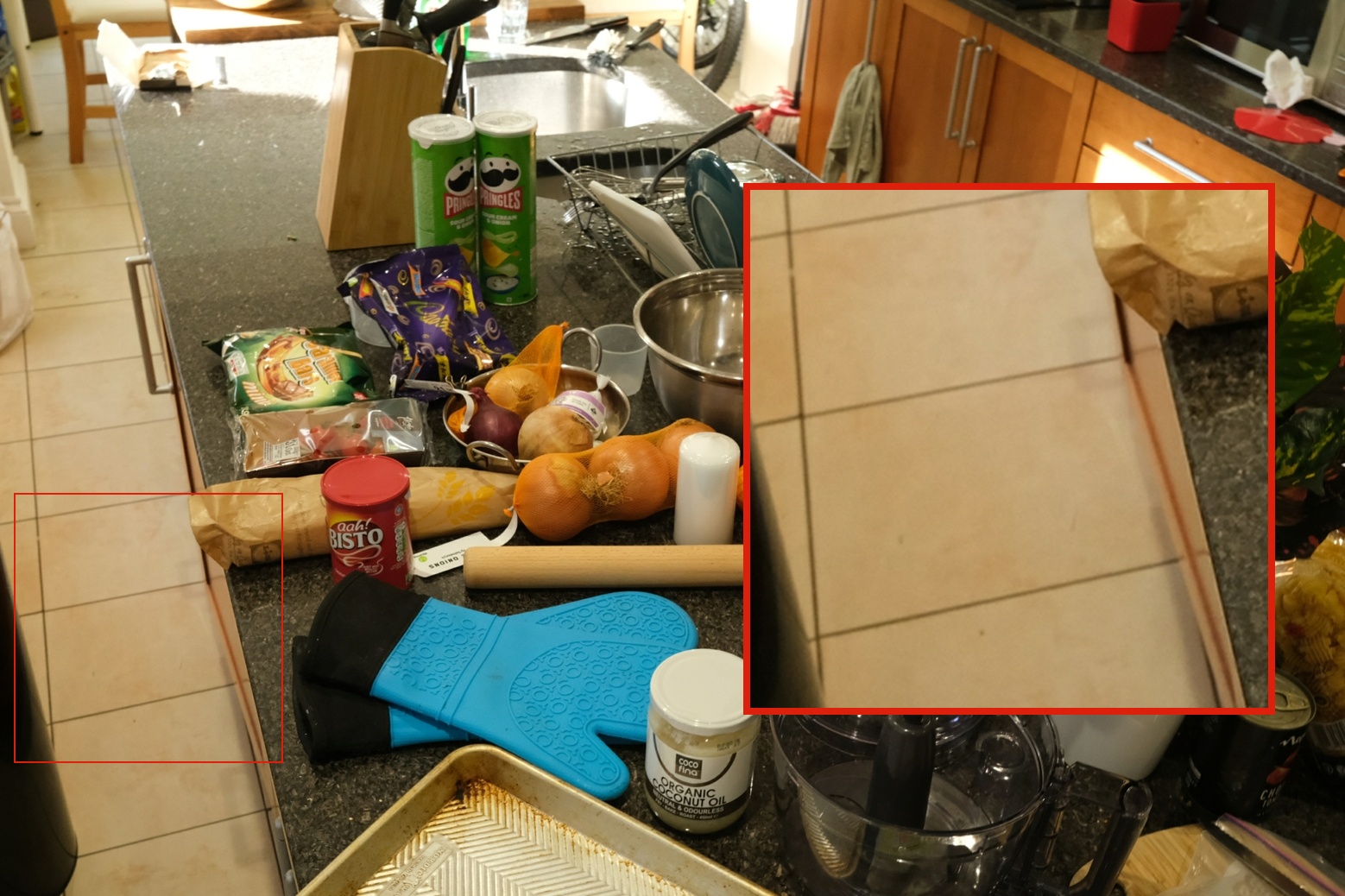} &
        \includegraphics[width=\resLen]{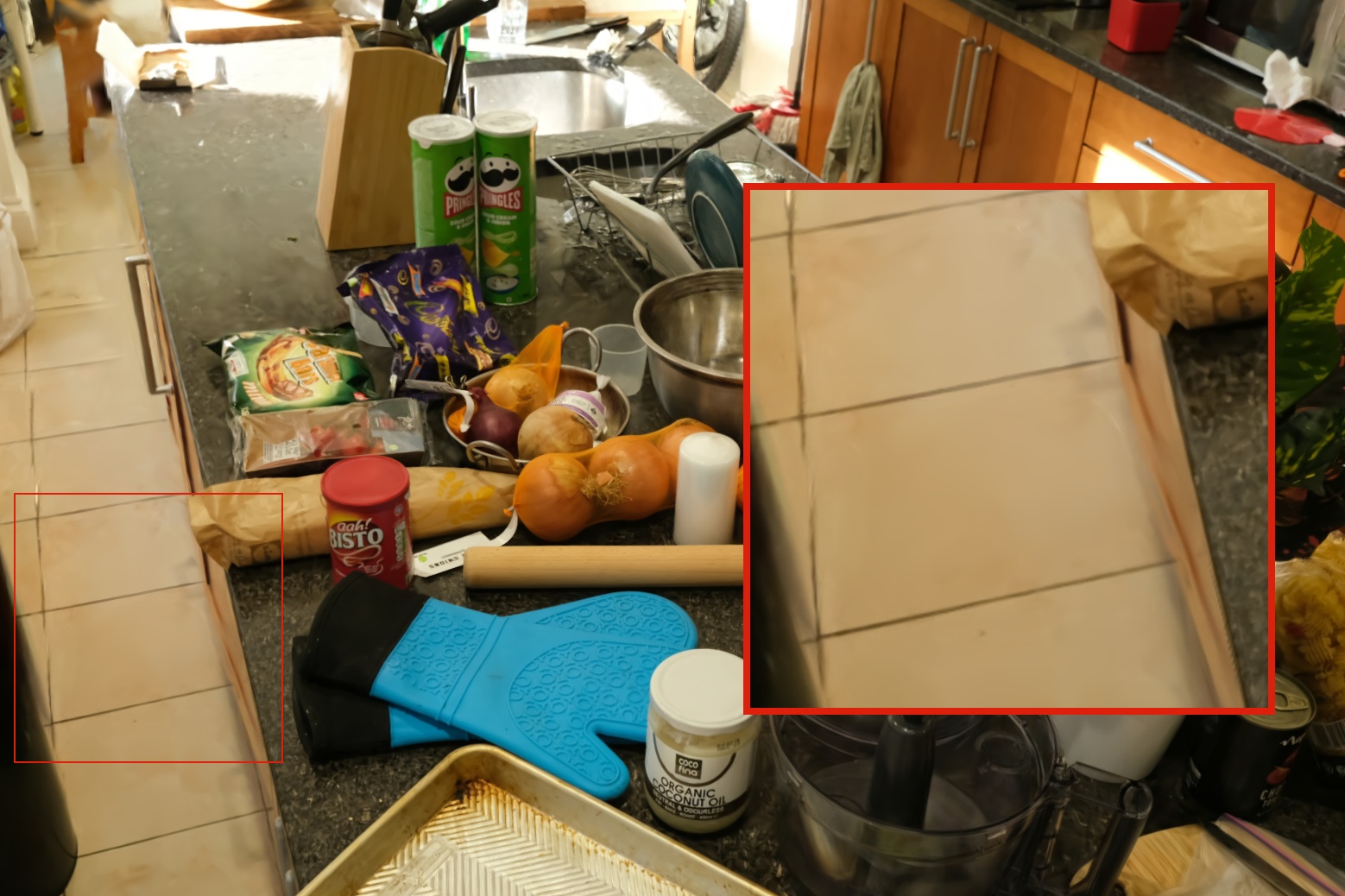} &
        \includegraphics[width=\resLen]{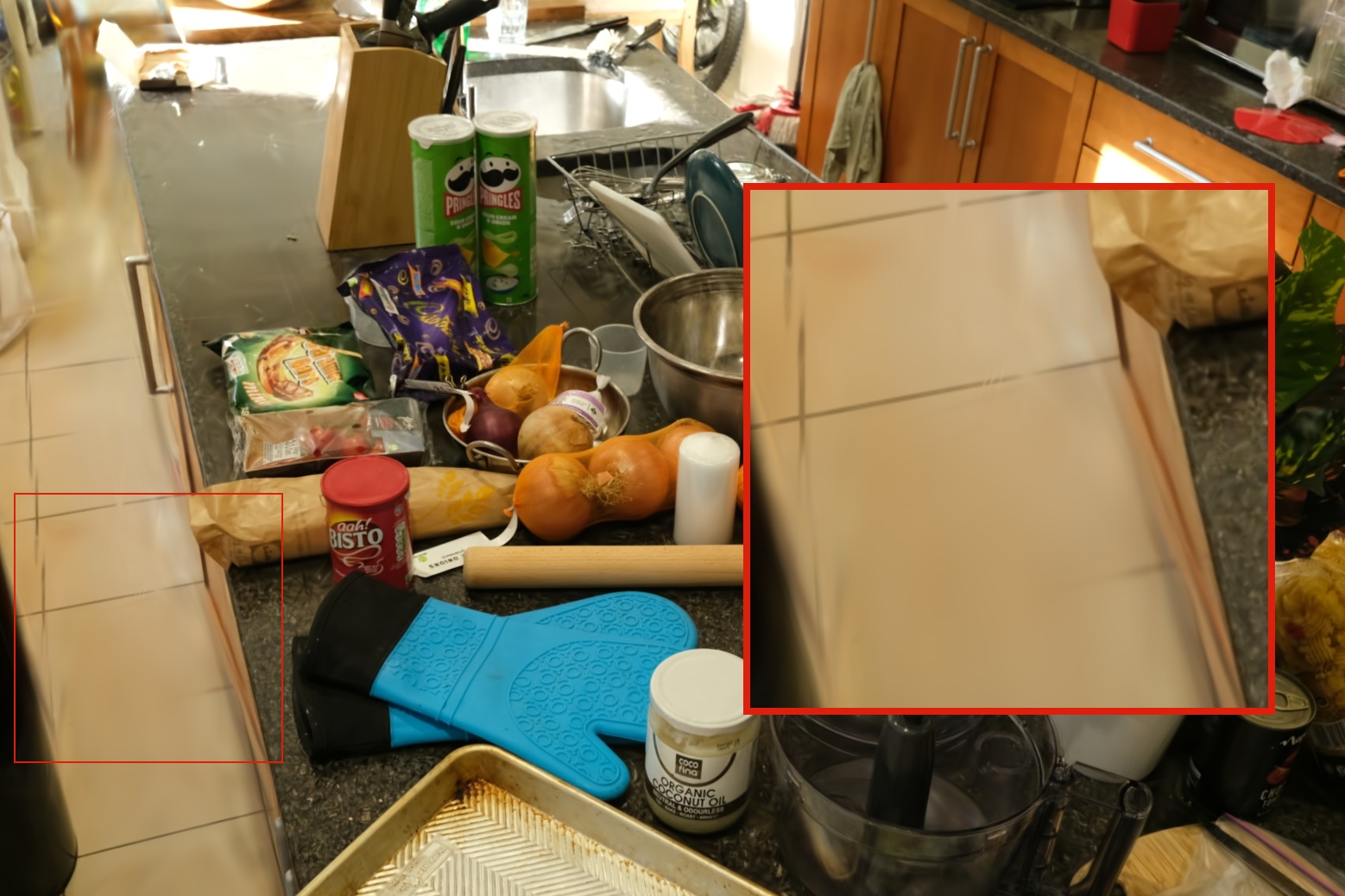} &
        \includegraphics[width=\resLen]{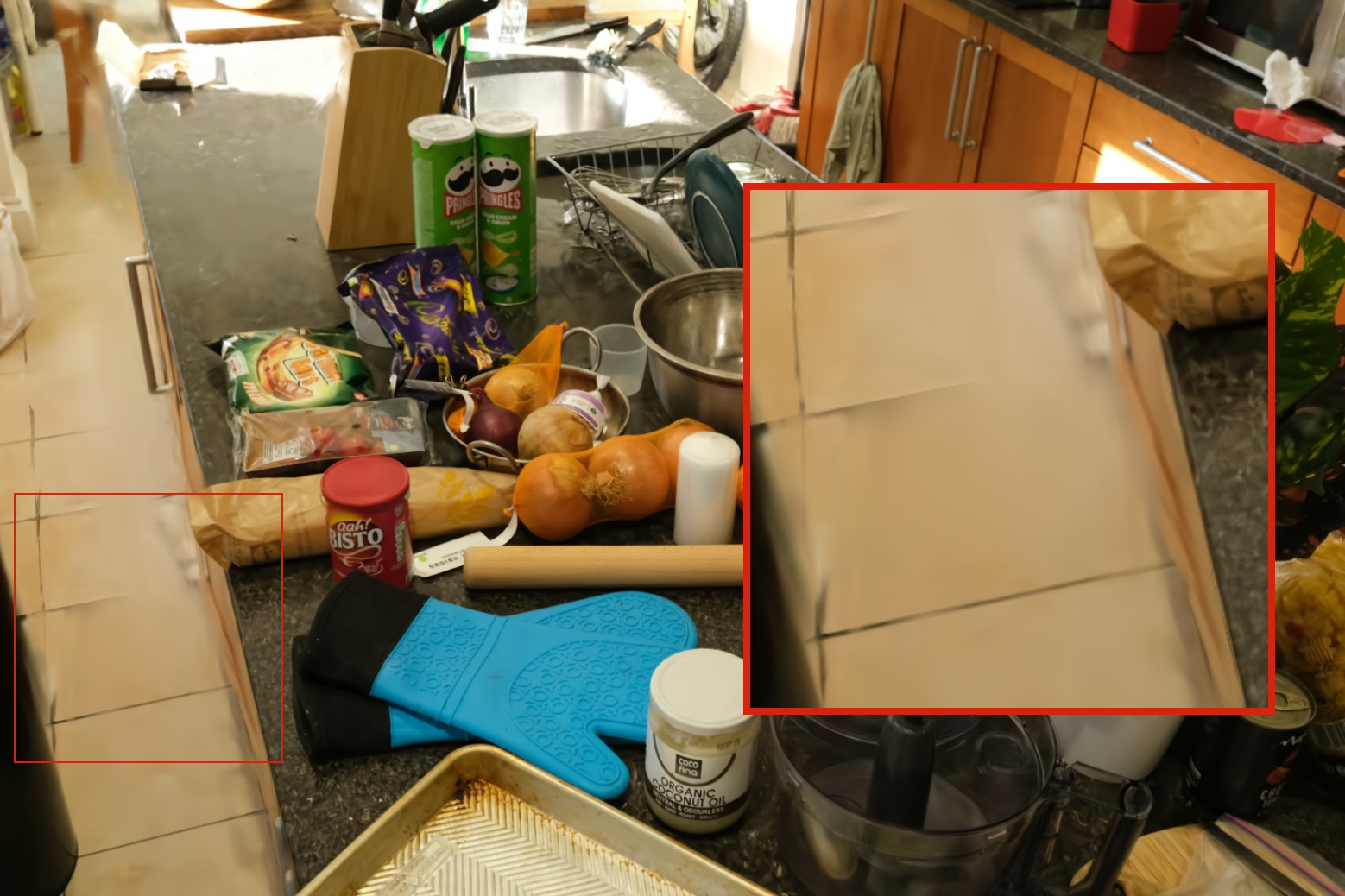} 
        \\[3pt]
        \includegraphics[width=\resLen]{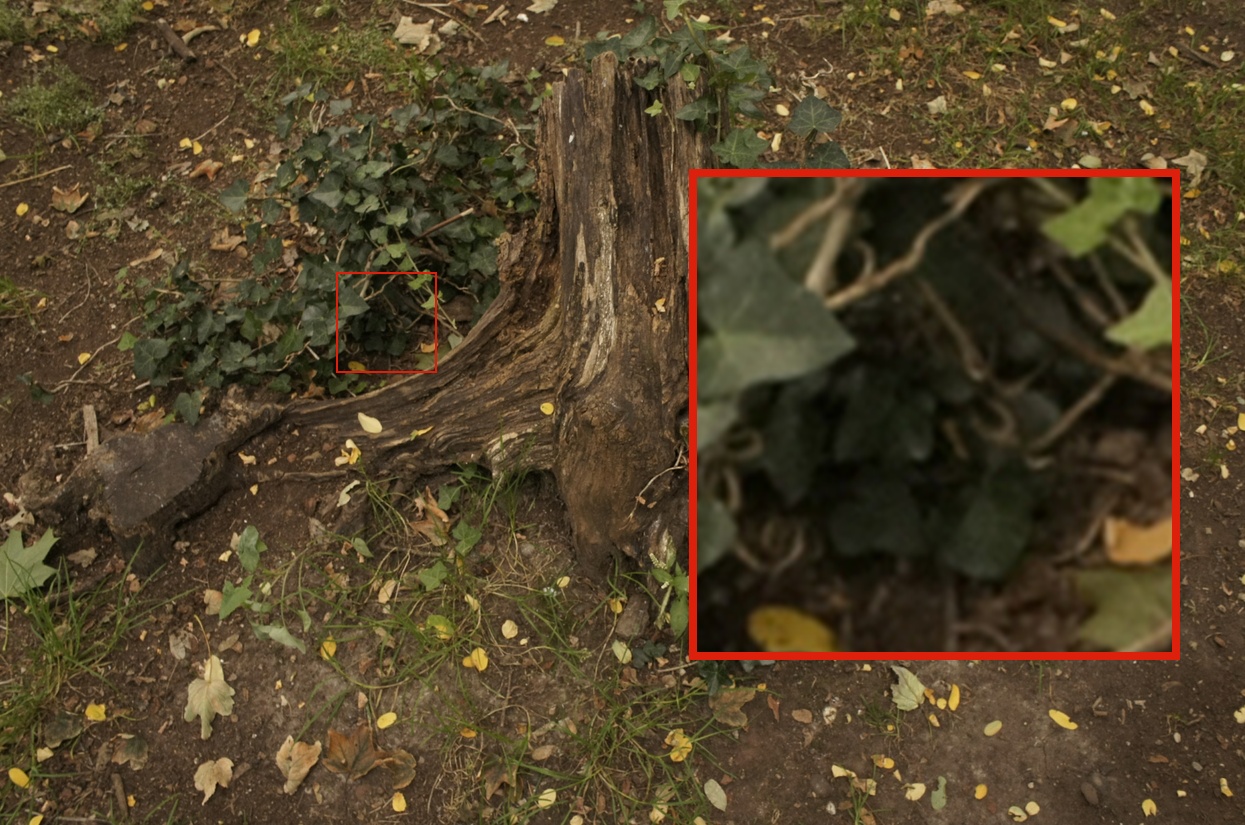} &
        \includegraphics[width=\resLen]{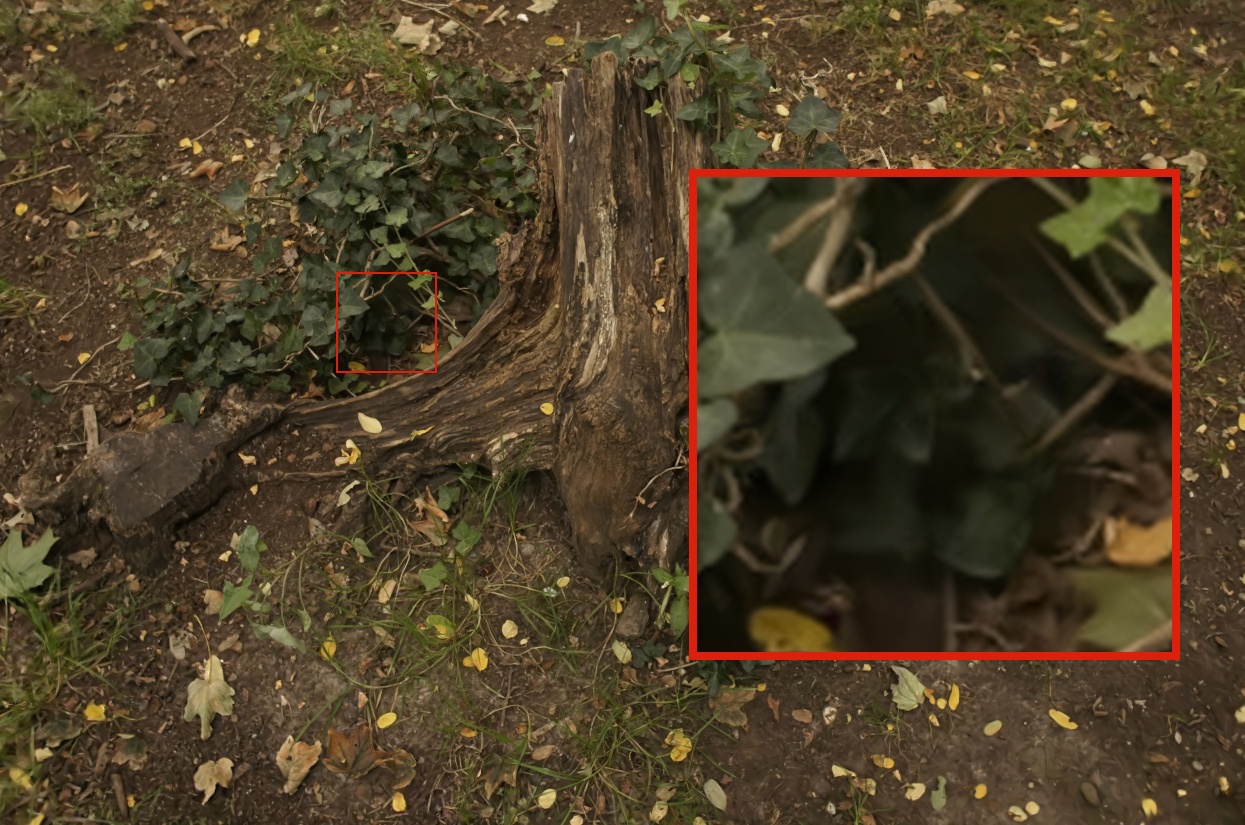} &
        \includegraphics[width=\resLen]{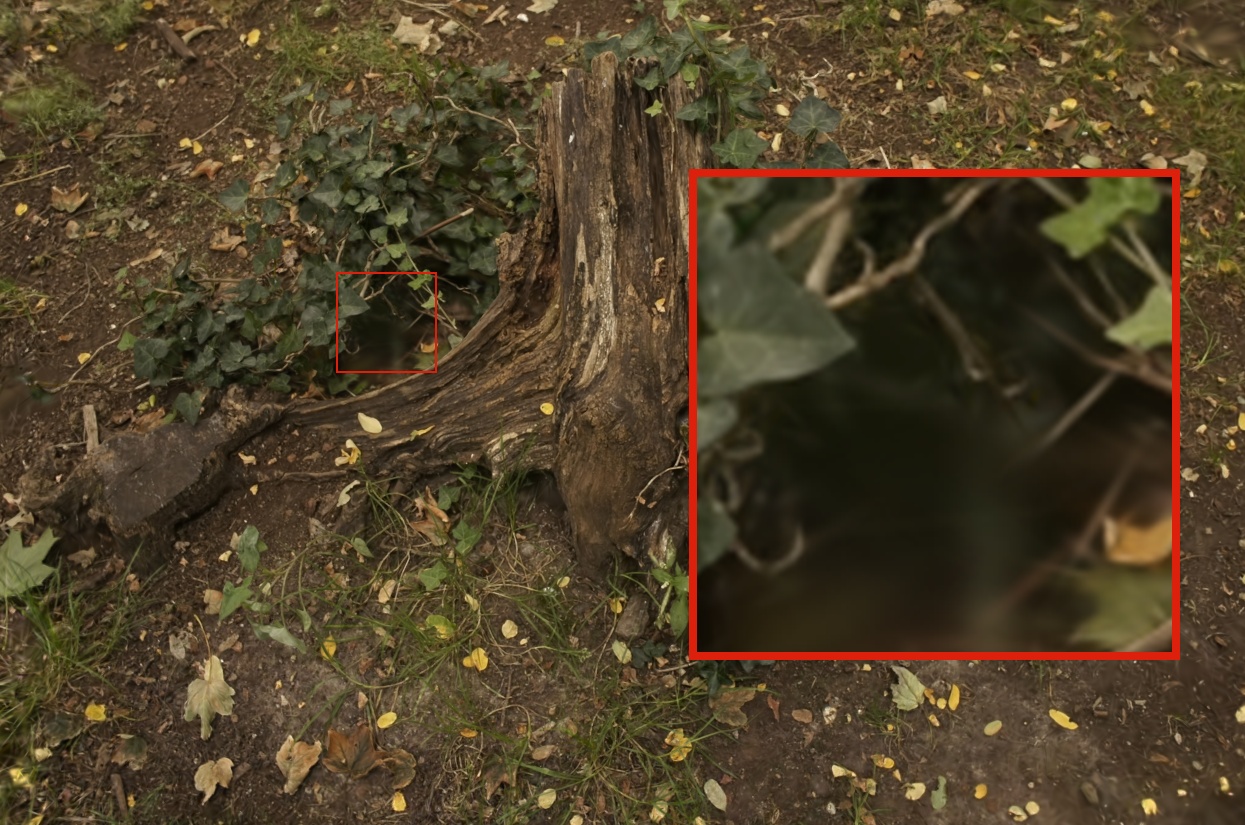} &
        \includegraphics[width=\resLen]{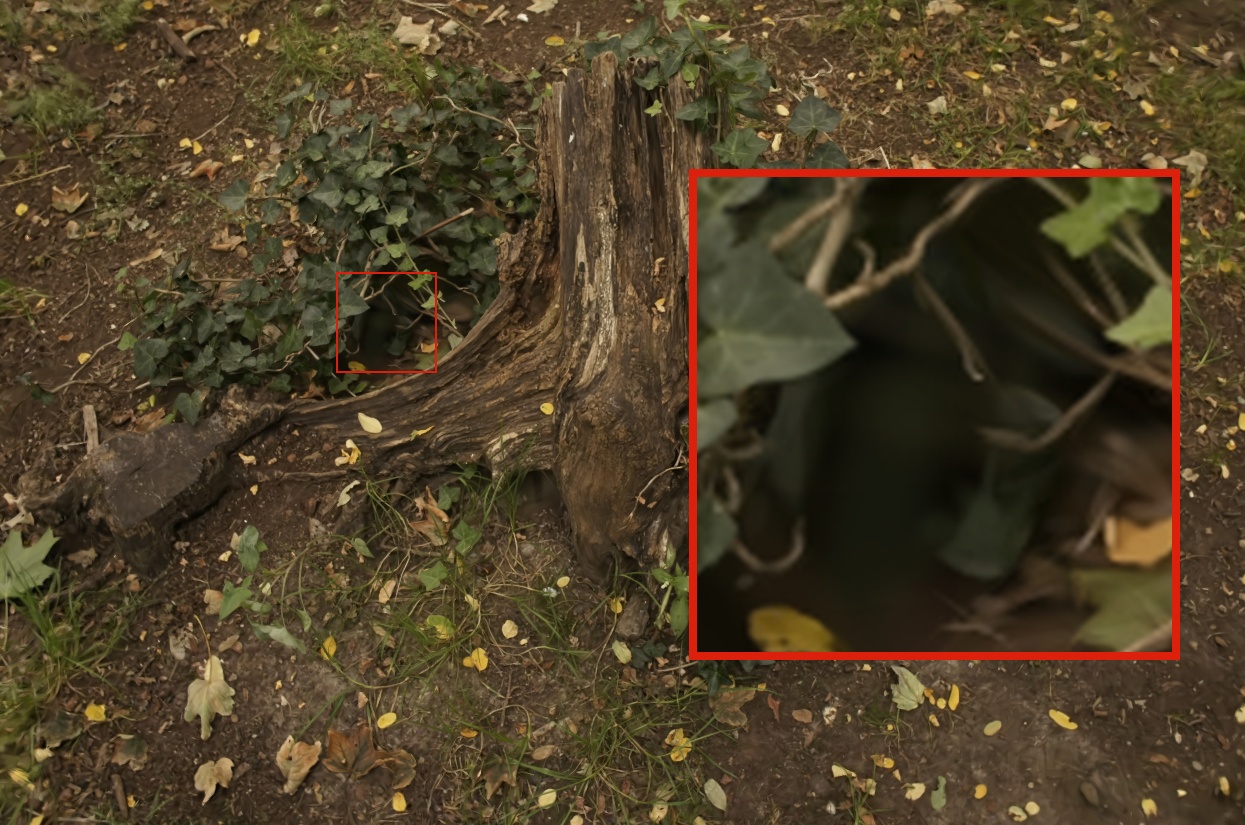} 

    \end{tabular}
    \caption{\textbf{Equal-time Comparison} between our method and baselines. We run the methods for the same period of wall-clock time, and compare the reconstruction quality. Our method produces comparable visual quality to 3DGS, and outperforms 3DGRT.
    }
    \label{fig:nvs_figures_more}
\end{figure*}

\begin{figure*}[t]
    \setlength{\resLen}{1.75in}
    \centering
    \small
    \setlength{\tabcolsep}{1pt}
    \renewcommand{\arraystretch}{0}
    \begin{tabular}{cccc}
        \textbf{Reference} & (a) \textbf{Ours} & (b) \textbf{RNG} & (c) \textbf{GS$^3$}
        \\[3pt]
        \includegraphics[width=\resLen]{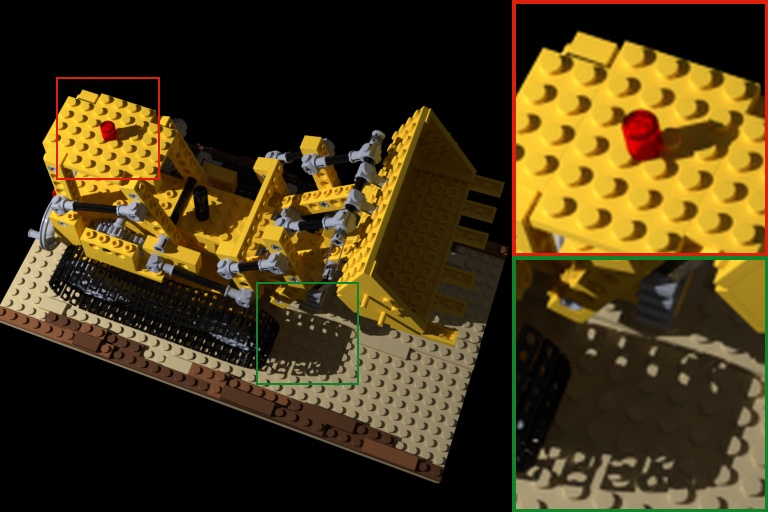} &
        \includegraphics[width=\resLen]{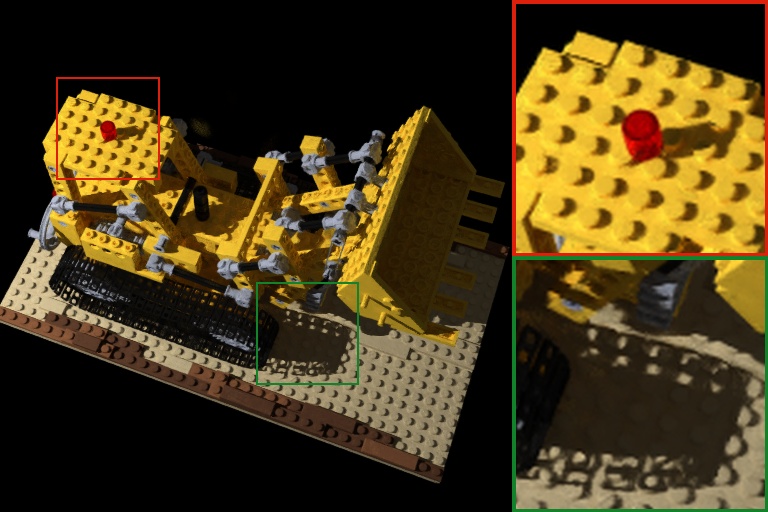} &
        \includegraphics[width=\resLen]{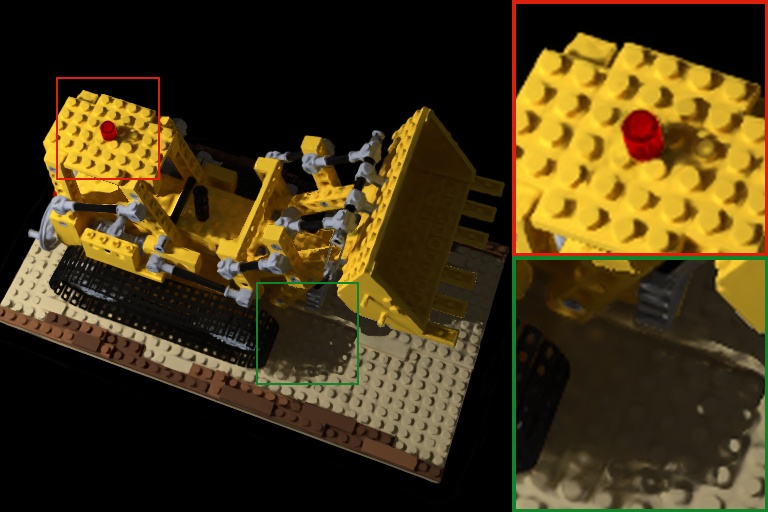} &
        \includegraphics[width=\resLen]{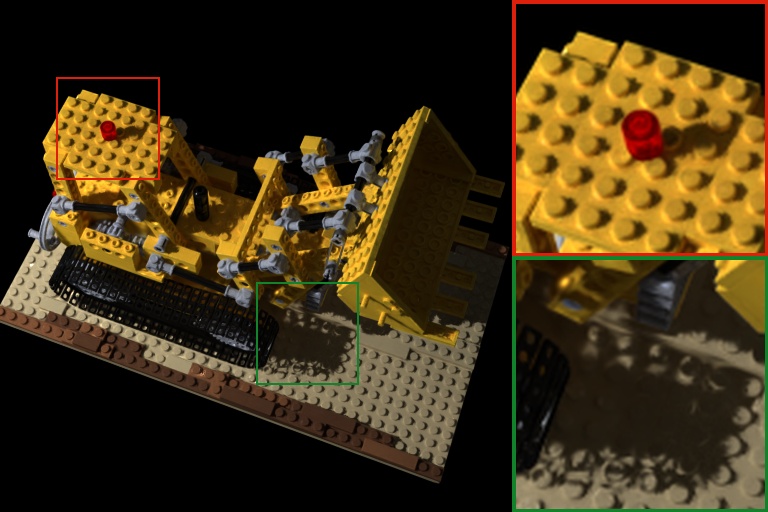}
        \\[2pt]
        \includegraphics[width=\resLen]{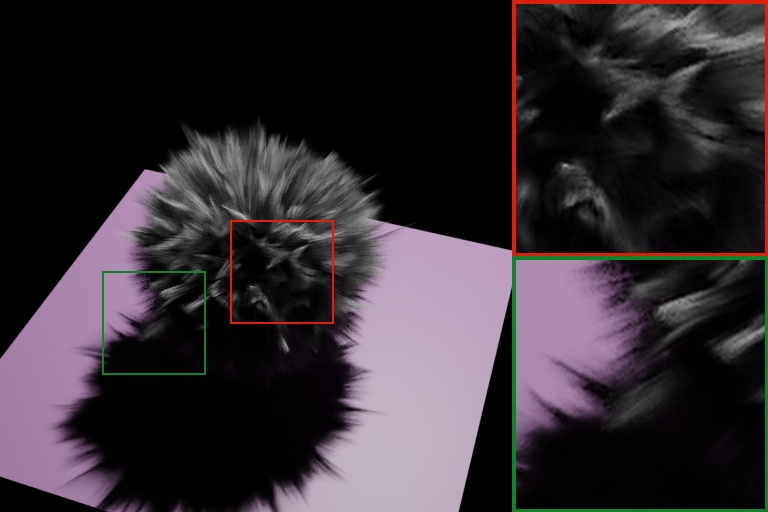} &
        \includegraphics[width=\resLen]{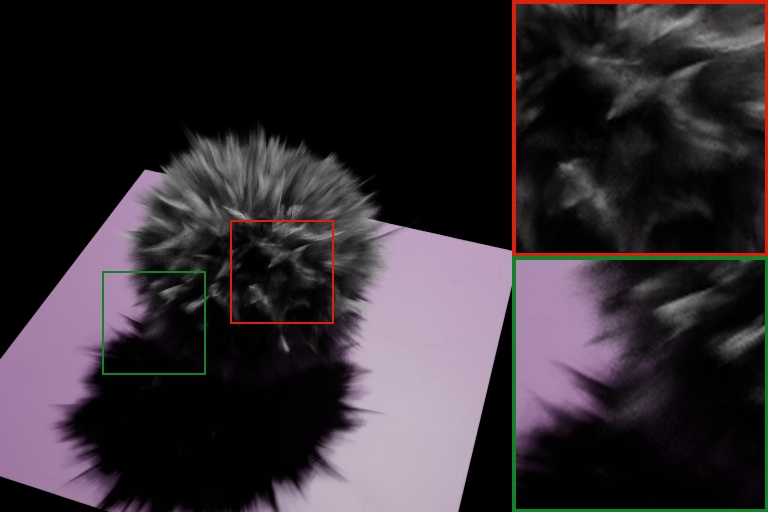} &
        \includegraphics[width=\resLen]{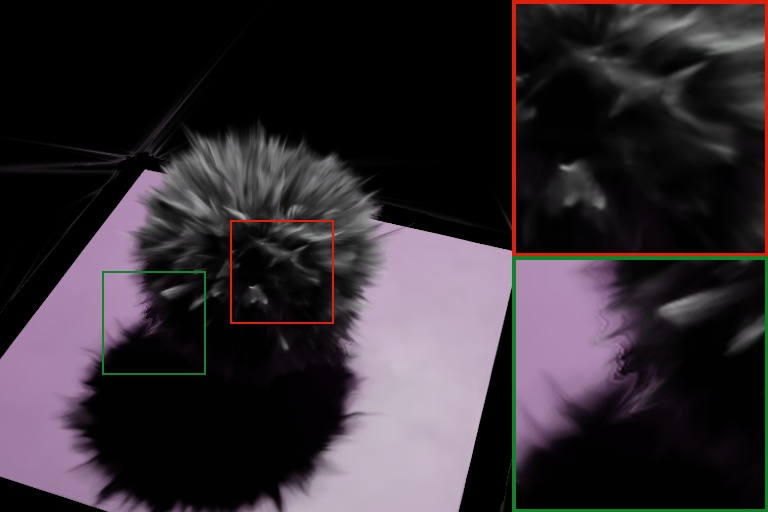} &
        \includegraphics[width=\resLen]{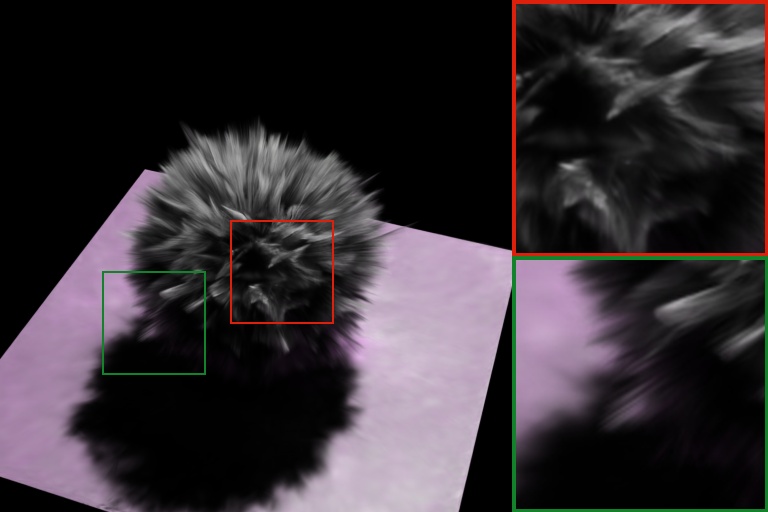}
        \\[2pt]
    \end{tabular}
    \caption{Visualization of reconstruction quality of our method and baselines. Our method produces significantly better geometry, and in particular, shadow quality.
    }
    \label{fig:relightable_figures_more}
\end{figure*}

In \autoref{fig:nvs_figures_more}, we show additional results of the equal-time novel view synthesis benchmark. In \autoref{fig:relightable_figures_more}, we show additional results from relightable Gaussian splatting evaluations. 

